%% file: Federated_Qlearning.tex
\documentclass[letter,english]{article}

\usepackage{geometry}
\usepackage[T1]{fontenc}
\usepackage[latin9]{inputenc}

\usepackage{bm}
\usepackage{amsmath}
\usepackage{amssymb} 
\usepackage[unicode=true,
 bookmarks=false,
 breaklinks=false,pdfborder={0 0 1},colorlinks=false]
 {hyperref}
\hypersetup{
 colorlinks,citecolor=blue,filecolor=blue,linkcolor=blue,urlcolor=blue}

\usepackage{multirow}
\usepackage{xcolor,colortbl}
\definecolor{Gray}{gray}{0.85}
\usepackage{enumitem}

\usepackage{amsthm}
\usepackage{cite}  
\usepackage{comment}
\usepackage{natbib}
\usepackage{booktabs,mathtools}

\usepackage{graphicx}
\usepackage[linesnumbered,ruled,vlined]{algorithm2e}
\usepackage{algorithmic}

\usepackage{float}
\usepackage{multirow}
\usepackage{cancel}

\usepackage{subcaption}

\usepackage{dsfont}
\usepackage{color}
\definecolor{yjc}{RGB}{125,0,0}
\definecolor{jiw}{RGB}{10,148,15}

\DeclareMathOperator*{\argmax}{arg\,max}

\allowdisplaybreaks

\newcommand{\mumin}{\mu_{\mathsf{min}}}
\newcommand{\Csim}{C_{\mathsf{het}}}
\newcommand{\mumax}{\mu_{\mathsf{max}}}

\newcommand{\muminavg}{\mu_{\mathsf{avg}}}

\newcommand{\tmix}{t_{\mathsf{mix}}}
\newcommand{\tmultimix}{t_{\mathsf{mix}}}

\newcommand{\tth}{t_{\mathsf{th}}}

\newcommand{\tmaxmix}{t_{\mathsf{mix}}^{\mathsf{max}}}

\newcommand{\syn}{\iota}
\newcommand{\nsyn}{\phi}

\newcommand{\Qavg}{Q}
\newcommand{\Qopt}{Q^{\star}}

\newcommand{\erravg}{{\Delta}}

\newcommand{\localvar}{\sigma_{\mathsf{local}}}

\newcommand{\fedsyncq}{{\sf{FedSynQ}}\xspace}
\newcommand{\fedqavg}{{\sf{FedAsynQ}}\xspace}
\newcommand{\fedqvavg}{{\sf{FedAsynQ-EqAvg}}\xspace}
\newcommand{\fedqwavg}{{\sf FedAsynQ-ImAvg}\xspace}

\input{macro.tex}

\allowdisplaybreaks

\title{The Blessing of Heterogeneity in Federated Q-Learning: \\ Linear Speedup and Beyond\footnotetext{Department of Electrical and Computer Engineering, Carnegie Mellon University, Pittsburgh, PA 15213, USA. Emails: {\tt\{jiinw,gaurij,yuejiec\}@andrew.cmu.edu}.}\footnotetext{A preliminary version of this paper was presented at the 2023 International Conference on Machine Learning (ICML).}}
\author{Jiin Woo \qquad Gauri Joshi \qquad Yuejie Chi  \medskip \\ 
Carnegie Mellon University}
\date{May 2023; Revised December 2023}

\begin{document}

\theoremstyle{plain} \newtheorem{lemma}{\textbf{Lemma}}
\newtheorem{proposition}{\textbf{Proposition}}
\newtheorem{theorem}{\textbf{Theorem}}
\newtheorem{assumption}{Assumption}
\newtheorem{definition}{Definition}
\newtheorem{claim}{\textbf{Claim}}
\theoremstyle{remark}\newtheorem{remark}{\textbf{Remark}}

\maketitle

\input{abstract}

\setcounter{tocdepth}{2}
\tableofcontents

\input{introduction}
\input{model}

\input{fedsyncq_results}

\input{fedasyncq_results}

\input{experiments}

\input{analysis}

\input{conclusion}

\section*{Acknowledgements}
This work is supported in part by the grants NSF CCF-2007911, CCF-2106778, CNS-2148212, and ONR N00014-19-1-2404 to Y. Chi, NSF-CCF 2007834, CCF-2045694, CNS-2112471, and ONR N00014-23-1-2149 to G. Joshi, and the Hsu Chang Memorial Fellowship at Carnegie Mellon University to J. Woo. 

\bibliographystyle{apalike}
\bibliography{bibfileRL}

\appendix

\input{preliminary}

\input{proof_syncq}

\input{fedasyncqnaive_proof_improved}

\input{fedasyncqweighted_proof}

\end{document}

%% file: macro.tex
\DeclareMathOperator{\ind}{\mathds{1}}  

\newcommand{\defn}{\coloneqq}

\newcommand{\mprob}{\mathbb{P}}
\newcommand{\mexp}{\mathbb{E}}

\newcommand{\Var}{\mathsf{Var}}

\newcommand{\bz}{\bm{z}}

\newcommand{\cA}{\mathcal{A}}
\newcommand{\cB}{\mathcal{B}}

\newcommand{\cM}{\mathcal{M}}

\newcommand{\cS}{{\mathcal{S}}}

\newcommand{\cU}{\mathcal{U}}

\newcommand{\cX}{\mathcal{X}}
\newcommand{\cY}{\mathcal{Y}}
\newcommand{\cZ}{\mathcal{Z}}

\newcommand{\mymid}{\,|\,} 

\usepackage{scalerel,stackengine}
\stackMath
\newcommand\reallywidehat[1]{%
\savestack{\tmpbox}{\stretchto{%
  \scaleto{%
    \scalerel*[\widthof{\ensuremath{#1}}]{\kern-.6pt\bigwedge\kern-.6pt}%
    {\rule[-\textheight/2]{1ex}{\textheight}}
  }{\textheight}%
}{0.5ex}}%
\stackon[1pt]{#1}{\tmpbox}%
}
\newcommand\reallywidecheck[1]{%
\savestack{\tmpbox}{\stretchto{%
  \scaleto{
    \scalerel*[\widthof{\ensuremath{#1}}]{\kern-.6pt\bigwedge\kern-.6pt}%
    {\rule[-\textheight/2]{1ex}{\textheight}}
  }{\textheight}%
}{0.5ex}}%
\stackon[1pt]{#1}{\scalebox{-1}{\tmpbox}}%
}

%% file: abstract.tex
\begin{abstract}
When the data used for reinforcement learning (RL) are collected by multiple agents in a distributed manner, federated versions of RL algorithms allow collaborative learning without the need for agents to share their local data. In this paper, we consider federated Q-learning, which aims to learn an optimal Q-function by {\em periodically} aggregating local Q-estimates trained on local data alone. Focusing on infinite-horizon tabular Markov decision processes, we provide sample complexity guarantees for both the synchronous and asynchronous variants of federated Q-learning. In both cases, our bounds exhibit a linear speedup with respect to the number of agents and near-optimal dependencies on other salient problem parameters.

 In the asynchronous setting, existing analyses of federated Q-learning, which adopt an equally weighted averaging of local Q-estimates, require that every agent covers the entire state-action space. In contrast, our improved sample complexity scales inverse proportionally to the minimum entry of the {\em average} stationary state-action occupancy distribution of all agents, thus only requiring the agents to {\em collectively cover} the entire state-action space, unveiling the {\em blessing of heterogeneity} in enabling collaborative learning by relaxing the coverage requirement of the single-agent case. However, its sample complexity still suffers when the local trajectories are highly heterogeneous. In response, we propose a novel federated Q-learning algorithm with importance averaging, giving larger weights to more frequently visited state-action pairs, which achieves a robust linear speedup as if all trajectories are centrally processed, regardless of the heterogeneity of local behavior policies. 

\end{abstract}

 \smallskip
 
 \noindent\textbf{Keywords:} federated Q-learning, periodic averaging, sample complexity, linear speedup, blessing of heterogeneity

%% file: introduction.tex
\section{Introduction}

Reinforcement Learning (RL) \citep{sutton2018reinforcement} is an area of machine learning for sequential decision making, aiming to learn an optimal policy that maximizes the total rewards via interactions with an unknown environment. RL is widely used in many real-world applications, such as autonomous driving, games, clinical trials, and recommendation systems. However, due to the high dimensionality of the state-action space, training of RL agents typically requires a significant amount of computation and data to achieve desirable performance. Moreover, data collection can be extremely time-consuming with limited access in the wild, especially when performed by a single agent.
On the other hand, it is possible to leverage multiple agents to collect data simultaneously, under the premise that they can learn a global policy collaboratively with the aid of a central server without the need of sharing local data. As a result, there is a growing need to conduct RL in a distributed or federated fashion.

Although there have been many studies analyzing federated learning \citep{kairouz2021advances} in other areas such as supervised machine learning \citep{mcmahan2017communication, bonawitz2019towards, wang2020tackling}, there are only a few recent works focused on federated RL. They consider issues such as robustness to adversarial attacks \citep{wu2021byzantine,fan2021fedrlfaulttolerant}, environment heterogeneity \citep{jin2022fedrlhete}, as well as sample and communication complexities \citep{doan2021finite,khodadadian22fedrlspeedup,shen2020a3c}. Encouragingly, some of these prior works offer non-asymptotic sample complexity analyses of federated RL algorithms that highlight a linear speedup of the required sample size in terms of the number of agents. However, the performance characterization of these federated algorithms is still far from complete.

\subsection{Federated Q-learning: prior art and limitations}
This paper focuses on Q-learning \citep{watkins1992q}, one of the most celebrated model-free RL algorithms, which aims to learn the optimal Q-function directly without forming an estimate of the model. Two sampling protocols are typically studied: synchronous sampling and asynchronous sampling. With synchronous sampling, all state-action pairs are updated uniformly assuming access to a generative model or a simulator \citep{kearns1999finite}. With asynchronous sampling, only the state-action pair that is visited by the behavior policy is updated at each time \citep{tsitsiklis1994asynchronous}. Despite its long history of theoretical investigation, the tight sample complexity of Q-learning in the single-agent setting has only recently been pinned down in \citet{li2021syncq}. As we shall elucidate, there remains a large gap in terms of the sample complexity requirement between the federated setting and the single-agent setting in terms of dependencies on salient problem parameters.

 To harness the power of multiple agents, \citet{khodadadian22fedrlspeedup} proposed and analyzed a federated variant of Q-learning with {\em asynchronous} sampling that periodically aggregates the local Q-estimates trained on local Markovian trajectories collected over $K$ agents. To set the stage, consider an infinite-horizon tabular Markov decision process (MDP) with state space $\cS$, action space $\cA$, and a discount factor $\gamma \in [0,1)$. To learn an $\varepsilon$-optimal Q-function estimate (in the $\ell_{\infty}$ sense), 
 \citet{khodadadian22fedrlspeedup} requires a per-agent sample size on the order of 
\begin{align} \label{eq:prior_fedq_bound}
\widetilde{O} \left( \frac{|\cS|^2}{K \mumin^5 (1-\gamma)^9 \varepsilon^2} \right) 
\end{align}
for sufficiently small $\varepsilon$, where $\mumin :=  \min_{1\leq k \leq K}\min_{ (s,a)\in  \cS \times \cA } \mu_{\mathsf{b}}^k(s,a) $ is the minimum entry of the stationary state-action occupancy distributions $\mu_{\mathsf{b}}^k$ of the sample trajectories over all agents, and $\widetilde{O}$ hides logarithmic terms. On the other hand, the sample requirement of single-agent Q-learning \citep{li2021syncq} for learning an $\varepsilon$-optimal Q-function is 
  \begin{align} \label{eq:asynq_bound}
    \widetilde{O} \left(
    \frac{ 1 }{\mumin (1-\gamma)^4   \varepsilon^2  } 
    \right ) 
  \end{align}
for sufficiently small $\varepsilon$.
Comparing the two sample complexity bounds reveals several drawbacks of existing analyses and raises the following natural questions.
\begin{itemize}[leftmargin=12pt] 
\item {\em Near-optimal sample size.} Despite the appealing linear speedup in terms of the number of agents $K$ shown in \citet{khodadadian22fedrlspeedup}, it has unfavorable dependencies on other salient problem parameters. In particular, since $1/\mumin \geq  |\cS||\cA|$, the sample complexity in \eqref{eq:prior_fedq_bound} will be better than that of the single-agent case in \eqref{eq:asynq_bound} only if $K$ is at least above the order of $\frac{|\cS|^6|\cA|^4}{ (1-\gamma)^5}$, which may not be practically feasible with large state-action space and long effective horizon.
 {\em Can we improve the dependency on the salient problem parameters for federated Q-learning while maintaining linear speedup?}

\item  {\em Benefits of heterogeneity.} Existing analyses in \citet{khodadadian22fedrlspeedup} require that each agent has a full coverage of the state-action space (i.e., $\mumin>0$), which is as stringent as the single-agent setting. However, given that the insufficient coverage of individual agents can be complemented by each other when agents have heterogeneous local trajectories, it may not be necessary to require full coverage of the state-action space from every agent. {\em Can we exploit the heterogeneity in the agents' local trajectories and relax the coverage requirement on individual agents?}
  \end{itemize}

\begin{table*}[t]
\centering
\resizebox{0.98\textwidth}{!}{
\begin{tabular}{c|c|c|c|c }
\toprule 
  \multirow{2}{*}{ sampling   }&   \multirow{2}{*}{  reference} & number of   &   \multirow{2}{*}{  coverage }& sample        \tabularnewline  
 &    &   agents &   &   complexity      \tabularnewline \toprule
 \multirow{3}{*}{synchronous}   &  \citet{wainwright2019stochastic,chen2020finite}   \vphantom{$\frac{1^{7^{7^7}}}{7^{7^{7^7}}}$} & $1$ & full & $\frac{ |\cS||\cA|  }{(1-\gamma)^5 \varepsilon^2} $  \vphantom{$\frac{7^{7^{7}}}{7^{7^{7^{7^{7^7}}}}}$}    \tabularnewline 
\cline{2-5}   &   \citep{li2021syncq}  \vphantom{$\frac{1^{7^{7^7}}}{7^{7^{7^7}}}$} & $1$ & full & $\frac{ |\cS||\cA|  }{(1-\gamma)^4 \varepsilon^2} $  \vphantom{$\frac{7^{7^{7}}}{7^{7^{7^{7^{7^7}}}}}$}     \tabularnewline 
\cline{2-5}  
    & {\fedsyncq (Theorem~\ref{thm:dist-syncq-local}) } \vphantom{$\frac{1^{7^{7^7}}}{7^{7^{7^7}}}$}   & $K$ & full &  $\frac{ |\cS||\cA|  }{K (1-\gamma)^5 \varepsilon^2} $  \vphantom{$\frac{7^{7^{7}}}{7^{7^{7^{7^{7^7}}}}}$}    \tabularnewline \toprule 
   &   \citet{qu2020finite}  & $1$ & full & $ \frac{\tmix}{ \mumin^2  (1-\gamma)^5 \varepsilon^2} $  \vphantom{$\frac{7^{7^{7}}}{7^{7^{7^{7^{7^7}}}}}$}   \tabularnewline
   \cline{2-5}  
   \multirow{6}{*}{asynchronous}  &   \citet{li2021asyncq} \vphantom{$\frac{1^{7^{7^7}}}{7^{7^{7^7}}}$}  & $1$ & full & $ \frac{1}{ \mumin  (1-\gamma)^5 \varepsilon^2} $   \vphantom{$\frac{7^{7^{7}}}{7^{7^{7^{7^{7^7}}}}}$}    \tabularnewline
   \cline{2-5}  
&   \citet{li2021syncq} \vphantom{$\frac{1^{7^{7^7}}}{7^{7^{7^7}}}$}  & $1$ & full & $ \frac{1}{ \mumin  (1-\gamma)^4 \varepsilon^2} $ \vphantom{$\frac{7^{7^{7}}}{7^{7^{7^{7^{7^7}}}}}$}     \tabularnewline
   \cline{2-5}  
 &  \fedqvavg \citep{khodadadian22fedrlspeedup}  \vphantom{$\frac{1^{7^{7^7}}}{7^{7^{7^7}}}$}     & $K$ & full & $ \frac{|\cS|^2}{K \mumin^5 (1-\gamma)^9 \varepsilon^2}$    \vphantom{$\frac{7^{7^{7}}}{7^{7^{7^{7^{7^7}}}}}$}  \tabularnewline
   \cline{2-5}
 &   {\fedqvavg (Theorem~\ref{thm:dist-asyncq-localupdate-tighter})}  \vphantom{$\frac{1^{7^{7^7}}}{7^{7^{7^7}}}$}   & $K$ & partial &$ \frac{\Csim}{K \muminavg  (1-\gamma)^5 \varepsilon^2} $   \vphantom{$\frac{7^{7^{7}}}{7^{7^{7^{7^{7^7}}}}}$}  \tabularnewline
   \cline{2-5}
 &  {\fedqwavg (Theorem~\ref{thm:dist-asyncwq-localupdate})} \vphantom{$\frac{1^{7^{7^7}}}{7^{7^{7^7}}}$}    & $K$ & partial &$ \frac{1}{K \muminavg  (1-\gamma)^5 \varepsilon^2} $  \vphantom{$\frac{7^{7^{7}}}{7^{7^{7^{7^{7^7}}}}}$}    \tabularnewline  \toprule 
\end{tabular} }
\caption{Comparison of sample complexity upper bounds of single-agent and federated Q-learning algorithms under synchronous and asynchronous sampling protocols to learn an $\varepsilon$-optimal Q-function in the $\ell_{\infty}$ sense, where logarithmic factors and burn-in costs are hidden.  
  Here, $\cS$ is the state space, $\cA$ is the action space, $\gamma$ is the discount factor, $K$ is the total number of agents, and $\tmix$ is the mixing time of the behavior policy. In addition, $\mumin = \min_{k,s,a} \mu_{\mathsf{b}}^k (s,a)$ denotes the minimum entry of the stationary state-action occupancy distributions $\mu_{\mathsf{b}}^k$ of all agents, $\muminavg :=  \min_{s,a}    \frac{1}{K} \sum_{k=1}^K \mu_{\mathsf{b}}^k(s,a) $ denotes the minimum entry of the average stationary state-action occupancy distribution of all agents, and $\Csim:= \max_{k,s,a}  K \mu_{\mathsf{b}}^k (s,a) / \big(  \sum_{k=1}^K \mu_{\mathsf{b}}^k(s,a) \big)$ captures the heterogeneity across the agents.  
}
\label{table:sample-comparison}
\end{table*}

\subsection{Summary of our contributions}

In this paper, we answer these questions in the affirmative, by providing a sample complexity analysis of federated Q-learning under both the synchronous and asynchronous settings. The main contributions are summarized as follows, with Table~\ref{table:sample-comparison} providing a comparison with the prior art.
\begin{itemize}[leftmargin=12pt] 
\item  {\em Sample complexity of federated synchronous Q-learning with equal averaging.} We show that with high probability, the sample complexity of federated synchronous Q-learning (\fedsyncq) to learn an $\varepsilon$-optimal Q-function in the $\ell_{\infty}$ sense is (see Theorem~\ref{thm:dist-syncq-local})
  \begin{align}
    \widetilde{O} \left(
    \frac{ |\cS||\cA|  }{K (1-\gamma)^5 \varepsilon^2}
    \right ),
  \end{align}
which exhibits a linear speedup with respect to the number of agents $K$ and nearly matches the tight sample complexity bound of single-agent synchronous Q-learning up to a factor of $1/(1-\gamma)$ in \citet{li2021syncq} for $K=1$.

\item {\em Sample complexity of federated asynchronous Q-learning with equal averaging.} We provide a sharpened sample complexity analysis of the algorithm developed in \citet{khodadadian22fedrlspeedup} for federated asynchronous Q-learning with equal averaging (\fedqvavg) that leads to new insights. To learn an $\varepsilon$-optimal Q-function in the $\ell_{\infty}$ sense, \fedqvavg requires at most (see Theorem~\ref{thm:dist-asyncq-localupdate-tighter})
  \begin{align}
   \widetilde{O} \left(
     \frac{\Csim}{K\muminavg (1-\gamma)^5 \varepsilon^2}
    \right) 
  \end{align}
samples per agent for sufficiently small $\varepsilon$ (ignoring the burn-in cost that depends on the mixing times of the Markovian trajectories over all agents), where $\muminavg =   \min_{s,a}    \frac{1}{K} \sum_{k=1}^K \mu_{\mathsf{b}}^k(s,a) \geq \mumin $ is  the minimum entry of the {\em average} stationary state-action occupancy distribution of all agents, and $\Csim = \max_{k,s,a}  \frac{ K \mu_{\mathsf{b}}^k (s,a) }{   \sum_{k=1}^K \mu_{\mathsf{b}}^k(s,a) } \in [1, 1/\muminavg] $ captures the heterogeneity of the behavior policies across agents. This sample complexity not only proves a linear speedup with respect to the number of agents, but also greatly sharpens the dependency on all the salient problem parameters --- including $1/(1-\gamma)$, $|\cS|$, and $1/\mumin$ --- by orders of magnitudes compared to the bound obtained in \citet{khodadadian22fedrlspeedup}.
More importantly, it uncovers that as long as the agents collectively cover the entire state-action space (i.e., $\muminavg>0$), \fedqvavg still enables learning even when individual agents fail to cover the entire state-action space (i.e., $\mumin=0$), unveiling the blessing of heterogeneity that was not elucidated in prior work \citep{khodadadian22fedrlspeedup}. 

\item {\em Sample complexity of federated asynchronous Q-learning with importance averaging.} Although heterogeneous behavior policies at agents may induce local trajectories covering different parts of the state-action space and relax the coverage requirement, equally weighting the local Q-estimates may hinder the convergence which is bottlenecked by the slowest converging agent. This is evident by the dependency on $\Csim$ in the sample complexity of \fedqvavg, which becomes larger when the local behavior policies are highly disparate. To address this issue, we propose a novel importance averaging scheme in federated Q-learning (\fedqwavg) that averages the local Q-estimates by assigning larger weights to more frequently updated local estimates. To learn an $\varepsilon$-optimal Q-function in the $\ell_{\infty}$ sense, \fedqwavg requires at most (see Theorem~\ref{thm:dist-asyncwq-localupdate})
  \begin{align} 
    \widetilde{O} \left(
    \frac{1}{K \muminavg  (1-\gamma)^5 \varepsilon^2}   
    \right) 
  \end{align}
  samples per agent for sufficiently small $\varepsilon$ (ignoring the burn-in cost that depends on the mixing times of the Markovian trajectories over all agents). This improves over that of \fedqvavg by removing the dependency on $\Csim$, which can be as large as $1/\muminavg$. More importantly, this suggests that \fedqwavg achieves stable linear speedup with respect to the profile of the local behavior policies while maintaining the blessing of heterogeneity that eases the burden of individual agents' coverage.

\end{itemize}

\subsection{Related work}

\paragraph{Analysis of single-agent Q-learning.} 
There has been extensive research on the convergence guarantees of Q-learning, focusing on the single-agent case.
Many initial studies have analyzed the asymptotic convergence of Q-learning \citep{tsitsiklis1994asynchronous,szepesvari1998asymptotic,jaakkola1994convergence,borkar2000ode}.
Later, \citet{even2003learning, beck2012error, wainwright2019stochastic,chen2020finite,li2021syncq} have studied the sample complexity of Q-learning under synchronous sampling, and \citet{even2003learning,beck2012error,qu2020finite,li2021syncq, li2021asyncq, chen2021lyapunov} have investigated the finite-time convergence of Q-learning under asynchronous sampling (also referred to as Markovian sampling). In addition, \citet{jin2018q,bai2019provably,zhang2020almost,li2021breaking,yang2021q} studied Q-learning with optimism for online RL, and \citet{shi2022pessimistic,yan2022efficacy} dealt with Q-learning with pessimism for offline RL.

\paragraph{Distributed and federated RL.} Several recent works have developed distributed versions of RL algorithms to accelerate training \citep{mnih2016a3c,espeholt2018impala,assran2019gossip}. Theoretical analysis of convergence and communication efficiency of these distributed RL algorithms have also been considered in recent works. For example, a collection of works
\citep{doan2019disttd,sun2020decentd,wang2020decentd,wai2020decentdsa,chen2021disttd,zeng2020decensamtrl} have analyzed the convergence of decentralized temporal difference (TD) learning.
Furthermore, \citet{chen2021dista3c,shen2020a3c} have analyzed the finite-time convergence of distributed actor-critic algorithms and \citet{chen2021distpg} proposed a communication-efficient policy gradient algorithm with provable convergence guarantees.

\paragraph{Notation.}  Throughout this paper, we denote by $\Delta(\cS)$ the probability simplex over a set $\cS$, and $[K]\coloneqq \{1,\cdots,K\}$ for any positive integer $K>0$. In addition, $f(\cdot)=\widetilde{O} ( g(\cdot) )$ or $f \lesssim g $ (resp.~$f(\cdot)=\widetilde{\Omega} ( g(\cdot) )$ or $f\gtrsim g$) 
means that $f(\cdot)$ is orderwise no larger than (resp.~no smaller than) $g(\cdot)$ modulo some logarithmic factors. The notation $f \asymp g $ means  $f \lesssim g$ and $f \gtrsim g$ hold simultaneously. 
 

%% file: model.tex

\section{Model and background}

In this section, we introduce the mathematical model and background of Markov decision processes.

\paragraph{Infinite-horizon Markov decision process.}
We consider an infinite-horizon Markov decision process (MDP), which is represented by $\mathcal{M} = (\cS,\cA, P, r,\gamma).$ Here, $\cS$ and $\cA$ denote the state space and the action space, respectively, $P: \cS\times\cA \times \cS \rightarrow [0,1]$ indicates the transition kernel such that $P(s' \,|\, {s,a})$ denotes the probability that action $a$ in state $s$ leads to state $s'$, $r:  \cS\times\cA \rightarrow [0,1]$ denotes a deterministic reward function, where $r(s,a)$ is the immediate reward for action $a$ in state $s,$ and $\gamma\in [0,1)$ is the discount factor.

\paragraph{Policy, value function, and Q-function.}
A {\em policy} is an action-selection rule denoted by the mapping $\pi: \cS \rightarrow \Delta(\cA)$, such that $\pi(a| s)$ is the probability of taking action $a$ in state $s$. For a given policy $\pi,$ the {\em value function} $V^{\pi}: \cS \rightarrow \mathbb{R}$, which measures the expected discounted cumulative reward from an initial state $s$, is defined as
\begin{align} \label{eq:def_V}
   \forall s\in \cS &: \qquad 
   V^{\pi}(s) \defn \mathbb{E} \left[ \sum_{t=0}^{\infty} \gamma^t r(s_t,a_t ) \,\big|\, s_0 =s \right] . 
\end{align}
Here, the expectation is taken with respect to the randomness of the trajectory $\{s_t,a_t,r_t\}_{t=0}^{\infty}$, sampled based on the transition kernel (i.e., $s_{t+1}\sim P(\cdot | s_t, a_t)$) and the policy $\pi$ (i.e., $a_t \sim \pi(\cdot|s_t)$) for any $t \ge 0$. Similarly, the state-action value function (i.e., {\em Q-function}) $Q^{\pi}: \cS \times \cA \rightarrow \mathbb{R}$, which measures the expected discounted cumulative reward from an initial state-action pair $(s,a)$, is defined as   
\begin{align*}
  \forall (s,a)\in \cS \times \cA &: \qquad 
  Q^{\pi}(s,a) \defn r(s ,a  )  + \mathbb{E} \left[  \sum_{t=1}^{\infty} \gamma^t r(s_t,a_t ) \,\big|\, s_0 =s, a_0 = a \right].
\end{align*}
Again here, the expectation is taken with respect to the randomness of the trajectory $\{s_t,a_t,r_t\}_{t=1}^{\infty}$ generated similarly as above. Since the rewards lie within $[0,1]$, it follows that for any policy $\pi$, 
\begin{equation}
 0\leq V^{\pi} \leq \frac{1}{1-\gamma}  , \qquad 0 \leq Q^{\pi} \leq \frac{1}{1-\gamma}.
\end{equation}

\paragraph{Optimal policy and Bellman's principle of optimality.} 
A policy that maximizes the value function uniformly over all states is called an {\em optimal policy} and denoted by $\pi^{\star}$. Note that the existence of such an optimal policy is always guaranteed \citep{puterman2014markov}, which also maximizes the Q-function simultaneously. The corresponding optimal value function and Q-function are denoted by $V^{\star} :=V^{\pi^{\star}}$ and $Q^{\star} := Q^{\pi^{\star}}$, respectively.  It is well-known that the optimal Q-function $Q^{\star}$ can be determined as the unique fixed point of the Bellman operator $\mathcal{T}$, given by
 \begin{equation} \label{eq:bellman}
 	\mathcal{T}(Q)(s,a) := r(s,a) + \gamma \mathop{\mathbb{E}}\limits_{s^{\prime} \sim P(\cdot| s,a)}  \Big[ \max_{a^{\prime}\in \cA} Q(s^{\prime}, a^{\prime}) \Big].
 \end{equation}
 Q-learning \citep{watkins1992q}, perhaps the most widely used model-free RL algorithm, seeks to learn the optimal Q-function based on samples collected from the underlying MDP without estimating the model.

%% file: fedsyncq_results.tex
\section{Federated synchronous Q-learning: algorithm and theory}
\label{sec:fed_syncQ}
In this section, we begin with understanding federated synchronous Q-learning, where all the state-action pairs are updated simultaneously  assuming access to a generative model or simulator at all the agents.

\subsection{Problem setting}
  
In the synchronous setting, each agent $ k \in [K]$ has access to a generative model, and generates a new sample
 \begin{equation}\label{eq:sync_sampling}  
s_t^k(s,a) \sim P(\cdot|s,a)
 \end{equation}
 for every state-action pair $(s,a) \in \cS \times \cA$ {\em independently} at every iteration $t$.
Our goal is to learn the optimal Q-function $Q^{\star}$ collaboratively by aggregating the local Q-learning estimates {\em periodically}.
 
\paragraph{Review: synchronous Q-learning with a single agent.} To facilitate algorithmic development, let us recall the synchronous Q-learning update rule with a single agent. Starting with certain initialization $Q_0$, at every iteration $t\geq 1$, the Q-function is updated according to
\begin{align} \label{eq:sync-q-learning}
 \forall (s,a)\in \cS \times \cA : \qquad Q_{t}(s, a) 
  &= (1-\eta) Q_{t-1}(s,a) + \eta \left(r(s,a) + \gamma \max_{a' \in \cA} Q_{t-1}(s_t(s,a),a') \right),
\end{align}
where $s_t(s,a)\sim P(\cdot|s,a)$ is drawn independently for every state-action pair $(s,a) \in \cS \times \cA$, and $\eta$ denotes the constant learning rate. The sample complexity of synchronous Q-learning has been recently investigated and sharpened in a number of works, e.g. \citet{li2021syncq,wainwright2019stochastic,chen2020finite}.

\subsection{Algorithm description}
We propose a natural federated synchronous Q-learning algorithm called \fedsyncq that alternates between local updates at agents and periodic averaging at a central server. The complete description is summarized in Algorithm~\ref{alg:dist-syncq-local}. \fedsyncq initializes a local Q-function as $Q_0^k = Q_0$ at each agent $k \in [K]$. Suppose at the beginning of each iteration $t\geq 1$, each agent maintains a local Q-function estimate $Q_{t-1}^{k}$ and a local value function estimate $V_{t-1}^k$, which are related via 
\begin{align}	\label{defn:Vt}
  \forall s\in \cS: \qquad
  V_t^k(s) :=  \max_{a\in \cA} Q_{t}^k(s,a).
\end{align} 
\fedsyncq proceeds according to the following steps in the rest of the $t$-th iteration.
\begin{enumerate}[leftmargin=12pt] 
\item {\em Local updates:} Each agent first independently updates {\em all} entries of its Q-estimate $Q_{t-1}^k$ to reach some {\em intermediate} estimate following the update rule:
\begin{align} \label{eq:syncq-local}
\forall (s,a) \in \cS \times \cA: \qquad  Q_{t-\frac{1}{2}}^k(s, a) &= (1-\eta) Q_{t-1}^k(s,a) + \eta \left( r(s,a) + \gamma   V_{t-1}^k(s_t^k(s,a) )  \right) ,  
\end{align}
where $s_t^k(s,a)$ is drawn according to \eqref{eq:sync_sampling}, and $\eta \geq 0$ is the learning rate.

\item {\em Periodic averaging:} These intermediate estimates will be periodically averaged by the server to form the updated estimate $Q_{t}^k$ at the end of the $t$-th iteration. Formally, denoting $\tau\geq 1$ as the synchronization period, it follows
\begin{align} \label{eq:syncq-periodic}
\forall (s,a) \in \cS \times \cA: \qquad  Q_{t}^k(s,a) =
  \begin{cases}
    \frac{1}{K} \sum_{k=1}^K Q_{t-\frac{1}{2}}^k(s,a) &~\text{if}~ t \equiv 0 \; (\text{mod}~ \tau)\\
    Q_{t-\frac{1}{2}}^k(s,a) &~\text{otherwise}
  \end{cases}.
\end{align}
\end{enumerate}
Denoting the number of total iterations by $T$, the algorithm outputs the final Q-estimate as the average of all local estimates, i.e. ${Q}_{T}=\frac{1}{K}\sum_{k}Q_T^k$. Without loss of generality, we assume the total number of iterations $T $ is divisible by $\tau$, where $C_{\mathsf{round}} = T/\tau$ is the rounds of communication. 

\begin{algorithm}[ht]
  \begin{algorithmic}[1] 
    \STATE \textbf{inputs:} learning rate $\eta$, discount factor $\gamma$, number of agents $K$, synchronization period $\tau$, number of iterations $T$. 
    \STATE  \textbf{initialization:} $Q_0^k = Q_0$ for all $k$. 
    \FOR{$t=1,\cdots,T$}
    \FOR{$k \in [K]$}    
    \STATE{ Draw  $s_t^k(s,a) \sim  P(\cdot \mymid s,a)$ for all $(s,a)\in \cS\times \cA$. }
    \STATE{ Compute $Q_{t-\frac{1}{2}}^k$ according to \eqref{eq:syncq-local}.}
    \STATE{ Compute $Q_t^k$ according to \eqref{eq:syncq-periodic}.}
    \ENDFOR
    \ENDFOR     
    \STATE \textbf{return:} ${Q}_{T}=\frac{1}{K}\sum_{k}Q_T^k$. 
  \end{algorithmic} 
  \caption{Federated Synchronous Q-learning (\fedsyncq)}
  \label{alg:dist-syncq-local}
\end{algorithm}

\subsection{Performance guarantee}
We are ready to provide the finite-time convergence analysis of Algorithm~\ref{alg:dist-syncq-local}.
\begin{theorem}[Finite-time convergence of \fedsyncq] \label{thm:dist-syncq-local}
Consider any given $\delta \in (0,1)$ and $\varepsilon\in (0, \frac{1}{1-\gamma}]$.
 Suppose that the initialization of Algorithm~\ref{alg:dist-syncq-local} satisfies $0 \leq Q_0  \le \frac{1}{1-\gamma}$,  
and the synchronization period $\tau$ obeys
  \begin{subequations}
\begin{align}\label{eq:sync_period}
 \tau \le  1+ \frac{1}{\eta} \min \left\{  \frac{1-\gamma}{8\gamma}, \, \frac{1}{K} \right\}. 
 \end{align}
There exist some sufficiently large constant $c_T>0$ and sufficiently small constant $c_\eta>0$, such that with probability at least $1-\delta$, the output of Algorithm~\ref{alg:dist-syncq-local} satisfies $\| Q_T - Q^{\star}\|_{\infty}\leq \varepsilon$, provided that the sample size per agent $T$ and the learning rate $\eta$ satisfy
  \begin{align}\label{eq:guarantee_sync}
  T
  &\ge  \frac{c_T}{K(1-\gamma)^5 \varepsilon^2} (\log((1-\gamma)^2\varepsilon))^2 \log{\frac{|\cS||\cA|K T}{\delta}}, \\
 \eta & = c_{\eta}        K (1-\gamma)^4 \varepsilon^2 \frac{1}{\log{\frac{|\cS||\cA|K T}{\delta}}}. \label{eq:lr_sync}
\end{align}
\end{subequations}
\end{theorem}
 
Theorem~\ref{thm:dist-syncq-local} suggests that to achieve an $\varepsilon$-accurate Q-function estimate in an $\ell_\infty$ sense, the number of samples required at each agent is no more than 
\begin{align*}
\widetilde{O}\left( \frac{|\cS| |\cA|}{K(1-\gamma)^5 \varepsilon^2}   \right),
\end{align*}
given that the agent collects $|\cS||\cA|$ samples at each iteration. A few implications are in order.

\paragraph{Linear speedup.} The sample complexity exhibits an appealing linear speedup with respect to the number of agents $K$. In comparison, the sharpest upper bound known for single-agent Q-learning \citep{li2021syncq} is $\widetilde{O}\left( \frac{|\cS| |\cA|}{(1-\gamma)^4 \min\{ \varepsilon, \varepsilon^2 \} }   \right)$, which matches with its algorithmic-dependent lower bound when $\varepsilon \in (0,1)$. Therefore, our federated setting enables faster learning as soon as the number of agents satisfies
$$ K \gtrsim \frac{1}{(1-\gamma)\max\left\{1, \varepsilon \right\}} $$
up to logarithmic factors. 
When $K=1$, our bound nearly matches with the lower bound of single-agent Q-learning up to a factor of $1/(1-\gamma)$, indicating its near-optimality.

\paragraph{Communication efficiency.} One key feature of our federated setting is the use of periodic averaging with the hope to improve communication efficiency. According to \eqref{eq:sync_period}, our theory requires that the synchronization period $\tau$ be inversely proportional to the learning rate $\eta$, which suggests that more frequent communication is needed to compensate the discrepancy of local updates when the learning rate is large. To provide insights, consider the parameter regime when $K\gtrsim \frac{1}{1-\gamma}$ and $\varepsilon \lesssim \frac{1}{K(1-\gamma)^2}$. Plugging the choice of the learning rate \eqref{eq:lr_sync} into the upper bound of $\tau$ in  \eqref{eq:sync_period}, we can choose the synchronization period as
$\tau \asymp \frac{1}{K^2(1-\gamma)^4\varepsilon^2}$ up to logarithmic factors, leading to a communication complexity no larger than
$ C_{\mathsf{round}}  = \frac{T}{\tau} \lesssim  \frac{K}{1-\gamma} $,
which is almost independent of the final accuracy $\varepsilon$.

%% file: fedasyncq_results.tex
\section{Federated asynchronous Q-learning: algorithm and theory}
\label{sec:fed_asyncQ}
In this section, we study the sample complexity of federated asynchronous Q-learning, where $K$ agents sample local trajectories using different behavior policies. 
In particular, we propose a novel aggregation algorithm \fedqwavg that leverages the heterogeneity of these policies and dramatically improves the sample complexity.

\subsection{Problem setting}

In the asynchronous setting, each agent $k\in [K]$ independently collects a sample trajectory $\{s_t^k,a_t^k,r_t^k\}_{t=0}^{\infty}$ from the same underlying MDP $\cM$ following some stationary {\em local} behavior policy $\pi_{\mathsf{b}}^k$ such that
\begin{align} \label{eq:sampling_async}
a_t^k \sim \pi_{\mathsf{b}}^k( \cdot | s_t^k), ~~~ r_t^k = r(s_t^k, a_t^k), ~~~ s_{t+1}^k \sim P(\cdot |  s_t^k, a_t^k)
\end{align}
for all $t\geq 0$, where the initial state is initialized as $s_0^k$ for each agent $k$.
Note that the behavior policies $\{\pi_{\mathsf{b}}^k\}_{k\in [K]}$ are heterogeneous across agents and can be different from the optimal policy $\pi^{\star}$. Contrary to the generative model considered in the synchronous setting, the samples collected under the asynchronous setting are no longer independent across time but are Markovian, making the analysis significantly more challenging. The sample trajectory at each agent can be viewed as sampling a time-homogeneous Markov chain over the set of state-action pairs. Throughout this paper, we make the following standard uniform ergodicity assumption \citep{paulin2015concentration,li2021asyncq}. 

\begin{assumption} [Uniform ergodicity]
  \label{assumption:uniform-ergodic}
  For every agent $k\in [K]$, the Markov chain induced by the stationary behavior policy $\pi_{\mathsf{b}}^k$ is uniformly ergodic over the entire state-action space $\cS \times \cA.$
\end{assumption}
Uniform ergodicity guarantees that the distribution of the state-action pair $(s_t, a_t)$ of a trajectory converges to the stationary distribution of the Markov chain geometrically fast regardless of the initial state-action pair, and eventually, each state-action pair is visited in proportion to the stationary distribution.

\paragraph{Key parameters.}
 Two important quantities concerning the resulting Markov chains will govern the performance guarantees. The first one is the stationary state-action distribution $\mu_{\mathsf{b}}^k$, which is the stationary distribution of the Markov chain induced by $\pi_{\mathsf{b}}^k$ over all state-action pairs; the second one is $\tmix^k$, which is the mixing time of the same Markov chain given by
\begin{equation} \label{def:mixing-time}
	\tmix^k := \min \Big\{ t ~\Big|~  \max_{ (s_0,a_0 ) \in \mathcal{S}\times \mathcal{A}} d_{\mathsf{TV}}\big( P_t^k (\cdot \,|\, s_0, a_0), \, \mu_{\mathsf{b}}^k \big) \leq \frac{1}{4} \Big\}  , 
\end{equation}
where $P_t^k(\cdot\,|\,s_0,a_0)$ denote the distribution of $(s_t, a_t)$ conditioned on $(s_0,a_0)$ for agent $k$, and $d_{\mathsf{TV}}(\cdot, \cdot)$ is the total variation distance. Further, let the largest mixing time of all the Markov chains induced by local behavior policies be
\begin{align}  \label{defn:mixing-time-MC}
  \tmaxmix \defn \max_{k \in [K]} \tmix^k.
\end{align}
In words, $\tmaxmix$ approximately indicates the time that the transition of every agent starts to follow its stationary distribution regardless of its initial state.
 
Let us further define a few key parameters that measure the coverage and heterogeneity of the stationary state-action distribution $\mu_{\mathsf{b}}^k$ across agents. First, define 
\begin{align}
  \label{defn:mu-min}
  \mumin &\defn \min_{k \in [K]} \; \mumin^k, \qquad\qquad\mbox{where}\qquad \mumin^k \defn \min_{(s,a)\in  \cS \times \cA } \mu_{\mathsf{b}}^k(s,a) .
\end{align}
State-action pairs with small stationary probabilities are visited less frequently, and therefore can become bottlenecks in improving the quality of Q-function estimates. Clearly, $\mumin \leq \frac{1}{|\cS||\cA|}$. In addition, denote
\begin{align}  \label{defn:mu-avg}
  \muminavg &\defn \min_{(s,a) \in \cS \times \cA}  \frac{1}{K} \sum_{k=1}^K \mu_{\mathsf{b}}^k(s,a)  .
\end{align}
In words, $\muminavg$ is the minimum entry of the {\em average} stationary state-action distribution of all agents. 
The difference between $\muminavg$ and $\mumin$ stands out when an individual agent fails to cover the entire state-action space. While $\mumin=0$ in such a case, $\muminavg$ can still be positive as long as each state-action pair is explored by at least one of the agents, i.e., $\sum_{k=1}^K \mu_{\mathsf{b}}^k(s,a)>0$. Note that $\muminavg$ is always greater than or equal to $\mumin$ since
  \begin{align} \label{eq:asyncwq-mu-comp}
    \muminavg 
    &= \min_{(s,a) \in \cS \times \cA} \frac{1}{K} \sum_{k=1}^K\mu_{\mathsf{b}}^k(s,a) \ge \min_{(s,a) \in \cS \times \cA, k \in [K]} \mu_{\mathsf{b}}^k(s,a) = \mumin.
  \end{align}
Last but not least, we measure the heterogeneity of the stationary state-action distributions across agents by
\begin{align} \label{eq:def_csim}
  \Csim \defn \max_{k \in [K]} \max_{(s,a) \in \cS \times \cA }\frac{ \mu^k_{\mathsf{b}}(s,a)}{ \frac{1}{K} \sum_{k=1}^K\mu_{\mathsf{b}}^k(s,a) },
\end{align}
which satisfies $1\leq \Csim \leq \min \{ K, \, 1/\muminavg\}$, and in particular, $\Csim =1$ when $\mu_b^k = \mu_b$ are all equal.

\paragraph{Review: asynchronous Q-learning with a single agent.}
Recall the update rule of asynchronous Q-learning with a single agent, where at each iteration $t \ge 1$, upon receiving a transition $(s_{t-1}, a_{t-1}, s_{t})$, the Q-estimate is updated via
\begin{align} \label{eq:async-q-learning}
  Q_{t}(s, a) 
  &=\left\{ \begin{array}{ll}
  (1-\eta) Q_{t-1}(s,a) + \eta \left(r(s,a) + \gamma \max_{a' \in \cA} Q_{t-1}(s_{t},a') \right), & \mbox{if~}(s,a) = (s_{t-1}, a_{t-1}), \\
  Q_t(s,a) & \mbox{otherwise},
  \end{array} \right.
\end{align}
where $\eta$ denotes the learning rate and $V_t$ is defined in \eqref{defn:Vt}. The sample complexity of asynchronous Q-learning has been recently investigated in \citet{li2021asyncq,li2021syncq,qu2020finite}.

\begin{figure}[t]
\centering
\includegraphics[width=0.56\textwidth]{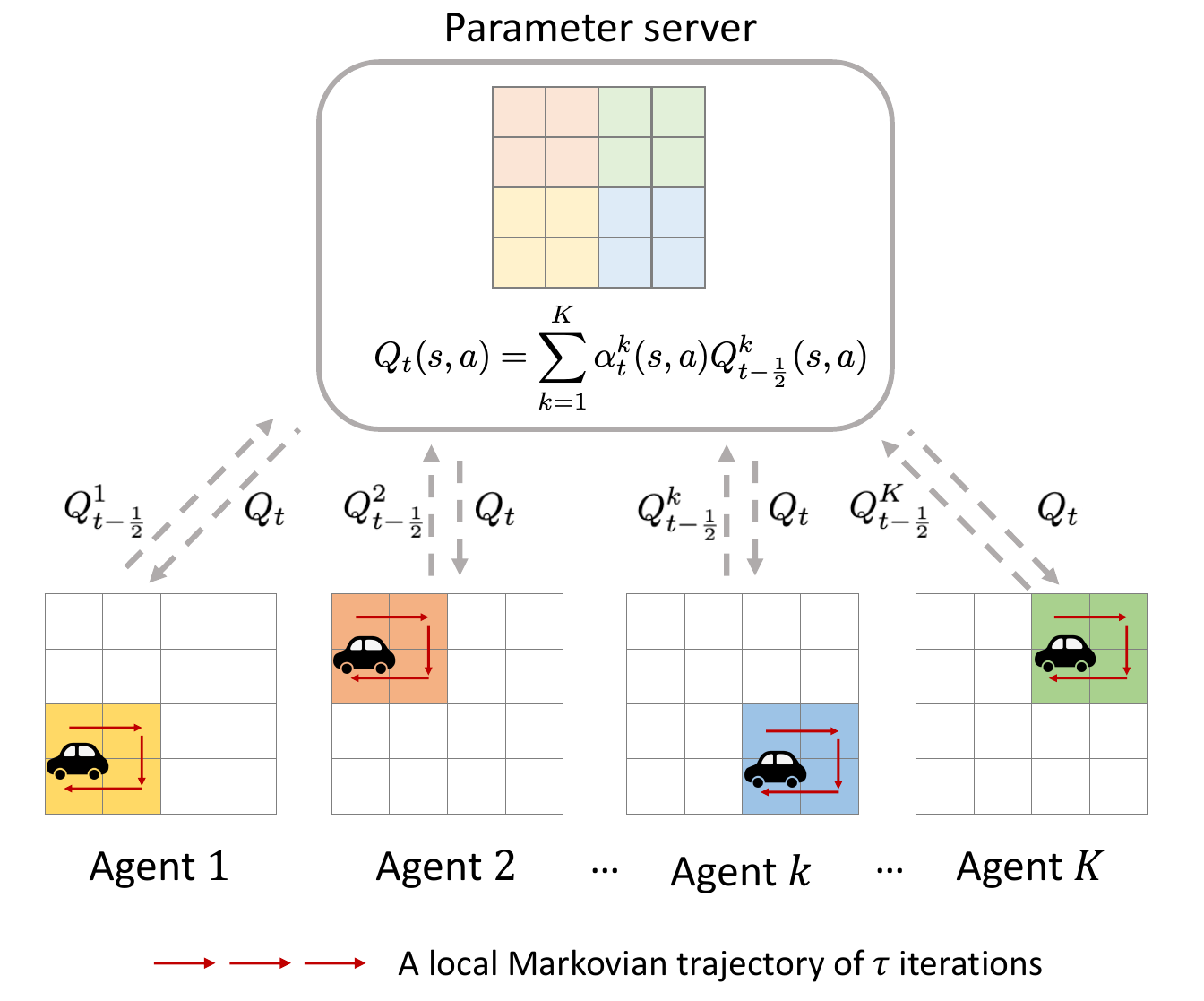}
\caption{Federated asynchronous Q-learning with $K$ agents and a parameter server. Each agent $k$ performs $\tau$ local updates on its local Q-table along a Markovian trajectory induced by behavior policy $\pi_{\mathsf{b}}^k$ and sends the Q-table to the server. The server averages and synchronizes the local Q-tables every $\tau$ iterations. For importance averaging, the agents additionally send the number of visits over all the state-action pairs within each synchronization period, which is not pictured. }
\label{fig:fedq-structure}
\end{figure}

\subsection{Algorithm description}
Similar to the synchronous setting, we describe a federated asynchronous Q-learning algorithm, called \fedqavg (see Algorithm~\ref{alg:dist-asyncq-local}), that learns the optimal Q-function by periodically averaging the local Q-estimates with the aid of a central server. See Figure~\ref{fig:fedq-structure} for an illustration. Inheriting the notation of $Q_t^k$ and $V_t^k$ from the synchronous setting (cf.~\eqref{defn:Vt}), \fedqavg proceeds as follows in the rest of the $t$-th iteration. 
\begin{enumerate}[leftmargin=12pt]
\item {\em Local updates:} Each agent $k$ samples a transition $(s_{t-1}^k, a_{t-1}^k,r_{t-1}^k, s_{t}^k)$ from its Markovian trajectory generated by the behavior policy $\pi_{\mathsf{b}}^k$ according to \eqref{eq:sampling_async} and updates a {\em single} entry of its local Q-estimate $Q_{t-1}^k$:
\begin{align} \label{eq:asyncq-local}
  Q_{t-\frac{1}{2}}^k(s, a) =
  \begin{cases}
    (1-\eta) Q_{t-1}^k(s,a) + \eta \big( r_{t-1}^k + \gamma  V_{t-1}^k(s_t^k) \big) &\quad   \mbox{if~}(s,a) = (s_{t-1}^k,a_{t-1}^k) \cr 
    Q_{t-1}^k(s,a), &\quad \mbox{otherwise}
  \end{cases},
\end{align}
where $\eta$ denotes the learning rate.

\item {\em Periodic averaging:} The intermediate local estimates will be averaged every $\tau$ iterations, where $\tau \ge 1$ is the synchronization period.
Here, we consider a more general weighted averaging scheme, where the updated estimate $Q_t^k$ is:
\begin{align} \label{eq:asyncq-periodic}
\forall (s,a) \in \cS \times \cA: \qquad  Q_{t}^k(s,a) =
  \begin{cases}
     \sum_{k=1}^K \alpha_t^k(s,a) Q_{t-\frac{1}{2}}^k(s,a) &~\text{if}~ t \equiv 0 \; (\text{mod}~ \tau)\\
    Q_{t-\frac{1}{2}}^k(s,a) &~\text{otherwise}
  \end{cases},
\end{align}
where $\alpha_t^k =[\alpha_t^k(s,a)]_{s\in \cS, a\in \cA} \in [0,1]^{|\cS||\cA|}$ is an entry-wise weight assigned to agent $k$ such that 
$$\forall (s,a) \in \cS \times \cA: \qquad \sum_{k=1}^K \alpha_t^k(s,a) = 1.$$ 
\end{enumerate}

After a total of $T$ iterations, \fedqavg outputs a global Q-estimate 
$Q_T(s,a) =\sum_{k=1}^K \alpha_T^k(s,a) Q_{T}^k(s,a)$ for  all $(s,a) \in \cS \times \cA$. In the subsections below, we provide two possible ways (equal and importance weighting) to choose $\alpha_t^k$ and their corresponding sample complexity analyses.
 
 \begin{algorithm}[t]
  \begin{algorithmic}[1] 
    \STATE \textbf{inputs:} learning rate $\{\eta\}$, discount factor $\gamma$, number of agents $K$, synchronization period $\tau$, total number of iterations $T$. 
    \STATE  \textbf{initialization:} $Q_0^k = Q_0$ for all $k \in [K]$. 
    \FOR{$t=1,\cdots,T$}
    \FOR{$k \in [K]$}    
    \STATE{ Draw action $a_{t-1}^k \sim \pi_{\mathsf{b}}^k(s_{t-1}^k)$, observe reward $r_{t-1}^k = r(s_{t-1}^k,a_{t-1}^k)$, and draw next state $s_{t}^k\sim P(\cdot\,|\,s_{t-1}^k,a_{t-1}^k)$.  }
    \STATE{ Compute $Q_{t-\frac{1}{2}}^k$ according to \eqref{eq:asyncq-local}.}
    \STATE{ Compute $Q_t^k$ according to \eqref{eq:asyncq-periodic}.}
    \ENDFOR
    \ENDFOR    
    \STATE \textbf{return:} $Q_T(s,a) =\sum_{k=1}^K \alpha_T^k(s,a) Q_{T}^k(s,a)$, for all $(s,a) \in \cS \times \cA$. 
  \end{algorithmic} 
  \caption{Federated Asynchronous Q-learning (\fedqavg)}
  \label{alg:dist-asyncq-local}
\end{algorithm}

\subsection{Performance guarantees with equal averaging} \label{sec:asyncq-analysis}

We begin with the most natural choice, which equally weights the local Q-estimates, that is,
\begin{align}  \label{eq:asyncq-weight-equal} 
\alpha_t^k(s,a) &= \frac{1}{K}.
\end{align}
We call the resulting scheme \fedqvavg, which is also analyzed in \citet{khodadadian22fedrlspeedup}.  We have the following improved performance guarantee in the next theorem.

\begin{theorem}[Finite-time convergence of \fedqvavg] \label{thm:dist-asyncq-localupdate-tighter}
 Consider any given $\delta \in (0,1)$ and $\varepsilon\in (0, \frac{1}{1-\gamma}]$.
 Suppose that the initialization of \fedqvavg satisfies $0 \leq Q_0  \le \frac{1}{1-\gamma}$.
There exist some sufficiently large constant $c_T>0$ and sufficiently small constant $c_\eta>0$, such that with probability at least $1-\delta$, the output of \fedqvavg satisfies $\| Q_T - Q^{\star}\|_{\infty}\leq \varepsilon$, provided that the synchronization period $\tau$, the sample size per agent $T$, and the learning rate $\eta$ satisfy
  \begin{subequations}\label{eq:guarantee_asyncvq-tighter}
  \begin{align} 
  \tau_0 \le \tau & \le \frac{1}{4 \eta} \min \left\{  \frac{1-\gamma}{4}, \, \frac{1}{K} \right\} ,\label{eq:asyncvq_period-tighter}  \\
  T
  &\ge  c_T  
  \left ( 
  \frac{\Csim }{ K \muminavg (1-\gamma)^5 \varepsilon^2} +T_0 \right)
  (\log((1-\gamma)^2\varepsilon))^2 \log{(TK)}\log{\frac{|\cS||\cA| {T^2 K}}{\delta}}, \\
 \eta & = c_{\eta}        \min\left \{
 \frac{K (1-\gamma)^4 \varepsilon^2}{\Csim } , \eta_0
 \right\}
 \frac{1}{ \log{(TK)}\log{\frac{|\cS||\cA| {T^2 K}}{\delta}} } , \label{eq:lr_asyncvq-tighter}
\end{align}
\end{subequations} 
 where $\tau_0= \frac{2176\tmaxmix}{\muminavg} \log{8K} \log{\frac{4|\cS||\cA|T^2}{\delta}}$, $T_0 =  \frac{1}{\muminavg (1-\gamma)\eta_0} $, and $\eta_0 = \frac{\muminavg  \min\{1-\gamma,K^{-1} \}}{\tmaxmix}$, independent of $\varepsilon$.
\end{theorem}
Theorem~\ref{thm:dist-asyncq-localupdate-tighter} implies that to achieve an $\varepsilon$-accurate estimate (in the $\ell_\infty$ sense), the sample complexity per agent of  \fedqvavg  is no more than 
\begin{align*}
\widetilde{O} \left ( \frac{ \Csim }{K \muminavg (1-\gamma)^5 \varepsilon^2} \right)
\end{align*}
for sufficiently small $\varepsilon$, when the burn-in cost $T_0$ --- representing the impact of the mixing times --- is amortized over time. A few implications are in order.
\paragraph{Linear speedup without full coverage.} 
The sample complexity of \fedqvavg shows linear speedup with respect to the number of agents, which is especially pronounced when the local behavior policies are similar, i.e., $\Csim\approx 1$. 
Notably, the guarantee holds as long as all agents collectively cover the entire state-action space (i.e., $\muminavg>0$), unveiling the benefit of heterogeneity in local behavior policies. This is surprising in view of the convergence guarantee provided in \citet{khodadadian22fedrlspeedup}, which requires each agent  visits the entire state-action space (i.e., $\mumin = 0$).
Moreover, our sample complexity has sharpened dependency on nearly all problem-dependent parameters compared to the bound $\widetilde{O} \Big(
  \frac{|\cS|^2}{K \mumin^5 (1-\gamma)^9 \varepsilon^2} \Big)$ obtained in \citet{khodadadian22fedrlspeedup} by at least a factor of 
  $$\frac{\muminavg |\cS|^2}{\Csim \mumin^5(1-\gamma)^4}\geq \frac{|\cS|^5|\cA|^3}{(1-\gamma)^4}. $$
For $K=1$, the bound nearly matches with the sharpest upper bound $\widetilde{O} \Big(\frac{1}{ \mumin  (1-\gamma)^4 \varepsilon^2} \Big)$ for the single-agent case \citep{li2021syncq} up to a factor of $1/(1-\gamma)$, when ignoring the burn-in cost. 
%

\paragraph{Communication efficiency.}
To provide further insights on the communication complexity of \fedqvavg, consider the regime when $\varepsilon$ is sufficiently small and the number of agents is sufficiently large such that $K\gtrsim \frac{1}{1-\gamma}$. By plugging the choice of the learning rate \eqref{eq:lr_asyncvq-tighter} into the upper bound of $\tau$ in \eqref{eq:asyncvq_period-tighter}, we can select the synchronization period as large as
$\tau \asymp \frac{\Csim}{ K^2 (1-\gamma)^4 \varepsilon^2 }$
up to logarithmic factors, which ensures the communication complexity $C_{\mathsf{round}} = T/\tau$ is no more than $\widetilde{O}\Big(\frac{K}{\muminavg(1-\gamma)}\Big)$.

\subsection{Performance guarantees with importance averaging}
 
In the asynchronous setting, heterogeneous behavior policies induce local trajectories that cover the state-action space in a non-uniform manner. As a result, agents may update the Q-estimate for a state-action pair at different frequencies, resulting in noisier Q-estimates of state-action pairs that an agent rarely visits.  
Equally-weighted averaging of such local Q-estimates is not efficient, because the convergence speed to the optimal Q-function for each state-action pair is bottlenecked with the slowest converging agent that visits it least frequently. This is highlighted by the impact of the heterogeneity factor $\Csim$ in the sample complexity of \fedqvavg, which scales linearly with $\Csim$, implying that increased heterogeneity among agents' trajectories may impede the convergence. For example, if only one agent exclusively visits a certain state-action pair $(s, a)$ with probability one, while other agents never visit that particular state-action pair, the heterogeneity factor becomes $\Csim=K$ when $K\leq 1/\muminavg$, canceling out the linear speedup.

Our key idea to prevent such inefficiency is to increase the contribution of frequently updated local Q-estimates, which are likely to have smaller errors. By assigning a weight inversely proportional to the error of the corresponding local estimate, we can balance the heterogeneous training progress of the local estimates and obtain an average estimate with much lower error. Combining this idea with the property that the local error decreases exponentially with the number of local visits, we propose an importance averaging scheme \fedqwavg with weights given by
\begin{align}   \label{eq:asyncq-weight-skewed}
 \alpha_t^k(s,a) &= \frac{(1-\eta)^{-N_{t-\tau, t}^k(s,a)}}{\sum_{k'=1}^K  (1-\eta)^{-N_{t-\tau, t}^{k'}(s,a)} }
\end{align}
for all $(s,a) \in \cS \times \cA$ and $k \in [K]$, where $N_{t-\tau, t}^k(s,a)$ represents the number of iterations between $[t-\tau, t)$ when the agent $k$ visits $(s,a)$.
The weights in \eqref{eq:asyncq-weight-skewed} can be calculated at the server based on the number of visits to each state-action pair by the agents in one synchronization period. Therefore, each agent needs to send its $N_{t-\tau, t}^k(s,a)$ for each $(s,a)$ along with its local Q-estimate, and \fedqwavg incurs twice the communication cost of \fedqvavg per iteration.

 We have the following theorem on the finite-time convergence of \fedqwavg.
 
\begin{theorem}[Finite-time convergence of \fedqwavg] \label{thm:dist-asyncwq-localupdate}
 Consider any given $\delta \in (0,1)$ and $\varepsilon\in (0, \frac{1}{1-\gamma}]$.
 Suppose that the initialization of \fedqwavg satisfies $0 \leq Q_0  \le \frac{1}{1-\gamma}$,    
and the synchronization period $\tau$ obeys
  \begin{subequations}
\begin{align}\label{eq:asyncwq_period} 
 \tau \le \frac{1}{4 \eta} \min \left\{  \frac{1-\gamma}{4}, \, \frac{1}{K} \right\} .
 \end{align}
There exist some sufficiently large constant $c_T>0$ and sufficiently small constant $c_\eta>0$, such that with probability at least $1-\delta$, the output of \fedqwavg satisfies $\| Q_T - Q^{\star}\|_{\infty}\leq \varepsilon$, provided that the sample size per agent $T$ and the learning rate $\eta$ satisfy
  \begin{align} \label{eq:guarantee_asyncwq}
  T
  &\ge  c_T \left ( 
  \frac{1}{ K \muminavg (1-\gamma)^5 \varepsilon^2} + \widetilde{T}_0 \right) 
  (\log((1-\gamma)^2\varepsilon))^2 \log{(TK)}\log{\frac{|\cS||\cA| {T^2 K}}{\delta}} , \\
 \eta & = c_{\eta}        \min\left \{
 K (1-\gamma)^4 \varepsilon^2 ,
  \widetilde{\eta}_0
 \right\}
 \frac{1}{\log{(TK)}\log{\frac{|\cS||\cA| {T^2 K}}{\delta}}}, \label{eq:lr_asyncwq}
\end{align}
\end{subequations}
where $\widetilde{T}_0 = \frac{ 1}{\muminavg(1-\gamma) \eta_0} $ and $\widetilde{\eta}_0 = \min \left \{ \frac{1}{\tmaxmix },  1-\gamma,  K^{-1} \right \}$, independent of $\varepsilon$.
\end{theorem}

Theorem~\ref{thm:dist-asyncwq-localupdate} implies that to achieve an $\varepsilon$-accurate estimate (in the $\ell_\infty$ sense), the sample complexity per agent of  \fedqwavg  is no more than 
\begin{align*}
\widetilde{O} \left ( \frac{1}{K \muminavg (1-\gamma)^5 \varepsilon^2} \right)
\end{align*}
for sufficiently small $\varepsilon$, when the burn-in cost $\widetilde{T}_0$ --- representing the impact of the mixing times --- is amortized over time. A few implications are in order.

\paragraph{Linear speedup without the curse of heterogeneity.}
The sample complexity of \fedqwavg is better than that of \fedqvavg, since it no longer depends on $\Csim$ which can be as large as $1/\muminavg$.
\fedqwavg  not only overcomes potential insufficient local coverage by exploiting the complementary coverage of agents' behavior policies, but also achieves linear speedup  with respect to the number of agents without suffering from the potential performance degradation due to the associated statistical heterogeneity as in \fedqvavg. In fact, the performance of \fedqwavg matches with centralized Q-learning as if we collect and process all data trajectories at the central server, up to the burn-in cost and logarithmic factors.

\paragraph{Communication efficiency.}
To provide further insights on the communication complexity of \fedqwavg, consider again the regime when $\varepsilon$ is sufficiently small and  $K\gtrsim \frac{1}{1-\gamma}$.
To minimize the communication frequency while preserving the sample efficiency, we again plug the choice of the learning rate \eqref{eq:lr_asyncwq} into \eqref{eq:asyncwq_period} and select the synchronization period as large as
$ \tau \asymp \frac{1}{ K^2 (1-\gamma)^4 \varepsilon^2 }$
up to logarithmic factors. Then, this ensures the communication complexity $C_{\mathsf{round}} = T/\tau$ is no more than $\widetilde{O}\Big(\frac{K}{\muminavg(1-\gamma)}\Big)$.

%% file: experiments.tex
\section{Numerical experiments}
\label{sec:exp}

In this section, we conduct numerical experiments to  demonstrate the performance of the asynchronous Q-learning algorithms (\fedqvavg and \fedqwavg).

\begin{figure}[ht]
\centering
\includegraphics[width=0.6\textwidth]{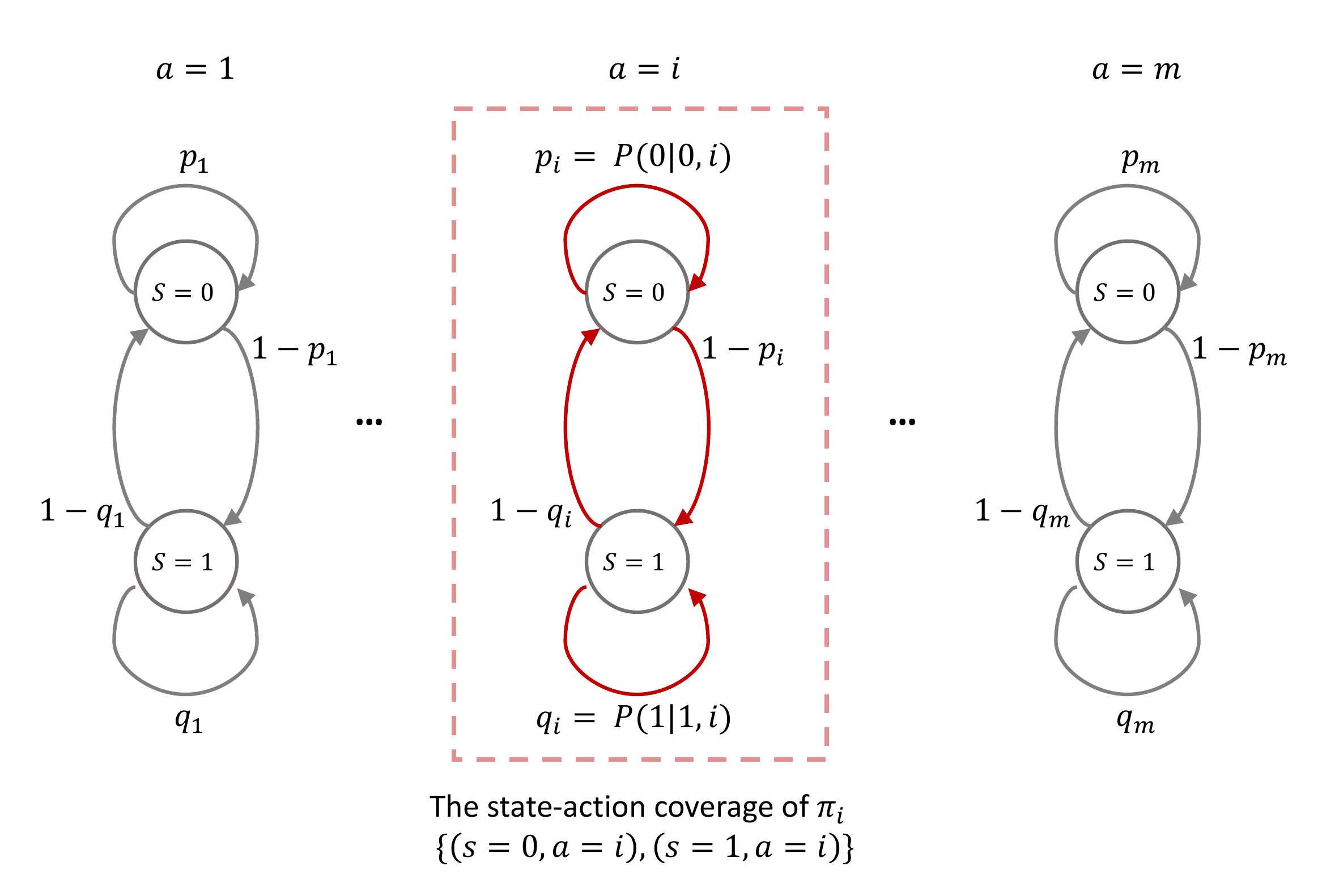}
\caption{An illustration of the constructed synthetic MDP $\cM$. The red arrows represent transitioning paths when action $a=i$ is taken in $s=0$ and $s=1$. A trajectory induced by $\pi_i$, which executes only action $i$ for any state, can cover only two state-action pairs, $(s=0,a=i)$ and $(s=1, a=i)$.}
\label{fig:mdp-diagram}
\end{figure}

\paragraph{Experimental setup.}
Consider an MDP $\cM = (\cS, \cA, P, r, \gamma)$ described in Figure~\ref{fig:mdp-diagram}, where $\cS = \{0, 1\}$ and $\cA = \{1, 2, \cdots, m \}$. The reward function $r$ is set as $r(s=1, a)=1$ and $r(s=0, a)=0$ for any action $a \in \cA$, and the discount factor is set as $\gamma = 0.9$. We now describe the transition kernel $P$. Here, we set the self-transitioning probabilities $p_a \defn P(0|0,a) $ and $q_a \defn P(1|1,a)$ uniformly at random from $[0.4, 0.6]$ for each $a \in \cA$, and set the probability of transitioning to the other state as $P(1-s|s,a) = 1- P(s|s,a)$ for each $s \in \cS$. 

We evaluate the proposed federated asynchronous Q-learning algorithms on the above MDP with $K$ agents selecting their behavior policies from $\Pi = \{\pi_1, ~ \pi_2, \cdots ,~\pi_{m}\}$, where the $i$-th policy always chooses action $i$ for any state, i.e., $\pi_i(i|s) = 1$ for all $s \in \cS$. Here, we assign  $\pi_{i}$ to agent $k \in [K]$ if $i \equiv k~(\text{mod} ~m)$.
Note that if an agent has a behavior policy $\pi_i$, it can visit only two state-action pairs, $(s=0,a=i)$ and $(s=1,a=i)$, as described in Figure~\ref{fig:mdp-diagram}.
Thus, each agent covers a subset of the state-action space, and at least $K=m$ agents are required to obtain local trajectories collectively covering the entire state-action space. Under this setting with $m=20$, we run the algorithms for $100$ simulations using samples randomly generated from the MDP and policies assigned to the agents. The Q-function is initialized with entries uniformly at random from $(0, \frac{1}{1-\gamma}]$ for each state-action pair.

\begin{figure}[htb]
\centering
\includegraphics[width=0.5\textwidth]{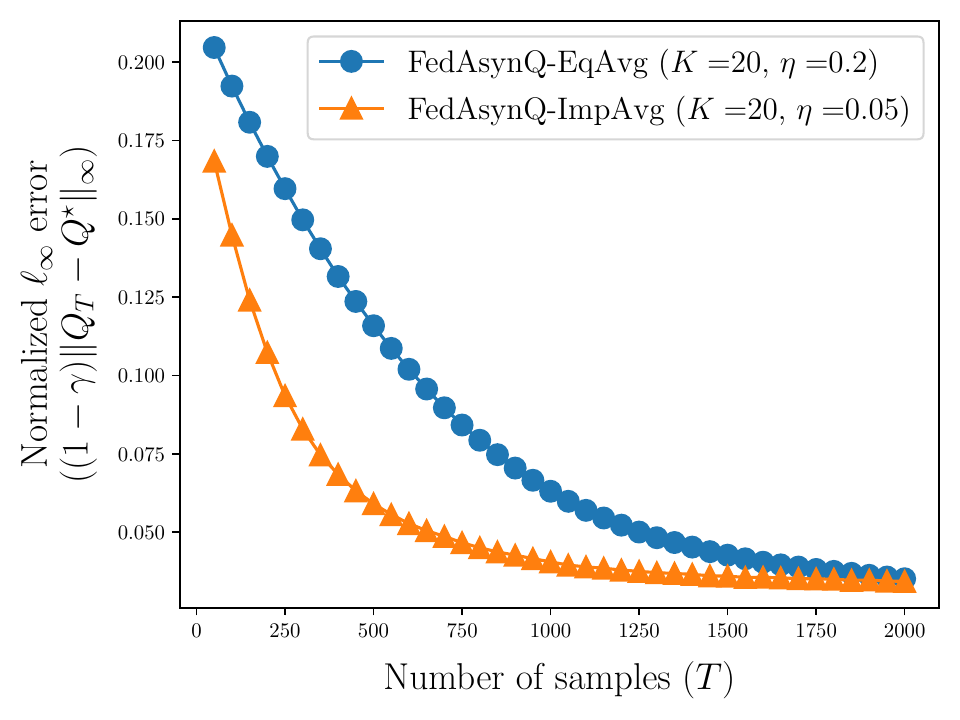}
\caption{The normalized $\ell_\infty$ error of the Q-estimates $(1-\gamma)\|Q_T-Q^{\star}\|_{\infty}$ with respect to the number of samples $T$ for both \fedqvavg and \fedqwavg, with $K=20$ and $\tau = 50$. Here, the learning rates of \fedqwavg and \fedqvavg are set as $\eta=0.05$ and $\eta = 0.2$, where each algorithm converges to the same error floor at the fastest speed, respectively.  }
\label{fig:fedq-errconv}
\end{figure}

\paragraph{Faster convergence of \fedqwavg.} 
Figure~\ref{fig:fedq-errconv} shows the normalized Q-estimate error $(1-\gamma)\|Q_T-Q^{\star}\|_{\infty}$ with respect to the sample size $T$, with $K=20$ and $\tau=50$.
Given the trajectories of agents collectively cover the entire state-action space, the global Q-estimates of both \fedqvavg and \fedqwavg converge to the optimal Q-function, yet at different speeds. 
Although \fedqvavg converges in the end, we can see that it converges much slower compared to \fedqwavg, because each entry of the Q-function is trained by only one agent while the other $m-1$ agents never contribute useful information. However, the vacuous values of the $m-1$ agents significantly slow down the global convergence under equal averaging.

\begin{figure}[htb]
  \centering
  \includegraphics[width=0.5\linewidth]{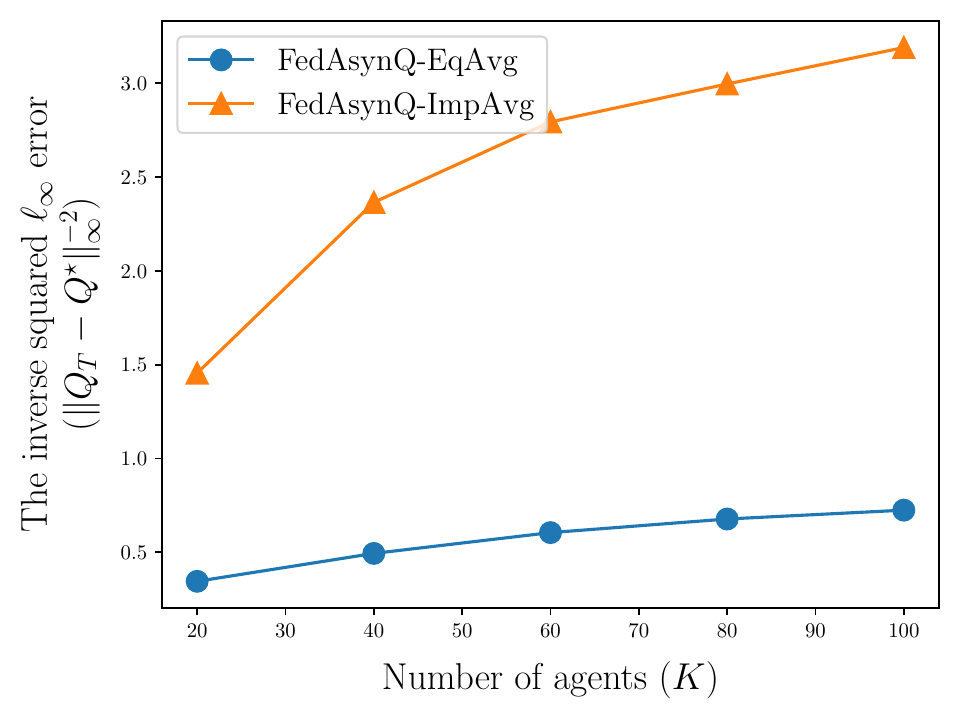}
\caption{The inverse squared $\ell_\infty$ error $\|Q_T-Q^{\star}\|_{\infty}^{-2}$ with 
respect to the number of agents $K=20,40,60,80,100$ for both \fedqvavg and \fedqwavg, with $T=300$ and $\tau=50$. }  \label{fig:fedq-K}
\end{figure}

\paragraph{Convergence speedup.}
Figure~\ref{fig:fedq-K} demonstrates the impact of the number of agents on the convergence speed of \fedqvavg and \fedqwavg. It can be observed that there is indeed a speedup in terms of the number of agents $K$ with respect to the squared $\ell_\infty$ error $\|Q_T-Q^{\star}\|_{\infty}^{-2}$, which is poised to scale linearly with respect to the number of agents.
In particular, the speedup is more rapid with \fedqwavg as $K$ increases, while it increases much slower with \fedqvavg. This shows that \fedqwavg achieves much better convergence speedup in terms of the number of agents.

\begin{figure}[htb]
  \centering
  \includegraphics[width=0.5\linewidth]{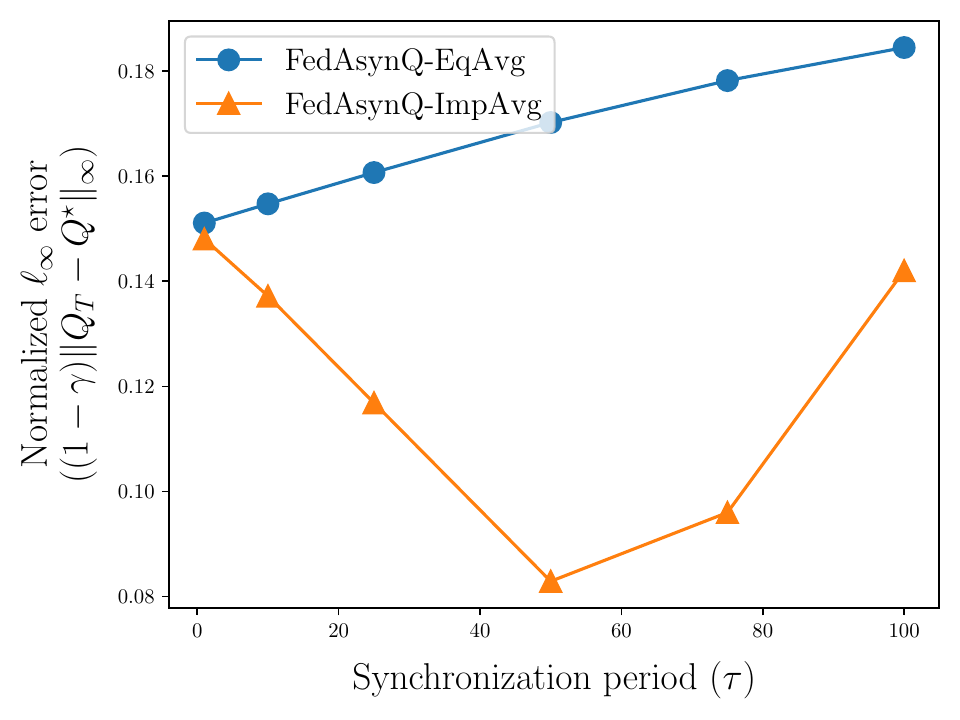}
\caption{The normalized $\ell_\infty$ error of the Q-estimates $(1-\gamma)\|Q_T-Q^{\star}\|_{\infty}$ with respect to the synchronization period $\tau =1, 10, 25, 50, 75, 100$ for both \fedqvavg and \fedqwavg, with $K=20$ and $T=300$.}   \label{fig:fedq-tau}
\end{figure}

\paragraph{Communication efficiency.} Figure~\ref{fig:fedq-tau} demonstrates the impact of the synchronization period $\tau$ on the convergence of \fedqwavg and \fedqvavg. 
With frequent averaging ($\tau=1$), \fedqwavg slightly outperforms \fedqvavg, but there is no significant difference because the heterogeneity between local Q-functions after just one local update is very small. The performance of \fedqvavg degrades as we increase $\tau$ since \fedqvavg cannot cope with the increased heterogeneity between local Q-estimates as we increase the number of local steps. On the other end, the performance of \fedqwavg improves first (i.e., $\tau=10,~25,~50$) as it balances the heterogeneity much better than \fedqvavg, but drops later if $\tau$ is too large (i.e., $\tau=75,~100$) due to the high variance of the averaged Q-estimates.

%% file: analysis.tex
\section{Analysis outline}
\label{sec:analysis}

Let the matrix $P \in \mathbb{R}^{|\cS||\cA| \times |\cA|}$ represent the transition kernel of the underlying MDP, where $P(s,a) = P(\cdot|s,a)$ is the probability vector corresponding to the state transition at the state-action pair $(s,a)$. 
For any vector $V \in \mathbb{R}^{|\mathcal{S}|}$, we define the variance parameter $\Var_{s,a}(V)$ with respect to the probability vector $P(s,a)$ as
\begin{equation}\label{eq:def_var}
\Var_{s,a}(V) \defn \mathbb{E}_{s'\sim P(\cdot|s,a)}\big[V(s') - P(s,a)V \big]^2 = P(s,a)(V\circ V)-[ P(s,a) V] \circ[P(s,a)V].
\end{equation}
Here, $\circ$ denotes the Hadamard product such that $a \circ b = [a_i b_i]_{i=1}^n$ for any vector $a = [a_i]_{i=1}^n, b = [b_i]_{i=1}^n \in \mathbb{R}^n$. With slight abuse of notation, we shall also assume $V^{\star}\in\mathbb{R}^{|\mathcal{S}|}$, $V_{t}^k\in\mathbb{R}^{|\mathcal{S}|}$, $Q^{\star}\in\mathbb{R}^{|\mathcal{S}||\mathcal{A}|}$, $Q_{t}^k\in\mathbb{R}^{|\mathcal{S}||\cA|}$, $Q_{t+\frac{1}{2}}^k\in\mathbb{R}^{|\mathcal{S}||\cA|}$ and
$r \in\mathbb{R}^{|\mathcal{S}||\mathcal{A}|}$ represent the corresponding functions in the matrix/vector form.

\subsection{Basic facts}
We first state a few basic facts that hold both for the synchronous and the asynchronous settings. It is easy to establish, by induction, that all iterates satisfy for all $1\leq k \leq K$ and $t\geq 0$ that
\begin{equation}\label{eq:bounded_iterates}
0 \leq Q_t^k \leq \frac{1}{1-\gamma}, \qquad 0 \leq V_t^k \leq \frac{1}{1-\gamma},
\end{equation}
as long as $0\leq Q_0 =Q_0^k \leq \frac{1}{1-\gamma}$; see a similar argument, e.g., in \citet[Lemma 4]{li2021syncq}. In addition, observe that
\begin{equation} \label{eq:vgaptoqgap}
\|  V_t^k - V^{\star} \|_{\infty} \le \| Q_{t}^k - Q^{\star} \|_{\infty}
 \end{equation}
 since
 $$\| V_t^k - V^{\star}\|_{\infty}   = \max_{s \in \cS} \Big|  \max_{a\in\cA} Q_t^k(s,a) - \max_{a\in\cA} Q^{\star}(s,a ) \Big| \leq  \max_{s \in \cS, a\in\cA} \big|Q_t^k(s,a) - Q^{\star}(s,a ) \big| \leq \| Q_{t}^k - Q^{\star}\|_{\infty}  .$$
    
 Letting $Q_t$ be the average of the local Q-estimates at the end of the $t$-th iteration, i.e.,
$\Qavg_t = \frac{1}{K} \sum_{k=1}^K Q_t^k$, it follows from \eqref{eq:syncq-periodic} and \eqref{eq:asyncq-periodic} that for all $t\geq 0$ that
\begin{equation} \label{eq:ave_preserve} 
\Qavg_{t} =   \frac{1}{K} \sum_{k=1}^K Q_{t}^k = \frac{1}{K} \sum_{k=1}^K Q_{t-\frac{1}{2}}^k . 
\end{equation}
Denote the error between $\Qavg_t$ and $\Qopt$ by
$$\erravg_t  = \Qopt - \Qavg_t,$$
which is the quantity we aim to control. From \eqref{eq:bounded_iterates}, it holds immediately that for all $t\geq 0$, 
\begin{equation}\label{eq:crude}
\| \erravg_t \|_{\infty} \leq \frac{1}{1-\gamma}.
\end{equation}

Next, we also introduce the following functions pertaining to periodic averaging. For any $t$, 
\begin{itemize}
\item define $\syn(t): =\tau \lfloor \frac{t}{\tau} \rfloor $ as the most recent synchronization step until $t$;
\item define $\nsyn(t):=\lfloor \frac{t}{\tau} \rfloor$ as the number of synchronization steps until $t$.
\end{itemize}

\input{fedsyncq_analysis}
 \input{fedasyncq_analysis_eqimproved}

 \input{fedasyncq_analysis_impavg}

%% file: fedsyncq_analysis.tex
\subsection{Proof outline of Theorem~\ref{thm:dist-syncq-local}}
 
  Define the local empirical transition matrix at the $t$-th iteration $P_t^k \in \{0,1\}^{|\cS||\cA| \times |\cS|}$ as
  \begin{align} 
    P_t^k((s,a), s') \defn
    \begin{cases}
      1, \quad \text{if} ~ s' = s_t^k(s,a) \\
      0, \quad \text{otherwise}
    \end{cases},
  \end{align}
 then the local update rule \eqref{eq:syncq-local} can be rewritten as
\begin{align} \label{eq:syncq-local-vec}
   Q_{t-\frac{1}{2}}^k &= (1-\eta) Q_{t-1}^k + \eta  \left( r + \gamma   P_t^k  V_{t-1}^k \right).  
\end{align}

The proof of Theorem~\ref{thm:dist-syncq-local} consists of the following steps.
\paragraph{Step 1: error decomposition.} 
To analyze the error $\erravg_t,$ we first decompose the error into three terms, each of which can be bounded in a simple form. From \eqref{eq:ave_preserve}, it follows that   
\begin{align}
\erravg_{t} =  \frac{1}{K} \sum_{k=1}^K \big( Q^{\star}  -  Q_{t-\frac{1}{2}}^k \big)   
  &\overset{\mathrm{(i)}}{=} \frac{1}{K} \sum_{k=1}^K \left( (1-\eta) (Q^{\star}  - Q_{t-1}^k  ) + \eta  (Q^{\star}  - r - \gamma P_{t}^k V^k_{t-1} ) \right) \cr
  & \overset{\mathrm{(ii)}}{=} (1-\eta ) \erravg_{t-1} + \eta  \frac{\gamma}{K} \sum_{k=1}^K \big( PV^{\star}  -P_{t}^k V^k_{t-1} \big)  \cr
    & = (1-\eta ) \erravg_{t-1} + \eta   \frac{\gamma}{K} \sum_{k=1}^K \big( P    - P_{t}^k \big)  V^k_{t-1} + \eta  \frac{\gamma}{K} \sum_{k=1}^K P \big( V^{\star} -  V^k_{t-1} \big)  , \nonumber
\end{align}
where (i) follows from \eqref{eq:syncq-local-vec}, and (ii) follows from Bellman's optimality equation $Q^{\star}  = r + \gamma  P V^{\star}$. By recursion over the above relation, we obtain
\begin{align} \label{eq:dist-syncq-error-decomp}
  \erravg_t  
  &=   \underbrace{ (1-\eta)^t \erravg_0 \vphantom{\frac{1}{K} \sum_{k=1}^K ( P - P_i^k  )} }_{=: E_t^1}  +  \underbrace{ \eta\frac{\gamma}{K}  \sum_{i=1}^{t}     (1-\eta)^{t-i} \sum_{k=1}^K ( P  - P_i^k  ) V^k_{i-1} }_{ =: E_t^2} +  \underbrace{ \eta \frac{\gamma}{K} \sum_{i=1}^{t}    (1-\eta)^{t-i}  \sum_{k=1}^K  P  (V^{\star} - V^k_{i-1}) }_{ =: E_t^3} . 
\end{align} 
Here, the first term $E_t^1$ denotes the initialization error stemming from the disparity between the initial Q-values and the optimal Q-values, which diminishes exponentially throughout iterations. The second term, $E_t^2$, comprises a weighted sum accounting for the difference between the true transition probability and the realized transition in each iteration, where the difference arises from the randomness of transitions. Lastly, the final term, $E_t^3$, represents a weighted sum of value estimation errors from preceding iterations, which introduces a recursive relation.

\paragraph{Step 2: bounding the error terms.}
Now, we obtain a bound of each of the error terms in \eqref{eq:dist-syncq-error-decomp} separately.

\begin{itemize}
\item  \textbf{Bounding $\| E_t^1\|_{\infty}$.} Using the fact that all agents start with the same initial Q-values, i.e., $Q_0^k= Q_0 $, the first error term is bounded as follows: 
\begin{align}
  \| E_t^1 \|_{\infty} = (1-\eta)^t
    \left\|   \erravg_0  \right\|_{\infty}  
  \le   \frac{ (1-\eta)^t}{1-\gamma},
\end{align}
where the last inequality follows from \eqref{eq:crude}.

\item \textbf{Bounding $\|E_t^2\|_{\infty}$.} 
Exploiting conditional independence across transitions in different iterations and applying Freedman's inequality \citep{freedman1975tail},
the second error term is bounded using Lemma~\ref{lemma:dist-syncq-error2-freedman} below, whose proof is provided in Appendix~\ref{proof:dist-syncq-error2-freedman}.  
\begin{lemma} \label{lemma:dist-syncq-error2-freedman}
  For any given $\delta  \in (0,1)$, the following holds
  \begin{align}
    \left\| E_t^2 \right\|_{\infty}
    \le \frac{8 \gamma}{1-\gamma}\sqrt{ \frac{\eta}{K} \log{\frac{|\cS||\cA|T}{\delta}}}
  \end{align}
  for all $0 \le t \le T$  with probability at least $1-\delta$, as long as $\eta$ satisfies $\eta \le \frac{K}{2} (\log{\frac{|\cS||\cA|T}{\delta}})^{-1}$. 
\end{lemma}

\item \textbf{Bounding $\| E_t^3\|_{\infty}$.} 
For $E_t^3$, we obtain the following recursive relation using Lemma~\ref{lemma:syncq-e3b} below, whose proof is provided in Appendix~\ref{proof:syncq-e3b}.
\begin{lemma} \label{lemma:syncq-e3b}
 Let $\beta$ be any integer that satisfies $0\leq \beta \leq \nsyn(T)$. For any given $\delta \in (0,1)$, the following holds   
\begin{align}
    \| E_t^3\|_{\infty}
   & \le \frac{2 \gamma}{1-\gamma} (1-\eta)^{\beta \tau} + \frac{16 \gamma \eta   \sqrt{\tau-1}}{(1-\gamma)} \sqrt{  \log{\frac{2|\cS| |\cA| K T}{\delta}}}  + \gamma (1+4 \eta (\tau-1)) \max_{\syn(t)-\beta \tau \le i <t} \| \erravg_{i} \|_{\infty} \nonumber
 \end{align}
for all $\beta \tau \le t \leq  T$ with probability at least $1-\delta$, as long as $\eta$ satisfies $\tau\eta<1/2$.  
\end{lemma}

\end{itemize}

\paragraph{Step 3: solving a recursive relation.}
By putting all the bounds derived in the previous step together, for any $\beta \tau \leq t \leq T$, the total error bound can be written in a simple recursive form as follows:
\begin{align} \label{eq:syncq-errorall}
  \left \| \erravg_{t} \right \|_{\infty}
  & \leq \zeta + \gamma (1+4  \eta(\tau-1)) \max_{\syn(t)-\beta \tau \le i <t} \| \erravg_{i} \|_{\infty}  
  \leq \zeta + \left(\frac{1+\gamma}{2}\right) \max_{\syn(t)-\beta \tau \le i <t} \| \erravg_{i} \|_{\infty},
\end{align}
where in the first inequality we introduce the short-hand notation
\begin{align} \label{eq:def_zeta}
  \zeta
  \defn
  \frac{4(1-\eta)^{\beta \tau}}{1-\gamma} 
  + \frac{8 \gamma}{1-\gamma}\sqrt{ \frac{\eta}{K} \log{\frac{|\cS||\cA|T}{\delta}}} 
  + \frac{16 \gamma \eta    \sqrt{\tau-1}}{(1-\gamma)} \sqrt{  \log{\frac{2|\cS| |\cA| K T}{\delta}}},
\end{align}
and the second inequality follows from the assumption
$  \tau-1 \le  \frac{1-\gamma}{8\gamma\eta}$.

By invoking the recursive relation in \eqref{eq:syncq-errorall} $L$ times, where the choices of $\beta$ and $L$ will be made momentarily, it follows that for any $L\beta\tau \leq t\leq T $,
\begin{align} \label{eq:syncq-errorall-recursive}
  \left \|\erravg_{t} \right \|_{\infty}
  &\le \sum_{i=0}^{L-1} \left( \frac{1+ \gamma}{2} \right)^i  \zeta  
     + \left( \frac{1+ \gamma}{2} \right)^L \max_{\syn(t)-L \beta \tau \le i<t} \| \erravg_i \|_{\infty} \cr
  &\le \frac{2}{1-\gamma}   \zeta + \left(\frac{1+\gamma}{2}\right)^L \left( \frac{1}{1-\gamma}\right),
\end{align}
where the second line uses the crude bound in \eqref{eq:crude}.

Setting $\beta = \left \lfloor \frac{1}{\tau} \sqrt{\frac{(1-\gamma)T}{2\eta }} \right \rfloor$ and $L = \left \lceil \sqrt{\frac{\eta T}{1-\gamma}} \right \rceil$, which ensures $L\beta\tau\leq T$, and plugging their choices into \eqref{eq:def_zeta} and \eqref{eq:syncq-errorall-recursive} at $t=T$, we obtain that
\begin{align}
  \|\erravg_T\|_{\infty}
 & \le \frac{8(1-\eta)^{\beta \tau}}{(1-\gamma)^2} + \frac{16 \gamma}{(1-\gamma)^2}\sqrt{ \frac{\eta}{K} \log{\frac{|\cS||\cA|T}{\delta}}} 
  + \frac{32 \gamma \eta  \sqrt{\tau-1}}{(1-\gamma)^2} \sqrt{  \log{\frac{2|\cS| |\cA| K T}{\delta}}} + \left(\frac{1+\gamma}{2}\right)^L \left( \frac{1}{1-\gamma}\right) \nonumber \\
&\leq   \frac{32}{(1-\gamma)^2}
  \left (
  \exp\left(-\frac{\sqrt{(1-\gamma)\eta T}}{2}\right)
  + \gamma\sqrt{ \frac{\eta}{K} \log{\frac{|\cS||\cA|T}{\delta}}} 
  + \gamma \eta   \sqrt{\tau-1} \sqrt{\log{\frac{|\cS||\cA|K T}{\delta}}}
  \right ) \nonumber \\
  & \leq  \frac{64}{(1-\gamma)^2}
  \left (
  \exp\left(-\frac{\sqrt{(1-\gamma)\eta T}}{2}\right)
  + \gamma\sqrt{ \frac{\eta}{K} \log{\frac{|\cS||\cA|KT}{\delta}}} 
  \right ) ,
\end{align}
where the second line follows from
\begin{align*}
(1-\eta)^{\beta \tau} 
  &\le  \exp( - \eta \beta\tau)\leq \exp\left(-\frac{\sqrt{(1-\gamma)\eta T}}{2}\right) ,\\
  \left(\frac{1+\gamma}{2}\right)^L &= \left(1-\frac{1-\gamma}{2}\right)^L \le \exp\left(-\frac{(1-\gamma)}{2}L \right) \le \exp\left(-\frac{\sqrt{(1-\gamma)\eta T}}{2}\right),
\end{align*}
 and the third line follows from the choice of the synchronization period such that
\begin{equation} 
\tau-1 \le \frac{1}{\eta} \min \left\{  \frac{1-\gamma}{8\gamma}, \, \frac{1}{K} \right\} .
\end{equation}

Thus, for any given $\varepsilon \in (0,\frac{1}{1-\gamma})$, we can guarantee that $\|\erravg_T\|_{\infty} \le \varepsilon$ if
\begin{align}
  T
  &\ge c_T \frac{1}{K(1-\gamma)^5 \varepsilon^2} (\log((1-\gamma)^2\varepsilon))^2 \log{\frac{|\cS||\cA|K T}{\delta}},  \nonumber \\
  \eta &= c_{\eta}        K (1-\gamma)^4 \varepsilon^2 \frac{1}{\log{\frac{|\cS||\cA|K T}{\delta}}}
 \end{align}
 for some sufficiently large $c_T$ and sufficiently small $c_\eta$.

%% file: fedasyncq_analysis_eqimproved.tex

\subsection{Proof outline of Theorem~\ref{thm:dist-asyncq-localupdate-tighter}}
\label{proof:dist-asyncq-localupdate-tighter}

For simplicity, we introduce the following notation.
Let $\cU_{v_1, v_2}^k(s,a)$ represent a set of iteration indices between $[v_1, v_2)$ for some $0 \le v_1 \le v_2 \le T$ where agent $k$ visits $(s,a)$, i.e.,
$$\cU_{v_1, v_2}^k(s,a) \defn \left\{ u \in [v_1, v_2) :\; (s_{u}^k, a_{u}^k) = (s,a) \right\},$$ 
and 
$N_{v_1, v_2}^k(s,a)$ denotes the number of visits of agent $k$ on $(s,a)$ during iterations between $[v_1, v_2)$, i.e., 
$$N_{v_1, v_2}^k(s,a) = |\cU_{v_1, v_2}^k(s,a)|.$$

Define the local empirical transition matrix at the $t$-th iteration $P_t^k \in \{0,1\}^{|\cS||\cA| \times |\cS|}$ as
\begin{align} 
  P_t^k((s,a), s') \defn
  \begin{cases}
    1 \quad \text{if} ~ (s,a,s') = (s_{t-1}^k, a_{t-1}^k, s_t^k) \\
    0 \quad \text{otherwise}
  \end{cases}.
\end{align}
Then the local update rule \eqref{eq:asyncq-local} can be rewritten as
\begin{align} \label{eq:asyncq-local-vector-tighter}
  Q_{t-\frac{1}{2}}^k(s, a) =
  \begin{cases}
    (1-\eta) Q_{t-1}^k(s,a) + \eta (r_{t-1}^k + \gamma P_t^k(s,a) V_{t-1}^k ) &\quad   \mbox{if~}(s,a) = (s_{t-1}^k,a_{t-1}^k) \cr 
    Q_{t-1}^k(s,a), &\quad \mbox{otherwise}
  \end{cases}.
\end{align}
 
The proof of Theorem~\ref{thm:dist-asyncq-localupdate-tighter} consists of the following steps.
\paragraph{Step 1: error decomposition.}
Consider any $0 \le t \le T$ such that $t \equiv 0 ~(\text{mod}~ \tau)$, i.e., $t$ is a synchronization step. 
To analyze $\Delta_t$, we first decompose the error for each $(s,a) \in \cS \times \cA$ as follows:
\begin{align} \label{eq:error_sync_point-tighter}
  \Delta_t(s,a)
  &= \frac{1}{K} \sum_{k=1}^K ( Q^{\star}(s,a) - Q_{t-\frac{1}{2}}^k(s,a)) \cr
  &= \left ( \frac{1}{K} \sum_{k=1}^K (1-\eta)^{N_{t-\tau, t}^k(s,a)} \right ) \Delta_{t-\tau}(s,a) \cr
  &\qquad + \frac{\gamma}{K} \sum_{k=1}^K \sum_{u \in \cU_{t-\tau, t}^k(s,a)} \eta (1-\eta)^{N_{u+1,t}^k(s,a)}  (P(s,a) -  P_{u+1}^k(s,a) ) V^k_{u} \cr
  &\qquad + \frac{\gamma}{K} \sum_{k=1}^K \sum_{u \in \cU_{t-\tau, t}^k(s,a) } \eta (1-\eta)^{N_{u+1,t}^k(s,a)}  P(s,a) (V^{\star} - V^k_{u}),
\end{align}
where we invoke the following recursive relation of the local error at iteration $u$ such that $(s_{u-1}, a_{u-1}) = (s,a)$:
\begin{align} \label{eq:asynq-local-onestepdecomp-tighter}
  &Q^{\star}(s,a) - Q_{u-\frac{1}{2}}^k(s,a) \cr
  &=  (1-\eta) (Q^{\star}(s,a)- Q_{u-1}^k(s,a)) + \eta (Q^{\star}(s,a) - r_{u-1}^k - \gamma P_u^k(s,a) V_{u-1}^k) \cr
  &= (1-\eta) (Q^{\star}(s,a)- Q_{u-1}^k(s,a)) + \eta (\gamma P(s,a) V^{\star} - \gamma P_u^k(s,a) V_{u-1}^k) \cr
  &= (1-\eta) (Q^{\star}(s,a)- Q_{u-1}^k(s,a)) + \gamma \eta (P(s,a) - P_u^k(s,a)) V_{u-1}^k + \gamma P(s,a) (V^{\star} - V_{u-1}^k).
\end{align}
Here, the second equality follows from Bellman's optimality equation. Denoting 
\begin{equation}\label{eq:def_lambda_vvv-tighter}
\lambda_{v_1,v_2}(s,a) \defn \frac{1}{K}\sum_{k=1}^K (1-\eta)^{N_{v_1, v_2}^k(s,a)}
\end{equation}
for any integer $0 \le v_1 \le v_2 \le T$, we apply recursion to the relation \eqref{eq:error_sync_point-tighter} over the synchronization periods, and obtain  
\begin{align} \label{eq:asyncvq-error-decomp-tighter}
&  \Delta_t(s,a) \nonumber \\
  &= \left( \prod_{h=0}^{\nsyn(t) -1} \lambda_{h\tau, (h+1)\tau}(s,a) \right)  \Delta_{0}(s,a) \cr
  &\quad +  \sum_{h=0}^{\nsyn(t) -1} \left ( \prod_{l = (h+1)}^{\nsyn(t)-1} \lambda_{l\tau, (l+1)\tau}(s,a) \right ) \frac{\gamma}{K}\sum_{k=1}^K \sum_{u \in \cU_{h \tau, (h+1)\tau}^k(s,a)} \eta (1-\eta)^{N_{u+1,(h+1)\tau}^k(s,a)}  (P(s,a) -  P_{u+1}^k(s,a) ) V^k_{u} \cr
  &\quad +  \sum_{h=0}^{\nsyn(t) -1}\left ( \prod_{l = (h+1)}^{\nsyn(t)-1} \lambda_{l\tau, (l+1)\tau}(s,a) \right )  \frac{\gamma}{K} \sum_{k=1}^K \sum_{u \in \cU_{h \tau, (h+1)\tau}^k(s,a) } \eta (1-\eta)^{N_{u+1,(h+1)\tau}^k(s,a)}  P(s,a) (V^{\star} - V^k_{u}) \cr
  &= \underbrace{ \omega_{0,t}(s,a) \Delta_{0}(s,a) \vphantom{ \sum_{k=1}^K \sum_{u \in \cU_{0, t}^k(s,a)}}}_{=: E_t^1(s,a)} 
+ \underbrace{ \gamma \sum_{k=1}^K \sum_{u \in \cU_{0, t}^k(s,a)} \omega_{u,t}^k(s,a) (P(s,a) -  P_{u+1}^k(s,a) ) V^k_{u} }_{=: E_t^2(s,a)} \cr
  &\qquad + \underbrace{ \gamma \sum_{k=1}^K \sum_{u \in \cU_{0, t}^k(s,a)} \omega_{u,t}^k(s,a)  P(s,a) (V^{\star} - V^k_{u}) }_{=: E_t^3(s,a)},
\end{align}
which is decomposed in a similar manner as \eqref{eq:dist-syncq-error-decomp}. Here, we define
\begin{subequations}\label{eq:asyncq-weight-def-tighter}
\begin{align} 
  \omega_{0,t}(s,a) &\defn \prod_{h = 0}^{\nsyn(t)-1} \lambda_{h\tau, (h+1)\tau}(s,a) ,   \\
  \omega_{u,t}^k(s,a) &\defn \frac{1}{K}   \eta (1-\eta)^{N_{u+1,(\nsyn(u)+1)\tau}^k(s,a)}   \prod_{l = \nsyn(u)+1}^{\nsyn(t)-1} \lambda_{l\tau, (l+1)\tau}(s,a).
\end{align}
\end{subequations}

We record the following useful lemma whose proof is provided in Appendix~\ref{proof:asyncq-wsum-tighter}.
\begin{lemma} \label{lemma:asyncq-wsum-tighter}
  Consider integers $v_1$ and $v_2$ such that $0 \le v_1 \le v_2  \le t \le T$, where $t \equiv 0 ~(\text{mod}~ \tau)$, and a state-action pair $(s,a) \in \cS \times \cA$. Suppose that $ \eta \tau \le 1$. The parameters defined in \eqref{eq:asyncq-weight-def-tighter} satisfy
  \begin{subequations}
  \begin{align}
    \label{eq:asyncq-lambda-tighter}  \lambda_{v_1,v_2}(s,a) & \le \exp\left (-\frac{\eta}{2K} \sum_{k=1}^K N_{v_1,v_2}^k(s,a)\right), \\
    \label{eq:asyncq-wsum-1-tighter}  \omega_{0,t}(s,a) + \sum_{k=1}^K \sum_{u \in \cU_{0, t}^k(s,a)} \omega_{u,t}^k (s,a) & = 1,\\
    \label{eq:asyncq-wsum-1-part-tighter} \sum_{k=1}^K \sum_{u \in \cU_{0, h'\tau}^k(s,a)} \omega_{u,t}^k (s,a) &\le \exp\left (-\frac{\eta}{2K} \sum_{k=1}^K N_{h'\tau,t}^k(s,a)\right) , \quad \forall 0\le h' \le \nsyn(t) ,\\
    \label{eq:asyncq-wsum-2-tighter}  \sum_{k=1}^K \sum_{u \in \cU_{0, t}^k(s,a)} (\omega_{u,t}^k (s,a) )^2 & \le \frac{2 \eta}{K}.
  \end{align}
  \end{subequations} 
\end{lemma}

\paragraph{Step 2: bounding the error terms.}
Here, we derive the bound of the error terms in \eqref{eq:asyncvq-error-decomp-tighter} separately for all the state-action pairs $(s,a) \in \cS \times \cA$. 

\begin{itemize}
\item \textbf{Bounding $|E_t^1(s,a)|$.} 
Using the initialization condition that $Q_0(s,a)= Q_0^k(s,a)$ for every agent $k \in [K],$ we bound the first term for any $(s,a) \in \cS \times \cA$ as follows: 
\begin{align}
  |E_t^1(s,a)|
  \le \omega_{0,t}(s,a) (\|Q_0 \|_{\infty} +  \|Q^{\star}\|_{\infty})
  \stackrel{\mathrm{(i)}}{\le} \frac{2 \omega_{0,t} (s,a)}{1-\gamma} 
  \stackrel{\mathrm{(ii)}}{\le}\frac{2}{1-\gamma}\exp\left (-\frac{\eta \muminavg t}{8} \right), 
\end{align}
where (i) holds because $\|Q_0 \|_{\infty}, ~ \|Q^{\star} \|_{\infty} \le \frac{1}{1-\gamma}$ (cf. \eqref{eq:bounded_iterates})
and (ii) follows from the fact that
\begin{align}
  \omega_{0,t} (s,a)
  \le \exp\left (-\frac{\eta}{2K} \sum_{k=1}^K N_{0,t}^k(s,a)\right)
  \le \exp\left (-\frac{\eta \muminavg t}{ 8 } \right),
\end{align}
where the first inequality holds according to \eqref{eq:asyncq-lambda-tighter} of Lemma~\ref{lemma:asyncq-wsum-tighter}, and the last inequality follows from the fact that $\sum_{k=1}^K N_{0, t}^k(s,a) \ge \frac{ K \muminavg t}{ 4 }$ for all $(s, a, k, h) \in\cS\times\cA\times [K]\times [T]$ at least with probability $1-\delta$ according to Lemma~\ref{lemma:dist-asyncwq-multi-visit-mix} and the union bound, as long as $t \ge \tth$.

\item \textbf{Bounding $|E_t^2(s,a)|$.} By carefully treating the statistical dependency via a decoupling argument and applying Freedman's inequality, we can obtain the following bound, whose proof is provided in Appendix~\ref{proof:asyncq-error2-freedman-tighter}.
\begin{lemma} \label{lemma:asyncq-error2-freedman-tighter}
  For any given $\delta  \in (0,1)$, the following holds for any $(s,a)  \in \cS \times \cA$ and $1 \le t \le T$:
  \begin{align}
    \left | E_t^2(s,a) \right|
    \le 
    \frac{ 7241 \gamma }{(1-\gamma)} \sqrt{\frac{\Csim \eta }{K} \log{(TK)} \log{\frac{4|\cS||\cA|T^2 K}{\delta}}}
  \end{align}
  with probability at least $1- 4 \delta$, as long as $\tau \ge \tth$ and $\frac{3}{T} \le \eta \le \min \Big\{\frac{1}{16\tau}, \frac{1}{4\tau K}, \frac{1}{128 K \Csim \log{(TK)}\log{\frac{4|\cS||\cA|T^2 K}{\delta}}} \Big\}$. 
\end{lemma}

\item \textbf{Bounding $|E_t^3(s,a)|$.}
  For $E_t^3$, we can obtain the following recursive relation, whose proof is provided in Appendix~\ref{proof:asyncvq-e3}.
\begin{lemma} \label{lemma:asyncvq-e3}
Let $\beta$ be any integer that satisfies $0< \beta \leq \nsyn(T)$. For any given $\delta \in (0,1)$, the following holds   
\begin{align}
  |E_t^3(s,a)|
  &\le \frac{2\gamma }{1-\gamma} \exp\left (-\frac{\eta \muminavg \beta \tau}{ 8} \right) + \frac{8 \gamma \eta  \sqrt{\tau-1}}{1-\gamma}\sqrt{\log{\frac{2|\cS||\cA|T K }{\delta}}} + \frac{1+\gamma}{2} \max_{\nsyn(t)-\beta \le h \le \nsyn(t)-1} \|\Delta_{h\tau}\|_{\infty},
\end{align}
for all $\beta \tau \le t \leq  T$ with probability at least $1-\delta$, as long as $\beta\tau \ge \tth$ and $\eta \le \min\{ \frac{1-\gamma}{4\gamma \tau}, \frac{1}{2\tau} \}$. 
\end{lemma}

\end{itemize}

\paragraph{Step 3: solving a recursive relation.}
By putting all the bounds derived in the previous step together, for any $\beta \tau \leq t \leq T$, the total error bound can be written in a simple recursive form as follows:
\begin{align}
  \|\Delta_t\|_{\infty}
  \le \theta + \frac{1+\gamma}{2} \max_{\nsyn(t)-\beta \le h \le \nsyn(t)-1} \|\Delta_{h\tau}\|_{\infty},
\end{align}
where we define
\begin{align} \label{eq:asyncvq_def_theta-tighter}
  \theta
  &\defn \frac{4}{1-\gamma} \exp\left (- { \frac{\eta \muminavg \beta \tau}{8} }\right)
    + \frac{ {7241} \gamma}{(1-\gamma)} \sqrt{\frac{\Csim \eta }{K} \log{(TK)}\log{\frac{4 |\cS||\cA| {T^2 K}}{\delta}} } 
    + \frac{8 \gamma \eta  \sqrt{\tau-1}}{1-\gamma}\sqrt{\log{\frac{2|\cS||\cA|T K }{\delta}}}.
\end{align}

Then, by invoking the recursive relation for $L_1$ times, where the choices of $\beta$ and $L_1$ will be made momentarily, it follows that for any $L_1\beta\tau \leq t\leq T $,
\begin{align} \label{eq:asyncvq-errorall-recursive-tighter}
  \|\Delta_t\|_{\infty}
  &\le \sum_{l=0}^{L_1 -1} \left( \frac{1+ \gamma}{2} \right)^l \theta + \left(\frac{1+\gamma}{2}\right)^{L_1} \max_{\nsyn(t)-\beta L \le i \le \nsyn(t)-1}  \|\Delta_{i \tau}\|_{\infty} 
  \le \frac{2}{1-\gamma} \left (\theta + \left(\frac{1+\gamma}{2} \right)^{L_1} \right ),
\end{align}
where the last inequality follows from \eqref{eq:crude}.

Setting $\beta= \left \lfloor \frac{1}{\tau} \sqrt{\frac{ {2}(1-\gamma)T}{ {\muminavg} \eta}} \right \rfloor$ and $L_1 = \left \lceil \frac{1}{2}\sqrt{\frac{ \muminavg \eta T}{(1-\gamma)}} \right \rceil$
, which ensures $L_1 \beta\tau\leq T$, and plugging the choices into \eqref{eq:asyncvq_def_theta-tighter} and \eqref{eq:asyncvq-errorall-recursive-tighter} at $t=T$, we obtain
\begin{align}
   \|\Delta_T\|_{\infty} \ 
  &\le \frac{8 \exp\left (- {\frac{\eta \muminavg \beta \tau}{8}} \right) }{(1-\gamma)^2} 
    + \frac{ {14481} \gamma}{(1-\gamma)^2} \sqrt{\frac{\Csim \eta }{K} \log{(TK)}\log{\frac{4 |\cS||\cA| {T^2 K}}{\delta}} } \cr
    &\qquad + \frac{16 \gamma \eta  \sqrt{\tau-1}}{(1-\gamma)^2}\sqrt{\log{\frac{2|\cS||\cA|T K }{\delta}}}
    + \frac{2}{1-\gamma} \left(\frac{1+\gamma}{2}\right)^L \nonumber \\
  &\le \frac{16}{(1-\gamma)^2}  \exp\left(-\frac{\sqrt{(1-\gamma)  {\muminavg} \eta T}}{8} \right) 
  + \frac{14481 \gamma}{(1-\gamma)^2} \sqrt{\frac{ \Csim \eta }{K} \log{(TK)}\log{\frac{4 |\cS||\cA| {T^2 K}}{\delta}} } \cr
  &\qquad  + \frac{16 \gamma \eta  \sqrt{\tau-1}}{(1-\gamma)^2}\sqrt{\log{\frac{2|\cS||\cA|T K }{\delta}}} \nonumber \\
  &\le \frac{14497}{(1-\gamma)^2}   \left ( \exp\left(-\frac{\sqrt{(1-\gamma) {\muminavg} \eta T}}{ {8}} \right) + \gamma \sqrt{\frac{ \Csim \eta}{K} \log{(TK)}\log{\frac{4 |\cS||\cA| {T^2 K}}{\delta}} }  \right)  ,
\end{align}
where the second line follows from
\begin{align*}
     \exp\left (-\frac{\eta \muminavg \beta \tau}{8} \right) &\le \exp\left(-\frac{\sqrt{(1-\gamma) {\muminavg} \eta T}}{8} \right), \\
    \left(\frac{1+\gamma}{2}\right)^{L_1} 
    &= \left(1-\frac{1-\gamma}{2} \right)^{L_1} 
    \le \exp \left( -\frac{1-\gamma}{2}L_1 \right) 
    \le \exp\left(-\frac{\sqrt{(1-\gamma) {\muminavg} \eta T}}{4} \right),
\end{align*}
and the third line follows from the choice of the synchronization period such that
\begin{equation} 
  {\tth }\le \tau \le \frac{1}{4 \eta} \min \left\{  \frac{1-\gamma}{4}, \, \frac{1}{K} \right\} .
\end{equation}

Thus, for any given $\varepsilon \in (0,\frac{1}{1-\gamma}]$, we can guarantee that $\|\Delta_T\|_{\infty} \le \varepsilon$ if
\begin{align*}
  T
  &\ge c_T (\log((1-\gamma)^2\varepsilon))^2 \log{(TK)}\log{\frac{4 |\cS||\cA| {T^2 K}}{\delta}}  \frac{1}{ {\muminavg}}
    \max \left \{
    \frac{\Csim }{K (1-\gamma)^5 \varepsilon^2}  ,
    \frac{\tmaxmix}{  {\muminavg} (1-\gamma)\min\{1-\gamma,K^{-1}\}}
    \right \} ,\cr  
  \eta 
         &= c_{\eta} \left(\log{(TK)}\log{\frac{4 |\cS||\cA| {T^2 K}}{\delta}} \right)^{-1} \min \left \{
         \frac{K (1-\gamma)^4 \varepsilon^2}{\Csim } ,
         \frac{ {\muminavg}  \min\{1-\gamma,K^{-1} \}}{\tmaxmix}
         \right \} 
\end{align*}
for some sufficiently large $c_T$ and sufficiently small $c_\eta$.

%% file: fedasyncq_analysis_impavg.tex

\subsection{Proof outline of Theorem~\ref{thm:dist-asyncwq-localupdate}}

The proof of Theorem~\ref{thm:dist-asyncwq-localupdate} consists of the following steps.
\paragraph{Step 1: error decomposition.}
Consider any $0 \le t \le T$ such that $t \equiv 0 ~(\text{mod}~ \tau)$, i.e., $t$ is a synchronization step. 
To analyze $\Delta_t$, invoking the recursive relation of the local error (cf. \eqref{eq:asynq-local-onestepdecomp-tighter}), we first decompose the error for each $(s,a) \in \cS \times \cA$ as follows:
\begin{align} \label{eq:asyncwq_error_sync_point}
  \Delta_t(s,a)
  &= \sum_{k=1}^K \alpha_t^k(s,a)  ( Q^{\star}(s,a) - Q_{t-\frac{1}{2}}^k(s,a)  ) \cr
  &= \left ( \sum_{k=1}^K \alpha_t^k(s,a)  (1-\eta)^{N_{t-\tau,t}^k(s,a)}\right ) \Delta_{t-\tau}(s,a) \cr
  &\qquad + \gamma \sum_{k=1}^K \alpha_t^k(s,a) \sum_{u \in \cU_{t-\tau, t}^k(s,a)} \eta (1-\eta)^{N_{u+1,t}^k(s,a)}  (P(s,a) -  P_{u+1}^k(s,a) ) V^k_{u} \cr
  &\qquad + \gamma \sum_{k=1}^K \alpha_t^k(s,a) \sum_{u \in \cU_{t-\tau, t}^k(s,a) } \eta (1-\eta)^{N_{u+1,t}^k(s,a)}  P(s,a) (V^{\star} - V^k_{u}) \cr
  &= \left ( \frac{K}{\sum_{k'=1}^K (1-\eta)^{-N_{t-\tau,t}^{k'}(s,a)}}\right ) \Delta_{t-\tau}(s,a) \cr
  &\qquad + \gamma \sum_{k=1}^K   \sum_{u \in \cU_{t-\tau, t}^k(s,a)} \frac{\eta (1-\eta)^{-N_{t-\tau,u+1}^k(s,a) }}{\sum_{k'=1}^K (1-\eta)^{-N_{t-\tau,t}^{k'}(s,a)}} (P(s,a) -  P_{u+1}^k(s,a) ) V^k_{u} \cr
  &\qquad + \gamma \sum_{k=1}^K \sum_{u \in \cU_{t-\tau, t}^k(s,a)} \frac{\eta (1-\eta)^{-N_{t-\tau,u+1}^k(s,a) }}{\sum_{k'=1}^K (1-\eta)^{-N_{t-\tau,t}^{k'}(s,a)}}  P(s,a) (V^{\star} - V^k_{u}),
\end{align}
where the last line uses the definition of $\alpha_t^k(s,a)$ in \eqref{eq:asyncq-weight-skewed}. Denoting 
\begin{equation}\label{eq:asyncwq_def_lambda_vvv}
\widetilde{\lambda}_{v_1,v_2}(s,a) \defn \frac{K}{\sum_{k=1}^K (1-\eta)^{N_{v_1,v_2}^k(s,a)}}
\end{equation}
for any integer $0 \le v_1 \le v_2 \le T$, we apply recursion to the relation \eqref{eq:asyncwq_error_sync_point} over the synchronization period, and obtain  
\begin{align} \label{eq:asyncwq-error-decomp}
 & \Delta_t(s,a) \cr
  &= \left( \prod_{h=0}^{\nsyn(t) -1} \widetilde{\lambda}_{h\tau, (h+1)\tau}(s,a) \right)  \Delta_{0}(s,a) \cr
  &\quad +  \sum_{h=0}^{\nsyn(t) -1} \left ( \prod_{l = (h+1)}^{\nsyn(t)-1} \widetilde{\lambda}_{l\tau, (l+1)\tau}(s,a) \right ) \gamma \sum_{k=1}^K   \sum_{u \in \cU_{h\tau, (h+1)\tau}^k(s,a)} \frac{\eta (1-\eta)^{-N_{h\tau,u+1}^k(s,a) }}{\sum_{k'=1}^K (1-\eta)^{-N_{h\tau,(h+1)\tau}^{k'}(s,a)}}  (P(s,a) -  P_{u+1}^k(s,a) ) V^k_{u} \cr
  &\quad +  \sum_{h=0}^{\nsyn(t) -1}\left ( \prod_{l = (h+1)}^{\nsyn(t)-1} \widetilde{\lambda}_{l\tau, (l+1)\tau}(s,a) \right )  \gamma \sum_{k=1}^K   \sum_{u \in \cU_{h\tau, (h+1)\tau}^k(s,a)} \frac{\eta (1-\eta)^{-N_{h\tau,u+1}^k(s,a) }}{\sum_{k'=1}^K (1-\eta)^{-N_{h\tau,(h+1)\tau}^{k'}(s,a)}} P(s,a) (V^{\star} - V^k_{u}) \cr
  &= \underbrace{ \widetilde{\omega}_{0,t}(s,a) \Delta_{0}(s,a) \vphantom{ \sum_{k=1}^K \sum_{u \in \cU_{0, t}^k(s,a)}}}_{=: E_t^1(s,a)} 
+ \underbrace{ \gamma \sum_{k=1}^K \sum_{u \in \cU_{0, t}^k(s,a)} \widetilde{\omega}_{u,t}^k(s,a) (P(s,a) -  P_{u+1}^k(s,a) ) V^k_{u} }_{=: E_t^2(s,a)} \cr
  &\qquad + \underbrace{ \gamma \sum_{k=1}^K \sum_{u \in \cU_{0, t}^k(s,a)} \widetilde{\omega}_{u,t}^k(s,a)  P(s,a) (V^{\star} - V^k_{u}) }_{=: E_t^3(s,a)},
\end{align}
which is again decomposed similarly as \eqref{eq:dist-syncq-error-decomp}. Here, we define
\begin{subequations}\label{eq:asyncwq-weight-def}
\begin{align} 
  \widetilde{\omega}_{0,t}(s,a) &\defn \prod_{h = 0}^{\nsyn(t)-1} \widetilde{\lambda}_{h\tau, (h+1)\tau}(s,a) ,   \\
  \widetilde{\omega}_{u,t}^k(s,a) &\defn \frac{\eta (1-\eta)^{-N_{\nsyn(u)\tau,u+1}^k(s,a) }}{\sum_{k'=1}^K (1-\eta)^{-N_{\nsyn(u)\tau,(\nsyn(u)+1)\tau}^{k'}(s,a)}} \left ( \prod_{l = \nsyn(u)+1}^{\nsyn(t)-1} \widetilde{\lambda}_{l\tau, (l+1)\tau}(s,a) \right ).
\end{align}
\end{subequations}

We record the following useful lemma whose proof is provided in Appendix~\ref{proof:asyncwq-wsum}.
\begin{lemma} \label{lemma:asyncwq-wsum}
Consider any integers $0 \le v_1 \le v_2  \le t \le T$ where $t \equiv 0 ~(\text{mod}~ \tau)$ and any state-action pair $(s,a) \in \cS \times \cA$. Suppose that $\eta\tau \le 1$, then the parameters defined in \eqref{eq:asyncwq-weight-def} satisfy
\begin{subequations}
 \begin{align}
    \label{eq:asyncwq-alpha}  \frac{1}{3K} \le  \alpha_{t}^k(s,a) & \le \frac{3}{K}, \\
    \label{eq:asyncwq-w0}  \widetilde{\omega}_{0,t}(s,a) &\le (1-\eta)^{\frac{1}{K} \sum_{k=1}^K N_{0,t}^k(s,a)},\\
    \label{eq:asyncwq-wsum-1}  \widetilde{\omega}_{0,t}(s,a) + \sum_{k=1}^K \sum_{u \in \cU_{0, t}^k(s,a)} \widetilde{\omega}_{u,t}^k(s,a) &= 1, \\
    \label{eq:asyncwq-wsum-1-part}  \sum_{k=1}^K \sum_{u \in \cU_{0, h'\tau}^k(s,a)} \widetilde{\omega}_{u,t}^k (s,a)& \le (1-\eta)^{\frac{1}{K}\sum_{k=1}^K N_{h'\tau, t}^k(s,a)}, \quad \forall 0\le h' \le \nsyn(t) , \\
    \label{eq:asyncwq-wsum-2} \sum_{k=1}^K \sum_{u \in \cU_{0, t}^k(s,a)} (\widetilde{\omega}_{u,t}^k(s,a))^2 & \le \frac{6 \eta}{K} .
\end{align}
\end{subequations} 
\end{lemma}

\paragraph{Step 2: bounding the error terms.}
Here, we derive the bound of each error term in \eqref{eq:asyncwq-error-decomp} separately for all the state-action pairs $(s,a) \in \cS \times \cA$. 

\begin{itemize}
\item  \textbf{Bounding $|E_t^1(s,a)|$.} Using the initialization condition that $Q_0(s,a)= Q_0^k(s,a)$ for every client $k \in [K],$ we bound the first term for any $(s,a) \in \cS \times \cA$ as follows: 
\begin{align}
  |E_t^1(s,a)|
  &\le \widetilde{\omega}_{0,t} (\|Q_0 \|_{\infty} +  \|Q^{\star}\|_{\infty})
  \stackrel{\mathrm{(i)}}{\le} \frac{2 \widetilde{\omega}_{0,t}}{1-\gamma} 
  \stackrel{\mathrm{(ii)}}{\le} \frac{2}{1-\gamma} (1-\eta)^{\frac{1}{K} \sum_{k=1}^K N_{0, t}^k(s,a)} 
  \stackrel{\mathrm{(iii)}}{\le} \frac{2}{1-\gamma}(1-\eta)^{\frac{1}{4} \muminavg  t},
\end{align}
where (i) holds because $\|Q_0 \|_{\infty}, ~ \|Q^{\star} \|_{\infty} \le \frac{1}{1-\gamma}$ (cf. \eqref{eq:bounded_iterates}), 
(ii) follows from \eqref{eq:asyncwq-w0} of Lemma~\ref{lemma:asyncwq-wsum},
and (iii) holds for all $(s, a, t) \in\cS\times\cA \times [T]$ with probability at least $1-\delta$ according to Lemma~\ref{lemma:dist-asyncwq-multi-visit-mix}, as long as {$t \ge \tth$}.

\item \textbf{Bounding $|E_t^2(s,a)|$.} By carefully treating the statistical dependency via a decoupling argument and applying Freedman's inequality, we can obtain the following bound, whose proof is provided in Appendix~\ref{proof:asyncwq-error2-freedman}.
\begin{lemma}\label{lemma:asyncwq-error2-freedman}
 For any given $\delta  \in (0,1)$, the following holds for any $(s,a)  \in \cS \times \cA$ and $1 \le t \le T$:
  \begin{align}
    \left | E_t^2(s,a) \right|
    \le 
\frac{2064 \gamma}{(1-\gamma)} \sqrt{\frac{\eta }{K} {\log{(TK)}} \log{\frac{4 |\cS||\cA| {T^2 K}}{\delta}} }
  \end{align}
  with probability at least $1-2\delta$, as long as 
  $$\frac{3}{T} < \eta \le \min \Big\{\frac{1}{16\tau}, \frac{K}{256 \log{(TK)}\log{\frac{4|\cS||\cA|T^2 K}{\delta}}},\, \frac{1}{34816 \tmaxmix \log{(8K)} \log{\frac{4 |\cS||\cA|T^2}{\delta}}} \Big\}.$$
\end{lemma}

\item \textbf{Bounding $|E_t^3(s,a)|$.}
For $E_t^3$, similarly to Lemma~\ref{lemma:asyncvq-e3}, we can obtain the following recursive relation, whose proof is provided in Appendix~\ref{proof:asyncwq-e3}. 
\begin{lemma} \label{lemma:asyncwq-e3}
Let $\beta$ be any integer that satisfies {$\frac{\tth}{\tau} \leq \beta \leq \nsyn(T)$}. For any given $\delta \in (0,1)$, the following holds   
\begin{align}
  |E_t^3(s,a)|
  &\le \frac{2(1-\eta)^{ \frac{\muminavg \beta \tau }{4} } }{1-\gamma} 
  + \frac{8 \gamma \eta  \sqrt{\tau-1}}{1-\gamma}\sqrt{\log{\frac{2|\cS||\cA|T K }{\delta}}} 
  + \frac{1+\gamma}{2} \max_{\nsyn(t)-\beta \le h \le \nsyn(t)-1}  \|\Delta_{h\tau}\|_{\infty},
\end{align}
for all $\beta \tau \le t \leq  T$ with probability at least $1-\delta$, as long as {$\eta \le \min\{ \frac{1-\gamma}{4\gamma \tau}, \frac{1}{2\tau} \}$}. 
\end{lemma}
\end{itemize}

\paragraph{Step 3: solving a recursive relation.}
By putting all the bounds derived in the previous step together, for any $\beta \tau \leq t \leq T$, the total error bound can be written in a simple recursive form as follows: 
\begin{align}
  \|\Delta_t\|_{\infty}
  \le \theta + \frac{1+\gamma}{2} \max_{\nsyn(t)-\beta \le h \le \nsyn(t)-1} \|\Delta_{h\tau}\|_{\infty},
\end{align}
where we define
\begin{align} \label{eq:asyncwq_def_theta}
  \widetilde{\theta}
  &\defn \frac{4}{1-\gamma} (1-\eta)^{\frac{\muminavg \beta \tau}{4} }
    + \frac{2064\gamma}{(1-\gamma)} \sqrt{\frac{\eta }{K} \log{(TK)}\log{\frac{4 |\cS||\cA| {T^2 K}}{\delta}} }
    + \frac{8 \gamma \eta  \sqrt{\tau-1}}{1-\gamma}\sqrt{\log{\frac{2|\cS||\cA|T K }{\delta}}} .
\end{align}

Then, by invoking the recursive relation for $L_2$ times, where the choices of $\beta$ and $L_2$ will be made momentarily, it follows that for any $L_2\beta\tau \leq t\leq T $,
\begin{align} \label{eq:asyncwq-errorall-recursive}
  \|\Delta_t\|_{\infty}
  &\le \sum_{l=0}^{L_2-1} \left( \frac{1+ \gamma}{2} \right)^l \widetilde{\theta} + \left(\frac{1+\gamma}{2}\right)^{L_2} \max_{\nsyn(t)-\beta L \le i \le \nsyn(t)-1}  \|\Delta_{i \tau}\|_{\infty} 
  \le \frac{2}{1-\gamma} \left (\theta + \left(\frac{1+\gamma}{2}\right)^{L_2} \right ),
\end{align}
where the last inequality follows from \eqref{eq:crude}.

Setting $L_2= \left \lceil \frac{1}{2}\sqrt{\frac{\muminavg \eta T}{(1-\gamma)}} \right \rceil$ and $\beta = \left \lfloor \frac{1}{\tau} \sqrt{\frac{2(1-\gamma)T}{\muminavg \eta }} \right \rfloor$, which ensures $L_2\beta\tau\leq T$, and plugging the choices into \eqref{eq:asyncwq_def_theta} and \eqref{eq:asyncwq-errorall-recursive} at $t=T$, we obtain
\begin{align} \label{eq:asyncwq-err-final}
  \|\Delta_T\|_{\infty}
  &\le \frac{8(1-\eta)^{\frac{\muminavg \beta \tau}{4} }}{(1-\gamma)^2} 
    + \frac{4128\gamma}{(1-\gamma)^2} \sqrt{\frac{\eta }{K} \log{(TK)}\log{\frac{4 |\cS||\cA| {T^2 K}}{\delta}} } \cr
    &\qquad + \frac{16 \gamma \eta  \sqrt{\tau-1}}{(1-\gamma)^2}\sqrt{\log{\frac{2|\cS||\cA|T K }{\delta}}} + \frac{2}{1-\gamma}\left(\frac{1+\gamma}{2}\right)^{L_2} \nonumber \\
  &\le \frac{16}{(1-\gamma)^2}  \exp\left(-\frac{\sqrt{(1-\gamma)\muminavg \eta T}}{4} \right)
  + \frac{4128\gamma}{(1-\gamma)^2} \sqrt{\frac{\eta }{K} \log{(TK)}\log{\frac{4 |\cS||\cA| {T^2 K}}{\delta}} } \cr
    &\qquad + \frac{16 \gamma \eta  \sqrt{\tau-1}}{(1-\gamma)^2}\sqrt{\log{\frac{2|\cS||\cA|T K }{\delta}}} \nonumber \\
  &\le \frac{4144}{(1-\gamma)^2}   \left ( \exp\left(-\frac{\sqrt{(1-\gamma)\muminavg \eta T}}{4} \right) + \gamma \sqrt{\frac{\eta}{K} \log{(TK)}\log{\frac{4 |\cS||\cA| {T^2 K}}{\delta}}}  \right),
\end{align}
where the second line follows from
\begin{align*}
    (1-\eta)^{\frac{\muminavg \beta \tau}{4} } 
    &\le \exp\left (-\frac{\eta \muminavg \beta \tau}{4} \right) 
    \le \exp\left(-\frac{\sqrt{(1-\gamma)\muminavg \eta T}}{4} \right), \\
    \left(\frac{1+\gamma}{2}\right)^{L_2} 
    &= \left(1-\frac{1-\gamma}{2} \right)^{L_2} 
    \le \exp \left( -\frac{1-\gamma}{2}L_2 \right) 
    \le \exp\left(-\frac{\sqrt{(1-\gamma)\muminavg \eta T}}{4} \right),
\end{align*}
and the third line follows from the choice of the synchronization period such that
\begin{equation} 
 \tau \le \frac{1}{4 \eta} \min \left\{  \frac{1-\gamma}{4}, \, \frac{1}{K} \right\} .
\end{equation}

Thus, for any given $\varepsilon \in (0,\frac{1}{1-\gamma})$, optimizing $\eta$ and $T$ to make \eqref{eq:asyncwq-err-final} bounded by $\varepsilon$ and recalling $\beta\tau \ge \tth$, we can guarantee that $\|\Delta_T\|_{\infty} \le \varepsilon$ if
\begin{align*}
  T
  &\ge c_T (\log((1-\gamma)^2\varepsilon))^2 \log{(TK)}\log{\frac{4 |\cS||\cA| {T^2 K}}{\delta}}  \frac{1}{\muminavg }  
  \max \left \{\frac{1}{K (1-\gamma)^5 \varepsilon^2} , 
  \frac{  \tmaxmix }{(1-\gamma)},
    \frac{1}{(1-\gamma) \min \left\{  1-\gamma,  K^{-1} \right\}}
  \right \} ,\\
\eta &= c_{\eta} \min \left \{
         K (1-\gamma)^4 \varepsilon^2 \frac{1}{\log{(TK)}\log{\frac{4 |\cS||\cA| {T^2 K}}{\delta}}},
         \frac{1}{\muminavg \tth}, \frac{1}{\tmaxmix \log{(TK)}\log{\frac{4 |\cS||\cA| {T^2 K}}{\delta}}}
         \right \} \cr  
    &= c_{\eta} \left ( \log{(TK)}\log{\frac{4 |\cS||\cA| {T^2 K}}{\delta}} \right)^{-1}\min \left \{
         K (1-\gamma)^4 \varepsilon^2 ,
         \frac{1}{\tmaxmix },
         \min \left\{  1-\gamma,  K^{-1} \right\}
        \right \}
\end{align*}
for some sufficiently large $c_T$ and sufficiently small $c_\eta$.

%% file: conclusion.tex
\section{Discussions}
We presented a sample complexity analysis of federated Q-learning in both synchronous and asynchronous settings. Our sample complexity not only leads to linear speedup with respect to the number of agents, but also significantly improves the dependencies on other salient problem parameters over the prior art. For federated asynchronous Q-learning, we proposed a novel importance averaging scheme that weighs the agents' local Q-estimates according to the number of visits to each state-action pair. This allows agents to leverage the blessing of heterogeneity of their local behavior policies and collaboratively learn the optimal Q-function that otherwise would not be possible, without requiring each individual agent to cover the entire state-action space.
Looking ahead, this work opens up many exciting future directions, some outlined below.
\begin{itemize}
\item {\em Improved sample complexity.} While our sample complexity bounds are near-optimal with respect to the size of the state-action space, it is still sub-optimal with respect to the effective horizon length as well as the mixing time when benchmarking with the sample complexity in the single-agent setting \citep{li2021syncq}. It will be interesting to close this gap, and further improve the sample complexity with variance reduction techniques \citep{wainwright2019variance,li2021asyncq} in the federated setting. 

\item {\em Understanding communication asynchrony across agents.} As a starting point, our work assumes that all agents communicate with the server in a synchronous manner to perform periodic averaging. However, in practical federated networks, some agents might be stragglers due to communication slowdowns, which warrants further investigation \citep{kairouz2021advances}. 

\item {\em Other RL settings and function approximation.} Besides the infinite-horizon tabular MDPs, it will be of great interest to extend our analysis framework to other RL settings including but not limited to the finite-horizon setting, the average reward setting, heterogeneous environments across the agents \citep{yang2023federated}, as well as incorporating function approximation.

\item {\em Federated offline RL.} In many applications, offline RL is attracting a growing amount of interest, which aims to explore history datasets to improve the learned policy without exploration, e.g. via pessimistic variants of Q-learning \citep{shi2022pessimistic}. It will be appealing to develop federated offline Q-learning algorithms to enable learning from geographically distributed history datasets.

\end{itemize}

%% file: preliminary.tex
\section{Preliminaries}

We record a few useful inequalities that will be used throughout our analysis. To start with, our analysis leverages Freedman's inequality
\citep{freedman1975tail}, which we record a user-friendly version as follows.

\begin{theorem}[Theorem~6 in \cite{li2021syncq}] \label{thm:Freedman}
Suppose that $Y_{n}=\sum_{k=1}^{n}X_{k}\in\mathbb{R}$,
where $\{X_{k}\}$ is a real-valued scalar sequence obeying 
\[
\left|X_{k}\right|\leq R\qquad\text{and}\qquad\mathbb{E}\left[X_{k}\mid\left\{ X_{j}\right\} _{j:j<k}\right]=0\quad\quad\quad\text{for all }k\geq1.
\]
Define
\[
W_{n}\coloneqq\sum_{k=1}^{n}\mathbb{E}_{k-1}\left[X_{k}^{2}\right],
\]
where we write $\mathbb{E}_{k-1}$ for the expectation conditional
on $\left\{ X_{j}\right\} _{j:j<k}$. Then for any given $\sigma^{2}\geq0$,
one has
\begin{equation}
\mathbb{P}\left\{ \left|Y_{n}\right|\geq\tau\text{ and }W_{n}\leq\sigma^{2}\right\} \leq2\exp\left(-\frac{\tau^{2}/2}{\sigma^{2}+R\tau/3}\right).\label{eq:Freedmans-general}
\end{equation}
In addition, suppose that $W_{n}\leq\sigma^{2}$ holds deterministically.
For any positive integer $m\geq1$, with probability at least $1-\delta$
one has
\begin{equation}
\left|Y_{n}\right|\leq\sqrt{8\max\Big\{ W_{n},\frac{\sigma^{2}}{2^m}\Big\}\log\frac{2m}{\delta}}+\frac{4}{3}R\log\frac{2m}{\delta}.\label{eq:Freedman-random}
\end{equation}
\end{theorem}

Another useful relation concerns the concentration of empirical distributions of uniformly ergodic Markov chains, which is rephrased from \citet{li2021asyncq}.
 

\begin{lemma}[{\cite[Lemma~8]{li2021asyncq}}]
  \label{lemma:Bernstein-state-occupancy}
  Consider any time homogeneous and uniformly ergodic Markov chain $(X_0, X_1, X_2,\ldots)$ with transition kernel $P$, finite state
space $\cX$, and stationary distribution $\mu$. Let $\tmix$ be the mixing time of the Markov chain and $\mumin $ be the minimum entry of the stationary distribution $\mu$. Consider any $0<\delta<1$. For any $x\in\mathcal{X}$, if $t \geq  \frac{443 \tmix}{\nu}\log\frac{4|\mathcal{X}|}{\delta}$ for $\nu \ge \mu(x)$, then
  \begin{align*}
    \forall y \in \mathcal{X}:\quad  \mathbb{P}_{X_{1}=y}\Bigg\{ \left|\sum_{i= 1}^{t}  \ind\{X_{i}=x\} - t \mu(x)\right| \geq \frac{1}{2}t \nu \Bigg\} 
    \leq \frac{\delta}{|\cX|}.
  \end{align*}
\end{lemma}
\begin{remark}
Lemma~\ref{lemma:Bernstein-state-occupancy} is a slightly generalized version of  in \citet[Lemma~8]{li2021asyncq}, where the concentration bound is characterized in terms of any given threshold $\nu \ge \mu(x)$, not scaling with the stationary distribution $\mu(x)$. It can be shown using the Bernstein's inequality for Markov chains \citep[Theorem~3.11]{paulin2015concentration} in the same manner as \citet[Lemma~8]{li2021asyncq}, except that the threshold is set to $\frac{\nu t}{2}$ instead of $\frac{\mu(x) t}{2}$. We omit further details for conciseness and refer interested readers to the proof in \cite{li2021asyncq}.
\end{remark}

In addition, we provide the concentration bound of the total number of visits of multiple agents agents with independent uniformly ergodic Markov chains, whose proof is provided in Appendix~\ref{proof:lemma:dist-asyncwq-multi-visit-mix}.
Denote
\begin{align} \label{eq:def_tth}
  \tth(s,a) \defn \frac{2176 \tmaxmix \log{8K} \log{\frac{4|\cS||\cA|T^2}{\delta}}}{\muminavg(s,a)}  ~~\text{and}~~ \tth \defn \frac{2176 \tmaxmix \log{8K} \log{\frac{4|\cS||\cA|T^2}{\delta}}}{\muminavg} .
\end{align}  
Here, $\muminavg(s,a) \defn \frac{1}{K} \sum_{k =1}^K \mu_{\mathsf{b}}^k(s,a)$ is the average behavior policy over all agents.

\begin{lemma} \label{lemma:dist-asyncwq-multi-visit-mix}
  Consider any $\delta \in (0,1)$. Under the asynchronous sampling, for any $(s,a) \in \cS \times \cA$ and $0 \le u < v \le T$ such that $v-u \ge \tth(s,a)$, the following holds : 
  \begin{align} \label{eq:asyncwq-multi-visit-bound}
    \frac{1}{4} (v-u)K \muminavg(s,a) \le  \sum_{k=1}^K N_{u,v}^k(s,a) \le  2(v-u)K \muminavg(s,a)
  \end{align}
  with probability at least $1-\frac{\delta}{|\cS||\cA|T^2}$.
\end{lemma}

%% file: proof_syncq.tex

\section{Proofs for  federated synchronous Q-learning (Section~\ref{sec:fed_syncQ})}
 Define the following actions
  \begin{equation}\label{eq:optimal_a}
  a^{\star}(s) = \argmax_{a \in \cA} Q^{\star}(s,a), \quad a_i^k(s) = \argmax_{a \in\cA} Q_i^k(s,a), \quad  a_i(s) = \argmax_{a \in \cA} \frac{1}{K}\sum_{k=1}^K Q_i^k(s,a)
  \end{equation}
for any state $s\in\cS$, which will be useful throughout the proof.

\subsection{Proof of Lemma~\ref{lemma:dist-syncq-error2-freedman}}
\label{proof:dist-syncq-error2-freedman}
 
For notation simplicity, let $z_i^k(s,a) \defn \eta (1-\eta)^{t-i}  (P(s,a) - P_{i}^k(s,a) ) V^k_{i-1}$, then the entries of $E_t^2 =[E_t^2(s,a)]$ can be written as
\begin{align}
  E_t^2(s,a) = \eta\frac{\gamma}{K}  \sum_{i=1}^{t} (1-\eta)^{t-i}  \sum_{k=1}^K (P(s,a) - P_i^k(s,a) ) V^k_{i-1}
  &=   \frac{\gamma}{K} \sum_{i=1}^{t} \sum_{k=1}^K z_i^k(s,a),
\end{align}
which we plan to bound by invoking Freedman's inequality (cf.~Theorem~\ref{thm:Freedman}) using the fact $z_i^k(s,a)$ is independent of the transition events of other agents $k' \neq k$ at $i$ and has zero mean conditioned on the events before iteration $i$, i.e.,
\begin{align}
  \label{eq:syncq_z_mean} &\mexp[z_i^k (s,a) | V_{i-1}^{K}, \ldots, V_{i-1}^{1}, \ldots, V_0^{K}, \ldots, V_0^1 ]=0,  \qquad \forall k \in [K],~ 1 \le i \le t. 
\end{align}
Before applying Freedman's inequality, we first derive the following properties of the variable $z_i^k(s,a)$.
\begin{itemize}
\item First, we can bound
\begin{align} \label{eq:syncq_z_max} 
  B_t(s,a) & \defn \max_{k \in [K], 1 \le i \le t} |z_i^k(s,a)|  \le \max_{k \in [K], 1 \le i \le t}  \eta \big(\| P(s,a)\|_{1} +  \| P_i^k(s,a)\|_1 \big) \| V^k_{i-1}\|_{\infty} \le \frac{2\eta}{1-\gamma}  ,  
\end{align}
where the first inequality uses $(1-\eta)^{t-i}\leq 1$, and the last inequality follows from  $\| P(s,a)\|_1 \leq 1$, $\| P_i^k(s,a)\|_1 \le 1$, and $\| V_{i-1}^k\|_{\infty} \le \frac{1}{1-\gamma}$ (cf.~\eqref{eq:bounded_iterates}).

\item Next, we have
\begin{align}  \label{eq:syncq_z_sqsum}
W_t(s,a)   &         \defn \sum_{i=1}^t \sum_{k=1}^K \mexp \big[(z_i^k(s,a))^2 | V_{i-1}^{K}, \ldots, V_{i-1}^1, \ldots, V_0^K, \ldots, V_0^1 \big] \nonumber \\
&   = \sum_{i=1}^t \sum_{k=1}^K \Var\big(z_i^k(s,a) | V_{i-1}^{K}, \ldots, V_{i-1}^1, \ldots, V_0^K, \ldots, V_0^1 \big) \cr
& =  \sum_{i=1}^t \sum_{k=1}^K \eta^2(1-\eta)^{2(t-i)} \Var_{s,a} (V_{i-1}^k) \cr
&\le  \frac{2K}{(1-\gamma)^2}  \sum_{i=1}^t \eta^2  (1-\eta)^{2(t-i)} \le \frac{2 \eta K }{(1-\gamma)^2} \defn \sigma^2,  
\end{align}
where we recall the definition of $\Var_{s,a}$ in \eqref{eq:def_var}.
Here, the first inequality holds since
$$\Var_{s,a} ( V_{i-1}^k) \le \| P(s,a)\|_{1} (\| V_{i-1}^k\|_{\infty})^2 + (\| P(s,a)\|_{1} \| V_{i-1}^k \|_{\infty})^2 \le \frac{2}{(1-\gamma)^2}$$
and the last inequality follows from  
\begin{align} \label{eq:lr_apple}
\sum_{i=1}^t   \eta^2  (1-\eta)^{2(t-i)} &\leq   \frac{\eta^2 (1-(1-\eta)^{2t})}{1-(1-\eta)^2} \le \eta.
\end{align}
\end{itemize}

By substituting the above bounds (cf. \eqref{eq:syncq_z_max} and \eqref{eq:syncq_z_sqsum}) and $m=1$ into Freedman's inequality (see Theorem~\ref{thm:Freedman}), it follows that for any $s\in \cS$, $a \in \cA$ and $t\in [T]$,  
\begin{align} \label{eq:dist-syncq-error2-freedman-zsumbound}
  \left | \sum_{i=1}^t \sum_{k=1}^K z_i^k(s,a) \right |
  &\le \sqrt{8 \max{\{W_t(s,a), \frac{\sigma^2}{2^m} \}} \log{\frac{2m |\cS||\cA|T}{\delta}}} + \frac{4}{3} B_t(s,a) \log{\frac{2m |\cS||\cA|T}{\delta}} \cr
  &\le
      \sqrt{ \frac{32 \eta K}{(1-\gamma)^2} \log{\frac{|\cS||\cA|T}{\delta}}} + \frac{6\eta}{1-\gamma} \log{\frac{|\cS||\cA|T}{\delta}} \cr  
  &\le
    \frac{8 \gamma}{1-\gamma}\sqrt{ \frac{\eta}{K} \log{\frac{|\cS||\cA|T}{\delta}}} 
\end{align}
with probability at least $1-\frac{\delta}{|\cS||\cA|T},$ where the last inequality holds under the assumption $\eta \le \frac{K}{2} (\log{\frac{|\cS||\cA|T}{\delta}})^{-1}$. Applying the union bound over all 
  $s\in \cS$, $a \in \cA$ and $t\in [T]$ then completes the proof.  
 
\subsection{Proof of Lemma~\ref{lemma:syncq-e3b}}
\label{proof:syncq-e3b}

For any $\beta \tau \leq t \leq T$ and $(s,a) \in \cS \times \cA,$ we can decompose the entries of $E_t^3 =[E_t^3(s,a)]$ as  
\begin{align} \label{eq:part-error3-bound}
 & |E_t^3(s,a)|
  = \left |\frac{\eta \gamma}{K}\sum_{i=0}^{t-1} \sum_{k=1}^K (1-\eta)^{t-i-1} P(s,a) ( V^{\star} - V^k_{i}) \right | \nonumber \\
  &\le \underbrace{ \left | \frac{\eta\gamma}{K}\sum_{i=0}^{\syn(t)-\beta \tau-1} \sum_{k=1}^K (1-\eta)^{t-i-1} P(s,a) (V^{\star} - V^k_{i}) \right | }_{=:E_t^{3a}(s,a)}
  + \underbrace{ \left |  \frac{\eta\gamma }{K}\sum_{i=\syn(t)-\beta\tau}^{t-1} \sum_{k=1}^K (1-\eta)^{t-i-1} P(s,a) ( V^{\star} - V^k_{i}) \right |}_{=:E_t^{3b}(s,a)}.
\end{align}
We shall bound these two terms separately.

\paragraph{Step 1: bounding $E_t^{3a}(s,a)$.} First, the bound of $E_t^{3a}$ is obtained as follows:
\begin{align}
  E_t^{3a}(s,a)
  & \le \eta \frac{\gamma}{K}\sum_{k=1}^K \sum_{i=0}^{\syn(t)-\beta \tau-1} (1-\eta)^{t-i} \| P(s,a)\|_{1} (\| V^{\star}\|_{\infty} + \| V^k_{i}\|_{\infty}) \nonumber \\
&  \le \frac{2 \eta \gamma}{1-\gamma} \sum_{i=0}^{\syn(t)-\beta\tau-1} (1-\eta)^{t-i-1}
  \le \frac{2 \gamma}{1-\gamma} (1-\eta)^{\beta \tau},
\end{align}
where the second inequality holds due to the fact that $\| P(s,a)\|_{1} \le 1$ and $\| V^{\star}\|_{\infty} \le \frac{1}{1-\gamma}$, $\| V^k_{i}\|_{\infty} \le \frac{1}{1-\gamma}$, and the last inequality follows from 
\begin{align*}
\sum_{i=0}^{\syn(t)-\beta\tau-1} (1-\eta)^{t-i-1}  &\le  (1-\eta)^{\beta \tau} +  (1-\eta)^{\beta\tau+1} + \ldots +   (1-\eta)^{t-1} \leq \frac{ (1-\eta)^{\beta\tau} }{1-(1-\eta)}\le \frac{(1-\eta)^{\beta\tau} }{\eta}.
\end{align*} 

\paragraph{Step 2: decomposing the bound on $E_t^{3b}(s,a)$.}  Next, $E_t^{3b}(s,a)$ can be bounded as follows
\begin{align}
  E_t^{3b}(s,a)
  &=  \left |  \frac{\eta\gamma }{K}\sum_{i=\syn(t)-\beta\tau}^{t-1} \sum_{k=1}^K (1-\eta)^{t-i-1} P(s,a) (V^{\star} - V^k_{i}) \right | \cr
  &\le \gamma \sum_{i=\syn(t)-\beta\tau}^{t-1} \eta (1-\eta)^{t-i-1} \left | \frac{1}{K} \sum_{k=1}^K P(s,a) (V^{\star} - V^k_{i}) \right | \cr
  &\le \gamma \sum_{i=\syn(t)-\beta\tau}^{t-1} \eta (1-\eta)^{t-i-1}  \left \| \frac{1}{K} \sum_{k=1}^K (V^{\star} - V^k_{i}) \right \|_{\infty}  , \label{eq:newyear}
\end{align}
where the second inequality holds since $\| P(s,a)\|_{1} \le 1$.
To continue, denoting  
\begin{equation}\label{eq:newyeardelta} 
  d_{v,w}^k(s,a) \defn Q^k_{w}(s, a) - Q^k_{v}(s, a)  ,
  \end{equation}
  we claim the following bound   for any $0 \le i < T$, which will be shown in Appendix~\ref{proof:syncq-avgvgap-base}:
  \begin{align} \label{eq:syncq-avgvgap-base}
    \left \|\frac{1}{K} \sum_{k=1}^K (V^{\star} - V_i^k) \right \|_{\infty}
    &\le \|\erravg_i\|_{\infty} + 2\max_{k} \big \| d_{\syn(i), i}^k \big \|_{\infty}.
  \end{align}

  In view of \eqref{eq:syncq-avgvgap-base}, it boils down to control $\max_{k} \big \| d_{\syn(i), i}^k \big \|_{\infty}$. For any $(s,a) \in \cS \times \cA$, $k \in [K]$, and $0 \le i <T$, by the definition \eqref{eq:newyeardelta}, it follows  that
  \begin{align} \label{eqn:syncq-tdsum-base}
    \big |  d_{\syn(i), i}^k(s,a) \big |
    & = \left |\sum_{j=\syn(i)}^{i-1} d_{j, j+1}^k(s,a) \right |
    \le \underbrace{ 2 \eta  \sum_{j=\syn(i)}^{i-1}  \| \Delta_j^k\|_{\infty} }_{\defn B_1}
    + \underbrace{ \gamma \eta \left | \sum_{j=\syn(i)}^{i-1}  ( P_{j+1}^k (s,a)  - P(s,a)) V^{\star} \right | }_{\defn B_2},
  \end{align}
  where 
  \begin{equation}\label{eq:define_local_Delta}
  \Delta_j^k =  Q^{\star}-  Q_j^k.
  \end{equation}
  The inequality \eqref{eqn:syncq-tdsum-base} holds by the local update rule:
  \begin{align}
    d_{j, j+1}^k(s,a)
    &= Q_{j+1}^k(s,a) - Q_j^k(s,a) \cr
    &= \eta ( r(s,a) + \gamma P_{j+1}^k (s,a) V_j^k - Q_j^k(s,a)) \cr
    &\overset{\mathrm{(i)}}{=} \eta ( r(s,a) + \gamma P_{j+1}^k (s,a) V_j^k - r(s,a) - \gamma P(s,a) V^{\star} + Q^{\star}(s,a) - Q_j^k(s,a)) \cr
    &= \eta  (\gamma  P_{j+1}^k (s,a)  V_j^k - \gamma  P(s,a) V^{\star} +  Q^{\star}(s,a) -  Q_j^k(s,a)) \cr
    &= \gamma \eta    P_{j+1}^k (s,a) ( V_j^k - V^{\star}) + \gamma \eta  (P_{j+1}^k (s,a)  - P(s,a)) V^{\star} + \eta \Delta_j^k(s,a) \nonumber \\
    & \leq 2 \eta  \| \Delta_j^k\|_{\infty} + \gamma \eta  (P_{j+1}^k (s,a)  - P(s,a)) V^{\star},
  \end{align}
  where (i) follows from Bellman's optimality equation, and the last inequality follows from $\| P_{j+1}^k (s,a)\|_1\leq 1$ and 
  $\| V_j^k - V^{\star} \|_{\infty}  \leq \| \Delta_j^k\|_{\infty}$ (cf. \eqref{eq:vgaptoqgap}).

Next, we bound each term in \eqref{eqn:syncq-tdsum-base} separately.
\begin{itemize}
    \item \textbf{Bounding $B_1$.}
    The local error $\| \Delta_j^k \|_{\infty}$ is bounded as stated in the following lemma, whose proof is provided in Appendix~\ref{proof:syncq-error-indi}.
  \begin{lemma} \label{lemma:syncq-error-indiv}
   Assume $\tau \eta \le \frac{1}{2}$. For any given $\delta \in (0,1)$,  the following bound holds for any $1 \le i \le T$ and $k \in [K]$:
    \begin{align} \label{eq:syncq-tdsum-base-1}
      \|\Delta_{i}^k\|_{\infty} \le
      \| \Delta_{\syn(i)} \|_{\infty}
      + \frac{2}{1-\gamma} \sqrt{ \eta \log{\frac{|\cS| |\cA| K T}{\delta}}}
    \end{align}
    with at least probability $1-\delta,$ where $\syn(i)$ is the most recent synchronization step until $i.$
  \end{lemma}
    Using the fact that $i-\syn(i) \le \tau -1$, we can claim that   
    \begin{align}
         2 \eta  \sum_{j=\syn(i)}^{i-1}  \| \Delta_j^k\|_{\infty} \le 2 \eta (\tau-1) \| \Delta_{\syn(i)} \|_{\infty} + \frac{4 \eta (\tau-1)}{1-\gamma} \sqrt{ \eta \log{\frac{|\cS| |\cA| K T}{\delta}}}.
    \end{align}

    \item \textbf{Bounding $B_2$.}
    Using the fact that the empirical transitions are independent and centered on the true transition probability, by invoking Hoeffding's inequality and the union bound, we can claim that the following holds for all $(s, a, k, t) \in\cS\times\cA\times [K]\times [T]$, 
  \begin{align} \label{eq:syncq-tdsum-base-2}
    \gamma \eta \left | \sum_{j=\syn(i)}^{i-1}   (P_{j+1}^k (s,a)  - P(s,a)) V^{\star} \right |
    &\le \frac{\gamma \eta}{1-\gamma} \sqrt{\frac{1}{2}\sum_{j=\syn(i)}^{i-1}  \log{\frac{|\cS||\cA|K T}{\delta}}}  \le \frac{\gamma \eta}{1-\gamma} \sqrt{(\tau-1) \log{\frac{|\cS||\cA|K T}{\delta}}}
  \end{align}
  with probability at least $1-\delta$ for any given $\delta \in (0,1)$, where $\tau$ is the synchronization period.
\end{itemize}

By substituting the bound of $B_1$ and $B_2$ into \eqref{eqn:syncq-tdsum-base}, and applying the union bound, we obtain that:
 for any given $\delta \in (0,1)$, the following holds for any $0 \le i \leq T$ and $k \in [K]$:
    \begin{align} \label{eq:newyeareve}
      \| d_{\syn(i), i}^k  \|_{\infty}
      &\le 2  \eta(\tau-1) \| \erravg_{\syn(i)} \|_{\infty}
        + \frac{4 \eta ((\tau-1) \sqrt{\eta} + \sqrt{\tau-1})}{(1-\gamma)} \sqrt{  \log{\frac{2|\cS| |\cA| K T}{\delta}}} \nonumber \\
        & \le 2  \eta(\tau-1) \| \erravg_{\syn(i)} \|_{\infty}
        + \frac{8 \eta \sqrt{\tau-1}  }{(1-\gamma)} \sqrt{  \log{\frac{2|\cS| |\cA| K T}{\delta}}} 
    \end{align}
    with at least probability $1-\delta$, where $\syn(i)$ is the most recent synchronization step until $i$. Here, the second line uses the fact $\eta\tau<1$.

 By combining \eqref{eq:newyeareve} and \eqref{eq:syncq-avgvgap-base} and substituting it into \eqref{eq:newyear} and using the fact that $\sum_{i=\syn(t)-\beta\tau}^{t-1} \eta (1-\eta)^{t-i-1} \le 1$, we can obtain the bound $E_t^{3b}(s,a)$ as follows:
  \begin{align}
    |E_t^{3b}(s,a)| 
    &\le \frac{16 \gamma \eta \sqrt{\tau-1}  }{(1-\gamma)} \sqrt{  \log{\frac{2|\cS| |\cA| K T}{\delta}}}    + \gamma \sum_{i=\syn(t)-\beta\tau}^{t-1} \eta (1-\eta)^{t-i-1}  \left (\|\Delta_i\|_{\infty} 
    + 4  \eta(\tau-1) \| \erravg_{\syn(i)} \|_{\infty} \right) \nonumber \\
    & \leq \frac{16 \gamma \eta \sqrt{\tau-1}  }{(1-\gamma)} \sqrt{  \log{\frac{2|\cS| |\cA| K T}{\delta}}} +\gamma (1+4\eta(\tau-1)) \max_{\syn(t)-\beta \tau \le i <t} \| \erravg_{i} \|_{\infty} .
  \end{align} 

\paragraph{Step 3: putting all together.}
 Now, we have the bounds of $E_t^{3a}$ and $E_t^{3b}$ separately derived above. By combining the bounds in \eqref{eq:part-error3-bound}, we can finally claim the advertised bound 
and this completes the proof.

\subsubsection{Proof of \eqref{eq:syncq-avgvgap-base}}
\label{proof:syncq-avgvgap-base}
On one end, it follows that   for any $s\in\cS$,  
  \begin{align} \label{eq:vdiff-upper}
    \frac{1}{K} \sum_{k=1}^K \left(  V^{\star}(s)- V_i^k(s) \right )
    &= Q^{\star}(s, a^{\star}(s)) - \frac{1}{K}\sum_{k=1}^K  Q_i^k(s,a_i^k(s)) \cr
    &\le  Q^{\star}(s, a^{\star}(s)) - \frac{1}{K}\sum_{k=1}^K  Q_i^k(s,a^{\star}(s)) = \Delta_i(s,a^{\star}(s)),
  \end{align}
  where we use the definitions in \eqref{eq:optimal_a}.
On the other end, it follows that 
  \begin{align} \label{eq:vdiff-lower-part1}
     \frac{1}{K} \sum_{k=1}^K \left( V^{\star}(s)- V_i^k(s) \right )  
    &=  Q^{\star}(s, a^{\star}(s)) - \frac{1}{K}\sum_{k=1}^K  Q_i^k(s,a_{\syn(i)}(s)) + \frac{1}{K}\sum_{k=1}^K  Q_i^k(s,a_{\syn(i)}(s))  - \frac{1}{K}\sum_{k=1}^K  Q_i^k(s,a_i^k(s)) \nonumber \\
    &\ge  Q^{\star}(s, a_{\syn(i)}(s))  - \frac{1}{K}\sum_{k=1}^K  Q_i^k(s,a_{\syn(i)}(s)) + \frac{1}{K}\sum_{k=1}^K  Q_i^k(s,a_{\syn(i)}(s))  - \frac{1}{K}\sum_{k=1}^K  Q_i^k(s,a_i^k(s)) \nonumber \\
    &= \Delta_i(s,a_{\syn(i)}(s)) + \frac{1}{K}\sum_{k=1}^K  Q_i^k(s,a_{\syn(i)}(s))  - \frac{1}{K}\sum_{k=1}^K  Q_i^k(s,a_i^k(s)),
    \end{align}
   where the  inequality follows from the fact that $a^{\star}(s)$ is the optimal action   for  state $s$. Notice that the latter terms can be further lower bounded as
   \begin{align} \label{eq:vdiff-lower-part2}
  & \frac{1}{K}\sum_{k=1}^K  Q_i^k(s,a_{\syn(i)}(s))  - \frac{1}{K}\sum_{k=1}^K  Q_i^k(s,a_i^k(s)) \nonumber \\
  & = \frac{1}{K}\sum_{k=1}^K  Q_i^k(s,a_{\syn(i)}(s))  - \frac{1}{K}\sum_{k=1}^K   Q_{\syn(i)}^k(s,a_{\syn(i)}(s)) + \frac{1}{K}\sum_{k=1}^K   Q_{\syn(i)}^k(s,a_{\syn(i)}(s)) \nonumber  \\
  & \qquad -\frac{1}{K}\sum_{k=1}^K  Q_{\syn(i)}^k(s,a_i^k(s)) + \frac{1}{K}\sum_{k=1}^K  Q_{\syn(i)}^k(s,a_i^k(s)) - \frac{1}{K}\sum_{k=1}^K  Q_i^k(s,a_i^k(s)) \nonumber \\
  & \geq   \frac{1}{K} \sum_{k=1}^K \big(d_{\syn(i),i}^k(s,a_{\syn(i)}(s)) - d_{\syn(i),i}^k(s,a_i^k(s))\big) ,
   \end{align}
  where the inequality follows from the definition \eqref{eq:newyeardelta} 
 and the fact that
  $$Q_{\syn(i)}^k(s,a_{\syn(i)}(s)) -  Q_{\syn(i)}^k(s,a_i^k(s)) \ge 0.$$
  The above holds, since
  $Q_{\syn(i)}^k = Q_{\syn(i)}$ for all $k \in [K]$ agents after periodic averaging at $\syn(i)$, and $a_{\syn(i)}(s)$ is the optimal action at state $s$ at time $\syn(i)$ for every agent.

Combining \eqref{eq:vdiff-upper}, \eqref{eq:vdiff-lower-part1} and \eqref{eq:vdiff-lower-part2}, we obtain
\begin{align*}
 \Delta_i(s,a_{\syn(i)}(s)) +\frac{1}{K} \sum_{k=1}^K \big(d_{\syn(i),i}^k(s,a_{\syn(i)}(s)) - d_{\syn(i),i}^k(s,a_i^k(s))\big)  \leq \frac{1}{K} \sum_{k=1}^K \left(  V^{\star}(s)- V_i^k(s) \right ) \leq \Delta_i(s,a^{\star}(s)),
 \end{align*}
which immediately implies \eqref{eq:syncq-avgvgap-base}.

\subsubsection{Proof of Lemma~\ref{lemma:syncq-error-indiv}}
\label{proof:syncq-error-indi}

By applying the decomposition in \eqref{eq:dist-syncq-error-decomp} to the local error for agent $k$, we decompose $\Delta_i^k$ as follows:
  \begin{align} \label{eq:syncq-error-indiv-decomp}
    \Delta_{i}^k (s,a)
    =  \underbrace{ (1-\eta)^{i-\syn(i)} \Delta_{\syn(i)}^k (s,a) \vphantom{\sum_{j=\syn(i)}^{i-1}}}_{\defn D_1}
    +  & \underbrace{  \gamma \sum_{j=\syn(i)+1}^{i} \eta (1-\eta)^{i-j} (P(s,a)-P_{j}^k(s,a)) V^{\star} }_{\defn D_2} \cr
    + & \underbrace{ \gamma \sum_{j=\syn(i)+1}^{i} \eta (1-\eta)^{i-j} P_{j}^k(s,a) (V^{\star}-V_{j-1}^k) }_{\defn D_3}.
  \end{align}  
We shall bound each term separately.
\begin{itemize}
    \item \textbf{Bounding $D_1$.}
        Since $\Delta_{\syn(i)}^k = \Delta_{\syn(i)}$ for every agent $k$ at the synchronization step $\syn(i)$,
        \begin{align}
            |D_1| \le (1-\eta)^{i-\syn(i)} \| \Delta_{\syn(i)} \|_{\infty}.
        \end{align}
    \item \textbf{Bounding $D_2$.}  
    In a similar manner to \eqref{eq:syncq-tdsum-base-2}, by invoking Hoeffding inequality and using the fact that $\sum_{j=\syn(i)+1}^{i} ( \eta (1-\eta)^{i-j})^2 \le \eta$ (cf.~\eqref{eq:lr_apple}), we can claim that the following holds for all $(s, a, k, t) \in\cS\times\cA\times [K]\times [T]$, 
    \begin{align} \label{eqn:syncq-bernstein-indiv}
    | D_2 |
    \le \gamma \sqrt{\sum_{j=\syn(i)+1}^{i} ( \eta (1-\eta)^{i-j})^2 \|V^{\star}\|_{\infty}^2 \log{\frac{|\cS| |\cA| K T}{\delta}}} 
    \le \frac{\gamma}{1-\gamma} \sqrt{ \eta \log{\frac{|\cS| |\cA| K T}{\delta}}}
  \end{align}
  with probability at least $1-\delta$ for any given $\delta \in (0,1)$.
  
    \item \textbf{Bounding $D_3$.} By bounding $\| V^{\star}-V_{j-1}^k \|_{\infty} $ with the local error $\| \Delta_{j-1}^k  \|_{\infty}$ (cf. \eqref{eq:vgaptoqgap}) and using $\|P_{j}^k(s,a)\|_{1}\leq 1$, we have
    \begin{align}
    |D_3|
    \le \gamma \sum_{j=\syn(i)+1}^{i} \eta (1-\eta)^{i-j} \|P_{j}^k(s,a)\|_{1} \| V^{\star}-V_{j-1}^k \|_{\infty}  
    \le \gamma \sum_{j=\syn(i)+1}^{i} \eta (1-\eta)^{i-j} \| \Delta_{j-1}^k\|_{\infty} .
    \end{align}
\end{itemize}

By combining the bounds obtained above in \eqref{eq:syncq-error-indiv-decomp},
we obtain the following recursive relation 
\begin{align}
    \|\Delta_{i}^k\|_{\infty}
    \le (1-\eta)^{i-\syn(i)} \| \Delta_{\syn(i)} \|_{\infty} 
    + \underbrace{ \frac{\gamma}{1-\gamma} \sqrt{ \eta \log{\frac{|\cS| |\cA| K T}{\delta}}} }_{\defn \rho}
    + \gamma \sum_{j=\syn(i)+1}^{i} \eta (1-\eta)^{i-j} \| \Delta_{j-1}^k \|_{\infty}.
  \end{align}

By invoking the recursive relation with some algebraic calculations, we obtain the following bound  
\begin{align}
    \|\Delta_{i}^k\|_{\infty} 
    &\le (1-\eta)^{i-\syn(i)} \| \Delta_{\syn(i)} \|_{\infty} + \rho  \cr
    &\quad + \gamma \sum_{j_1=\syn(i)+1}^{i} \eta (1-\eta)^{i-j_1} \left ((1-\eta)^{j_1-1-\syn(i)} \| \Delta_{\syn(i)} \|_{\infty} + \rho + \gamma \sum_{j_2=\syn(i)+1}^{j_1-1} \eta (1-\eta)^{j_1-1-j_2}  \| \Delta_{j_2-1}^k \|_{\infty} \right) \cr
    &= \left((1-\eta)^{i-\syn(i)}  + \gamma \sum_{j_1=\syn(i)+1}^{i} \eta (1-\eta)^{i-1-\syn(i)} \right)\| \Delta_{\syn(i)} \|_{\infty} 
    + \left(1 + \gamma \sum_{j_1=\syn(i)+1}^{i} \eta (1-\eta)^{i-j_1} \right) \rho  \cr
    &\qquad+ \gamma^2 \sum_{j_1=\syn(i)+1}^{i} \sum_{j_2=\syn(i)+1}^{j_1-1} \eta ^2  (1-\eta)^{i-1-j_2}  \| \Delta_{j_2-1}^k \|_{\infty} \cr  
    &\le \left((1-\eta)^{i-\syn(i)}  + \gamma \sum_{j_1=\syn(i)+1}^{i} \eta (1-\eta)^{i-1-\syn(i)} \right)\| \Delta_{\syn(i)} \|_{\infty} 
    + \left(1 + \gamma \sum_{j_1=\syn(i)+1}^{i} \eta (1-\eta)^{i-j_1}  \right) \rho  \cr
    &\qquad+ \gamma^2 \sum_{j_1=\syn(i)+1}^{i} \sum_{j_2=\syn(i)+1}^{j_1-1} \eta ^2  (1-\eta)^{i-1-j_2}  \left ( (1-\eta)^{j_2-1-\syn(i)} \| \Delta_{\syn(i)} \|_{\infty} + \rho  + \cdots \right) \cr
    &\le \left((1-\eta)^{i-\syn(i)}  + \gamma \sum_{j_1=\syn(i)+1}^{i} \eta (1-\eta)^{i-1-\syn(i)} + \cdots + \gamma^l \sum_{j_1=\syn(i)+1}^{i}  \cdots  \sum_{j_l=\syn(i)+1}^{j_{l-1}-1} \eta^l  (1-\eta)^{i-l-\syn(i)}   \right)\| \Delta_{\syn(i)} \|_{\infty} \cr 
    &\qquad + \left(1 + \gamma \sum_{j_1=\syn(i)+1}^{i} \eta (1-\eta)^{i-j_1} + \cdots +  \gamma^l \sum_{j_1=\syn(i)+1}^{i}  \cdots   \sum_{j_l=\syn(i)+1}^{j_{l-1}-1} \eta^l (1-\eta)^{i-l +1 -j_l}\right) \rho  \cr    
    &\qquad+ \gamma^{l+1} \sum_{j_1=\syn(i)+1}^{i} \cdots \sum_{j_{l+1}=\syn(i)+1}^{j_l-1} \eta^{l+1}  (1-\eta)^{i-l-j_{l+1}}  \left ( \| \Delta_{j_{l+1}-1}^k \|\right) \cr        
    &\stackrel{\mathrm{(i)}}{\le} \sum_{l=0}^{i-\syn(i)} \gamma^l \binom{i-\syn(i)}{l} \eta^l (1-\eta)^{i-\syn(i)-l} \| \Delta_{\syn(i)}^k \|_{\infty}
    + \sum_{l=0}^{i-\syn(i)-1} \gamma^l \binom{i-\syn(i)}{l} \eta^l \rho \cr
    &\le ((1-\eta) + \gamma \eta )^{i-\syn(i)} \| \Delta_{\syn(i)}^k \|_{\infty} + (1+\gamma \eta)^{i-\syn(i)} \rho \cr
    &\stackrel{\mathrm{(ii)}}{\le} \| \Delta_{\syn(i)}^k \|_{\infty} + 2 \rho,
\end{align}
where (i) follows from $\Delta_{j_{i-\syn(i)}-1}^k = \Delta_{\syn(i)}^k $ since $j_l \le i-l+1$,
\begin{align*}
\sum_{j_1=\syn(i)+1}^{i} \sum_{j_2=\syn(i)+1}^{j_1-1} \cdots  \sum_{j_l=\syn(i)+1}^{j_{l-1}-1} \eta^l  (1-\eta)^{i-l-\syn(i)} =  \binom{i-\syn(i)}{l} \eta^l  (1-\eta)^{i-l-\syn(i)},\\
\sum_{j_1=\syn(i)+1}^{i}  \cdots  \sum_{j_l=\syn(i)+1}^{j_{l-1}-1} \eta^l (1-\eta)^{i-l +1 -j_l}
\le \sum_{j_1=\syn(i)+1}^{i}  \cdots  \sum_{j_l=\syn(i)+1}^{j_{l-1}-1} \eta^l \le \binom{i-\syn(i)}{l} \eta^l ,
\end{align*}
and (ii) follows from
$(1+\gamma \eta)^{i-\syn(i)} \le (1+\gamma \eta)^{\tau} \le e^{\tau \eta} \le 2$
since $i-\syn(i) \le \tau$ and $\tau \eta \le \frac{1}{2}$. This completes the proof.

%% file: fedasyncqnaive_proof_improved.tex

\section{Proofs for federated asynchronous Q-learning (Section~\ref{sec:fed_asyncQ})}

\subsection{Proof of Lemma~\ref{lemma:dist-asyncwq-multi-visit-mix}}
\label{proof:lemma:dist-asyncwq-multi-visit-mix}
 
To describe the joint probabilistic transitions of $K$ agents formally, we first introduce the following Markov chain $X_t = (X_t^1,\ldots, X_t^K)$, $t=0, 1,\ldots$, where $X_t^k \in \cS\times \cA$ is the state-action pair visited by agent $k$ at time $t$. The joint transition kernel $P$ of $K$ agents is given by
  \begin{align}
    P &\defn \begin{pmatrix}
      P^1 \\
      & P^2\\
      &  & \ddots\\
      &  &  & P^K
    \end{pmatrix} ,
  \end{align}    
  where $P^k$ is the transition kernel of agent $k$, $k=1,\ldots, K$.  Since the agents are independent, the stationary distribution of the joint Markov chain is $\mu$, given by
  \begin{align} \label{eq:def_mu_joint}
     \mu(x) &\defn \prod_{k=1}^K \mu_{\mathsf{b}}^k(x^k),\quad ~~\forall x = (x^1, x^2, \cdots, x^K) \in (\cS \times \cA)^K,  
  \end{align}
  where $\mu_{\mathsf{b}}^k$ denotes the stationary distribution of agent $k$, which are induced by its behavior policy $\pi_{\mathsf{b}}^k$. Next, we define the mixing time of the joint Markov chain as follows:
  \begin{align}
    \tmultimix(\epsilon) &\defn \min \left \{ t ~ \bigg | ~ \sup_{x_0 \in (\cS \times \cA)^K} d_{\mathsf{TV}} (P_{t}(\cdot| x_0),\, \mu ) \le \epsilon  \right \} ~~\text{and}~~ \tmultimix \defn \tmultimix\left(\frac{1}{4}\right),
  \end{align}
  where 
  \begin{equation}\label{eq:def_jointPt}
  P_{t}(\cdot| x_0) = \prod_{k=1}^K P_t^k (\cdot | x_0^k)
  \end{equation} 
  denotes the distribution of the joint state-action pairs of all agents after $t$ transitions starting from $x_0=(x_0^1,\ldots, x_0^K)$. The mixing time of the joint Markov chain can be connected to those of the individual chains via the following relation
   \begin{align} 
    \tmultimix(\epsilon)  & \leq   \max_k \tmix^k(\epsilon /K )     , \qquad \tmix     \le 4\log{8K} \max_{k \in [K]} \tmix^k, \, \label{eq:tmultimix}  
  \end{align}
  which will be proven at the end of the proof.

We now turn to the proof of Lemma~\ref{lemma:dist-asyncwq-multi-visit-mix}.  Define the event
  \begin{align}
    \cB_{u,v}(s,a) \defn \left \{ \left |\sum_{k=1}^K N_{u,v}^k (s,a) - (v-u) \sum_{k=1}^K \mu_{\mathsf{b}}^k(s,a) \right | \ge \frac{1}{2}(v-u) \sum_{k=1}^K \mu_{\mathsf{b}}^k(s,a) \right \}.
  \end{align}
We first establish that
  \begin{align} \label{eq:dist-asyncwq-multi-visit-mix-base}
   \max_{x_0 \in (\cS \times \cA)^K} \mprob \bigg\{ \cB_{u,v}(s,a) \big | \{(s_0^k, a_0^k)\}_{k=1}^K  = x_0 \bigg\}  \le \frac{\delta}{|\cS||\cA|T^2}
  \end{align}
  for any $(s,a) \in \cS \times \cA$ and $1 \le u < v \le T$  provided that
  $u \ge  \tth(s,a) /2$ and $v-u \ge \tth(s,a) /2$.  
  To this end, we decompose the probability into two terms as follows:
  \begin{align*}
    \mprob \bigg \{ \cB_{u,v}(s,a) \big| \{(s_0^k, a_0^k)\}_{k=1}^K  = x_0 \bigg\}
    &= \underbrace{\mprob \bigg\{ \cB_{u,v}(s,a) \big| \{(s_0^k, a_0^k)\}_{k=1}^K \sim \mu \bigg \}}_{=: G_1} \cr
    &\quad + \underbrace{\mprob \bigg \{ \cB_{u,v}(s,a) \big| \{(s_0^k, a_0^k)\}_{k=1}^K  = x_0 \bigg \}
    - \mprob \bigg \{ \cB_{u,v}(s,a) \big| \{(s_0^k, a_0^k)\}_{k=1}^K \sim \mu \bigg \} }_{ =:G_2},
  \end{align*}
  and show each of the terms is bounded by $\frac{\delta}{2|\cS||\cA|T^2}$ for any $x_0 \in (\cS \times \cA)^K$. 
  We shall derive the bounds of these two terms separately.
 
 \paragraph{Step 1: bounding $G_1$.}
      This is for the case that the distribution of the initial state follows the joint stationary distribution.  Since the total number of visits can be written as
  \begin{align*} 
    \sum_{k=1}^K N_{u,v}^k(s,a) = \sum_{k=1}^K \sum_{i=u+1}^v Z_i^k(s,a) = \sum_{i=u+1}^v \bar{Z}_i(s,a),
  \end{align*}
  where
  \begin{equation*}
    Z_i^k(s,a) = 
    \begin{cases}
      1,\quad & \text{if }(s,a) \in (s_{i-1}^k, a_{i-1}^k)\\
      0, & \text{otherwise}
    \end{cases}
    \quad \text{and} \quad
    \bar{Z}_i(s,a) = \sum_{k=1}^K Z_i^k(s,a),
  \end{equation*}
  and
    \begin{align*} 
    \nu_{u,v}(s,a) &\defn \mexp_{(s_0^k, a_0^k) \sim \mu^k \forall k \in [K]} \left [\sum_{i=u+1}^v \bar{Z}_i(s,a) \right ] = (v-u) \sum_{k=1}^K \mu_{\mathsf{b}}^k(s,a) ,
  \end{align*}
  we can invoke Bernstein's inequality for Markov chains \citep[Theorem~3.11]{paulin2015concentration} and obtain
  \begin{align} \label{eq:markov_concentration}
    G_1& = \mprob_{\{(s_0^k, a_0^k)\}_{k=1}^K \sim \mu
      } \left [ \left | \sum_{i=u+1}^v \bar{Z}_i(s,a)  - \nu_{u,v}(s,a) \right | \ge  \frac{1}{2} \nu_{u,v}(s,a) \right ] \cr
    &\le 2 \exp \left ( -\frac{(\nu_{u,v}(s,a)/2)^2 \gamma_{\text{ps}}}{8((v-u)+1/\gamma_{\text{ps}}) V_f + 20 C (\nu_{u,v}(s,a)/2) } \right). 
  \end{align}    
 Here,  $\gamma_{\text{ps}}$ is the pseudo spectral gap satisfying
 \begin{subequations}\label{eq:param_mc}
 \begin{align}
    \gamma_{\text{ps}} &\ge \frac{1}{2 \tmultimix} \label{eq:gammaps} 
    \end{align}  
 for uniformly ergodic Markov chains according to \citet[Proposition~3.4]{paulin2015concentration}. The parameters $C$ and $V_f$ are defined and bounded as follows
  \begin{align}
    C &\defn \max_{u <i \le v} \left | \bar{Z}_i(s,a) - \mexp[\bar{Z}_i(s,a)] \right| \le K, \\
    V_f &\defn \Var (\bar{Z}_i(s,a)) = \sum_{k=1}^K (1-\mu_{\mathsf{b}}^k(s,a)) \mu_{\mathsf{b}}^k(s,a) \le \sum_{k=1}^K \mu_{\mathsf{b}}^k(s,a) .\label{eq:vf}  
  \end{align}
  \end{subequations}
Plugging \eqref{eq:param_mc} into \eqref{eq:markov_concentration}, we have
      \begin{align}
  G_1  &\le 2 \exp \left ( -\frac{(\nu_{u,v}(s,a))^2 }{8\tmultimix (24 (v-u) (\sum_{k=1}^K \mu_{\mathsf{b}}^k(s,a)) + 10 K \nu_{u,v}(s,a)) } \right) \cr
    &\le 2 \exp \left ( -\frac{(v-u) (\sum_{k=1}^K \mu_{\mathsf{b}}^k(s,a)) }{8\tmultimix (24  + 10 K ) } \right)  \le \frac{\delta}{2|\cS||\cA|T^2}, 
  \end{align}
  where the last inequality holds since $(v-u)$ is large enough to satisfy the following condition:  
  \begin{align*}
    v-u
    &\ge \frac{\tth(s,a)}{2} 
    \ge \frac{1088 (\max_{k \in [K]} \tmix^k) \log{8K} \log{\frac{4|\cS||\cA|T^2}{\delta}}}{\frac{1}{K}\sum_{k=1}^K\mu_{\mathsf{b}}^k(s,a)}
      \ge \frac{272 \tmultimix \log{\frac{4|\cS||\cA|T^2}{\delta}}}{\frac{1}{K}\sum_{k=1}^K\mu_{\mathsf{b}}^k(s,a)}.
  \end{align*}

\paragraph{Step 2: bounding $G_2$.}  By the same argument of \citet[Section A.1]{li2021asyncq},  using the fact that the difference caused by the initial state becomes very small after sufficiently long time,  we have
\begin{align}
G_2 & : = \mprob \bigg \{ \cB_{u,v}(s,a) \big| \{(s_0^k, a_0^k)\}_{k=1}^K  = x_0 \bigg \}
    - \mprob \bigg \{ \cB_{u,v}(s,a) \big| \{(s_0^k, a_0^k)\}_{k=1}^K \sim \mu \bigg \}  \nonumber \\
    &\le d_{\mathsf{TV}} (P_{u}(\cdot | x_0), \mu ) \le \frac{\delta}{2|\cS||\cA|T^2},
  \end{align}
where the last inequality holds due to
  \begin{align}
    u
    \ge \frac{\tth(s,a)}{2}
    \ge 4 \log{\frac{4|\cS||\cA|T^2 K}{\delta}}  \max_{k\in [K]} \tmix^k
    \ge \max_{k \in [K]} \tmix^k\left(\frac{\delta}{2|\cS||\cA|T^2 K} \right)
    \ge \tmultimix\left(\frac{\delta}{2|\cS||\cA|T^2} \right).
  \end{align}
Here, the second inequality follows from      
  the fact that $\tmix^k(\epsilon) \le  2\tmix^k \log_2{\frac{2}{\epsilon}}  $ \citep{paulin2015concentration}, and the last inequality follows from \eqref{eq:tmultimix}.
  
\paragraph{Step 3: summing things up.}
By combining the above bound, we complete the proof of \eqref{eq:dist-asyncwq-multi-visit-mix-base}, provided that   $u \ge  \tth(s,a) /2$ and
 $v-u \ge \tth(s,a) $.
Then, we can obtain the following bound for any $(s,a) \in \cS \times \cA$ and $0 \le u <v \le T$:  
  \begin{align}
    &\mprob \left\{ \frac{1}{4}(v-u) \sum_{k=1}^K \mu_{\mathsf{b}}^k(s,a) \le \sum_{k=1}^K N_{u,v}^k (s,a) \le 2(v-u) \sum_{k=1}^K \mu_{\mathsf{b}}^k(s,a) \right \}  \cr 
& \leq   \mprob \left \{ \left |\sum_{k=1}^K N_{u+\frac{\tth(s,a)}{2},v}^k (s,a) - \left(v-u-\frac{\tth(s,a)}{2} \right) \sum_{k=1}^K \mu_{\mathsf{b}}^k(s,a) \right | \ge \frac{1}{2} \left(v-u-\frac{\tth(s,a)}{2} \right) \sum_{k=1}^K \mu_{\mathsf{b}}^k(s,a) \right \} \cr 
    &= \max_{x_0 \in (\cS \times \cA)^K} \mprob \left \{ \cB_{u+\frac{\tth(s,a)}{2},v}(s,a) \bigg | \{(s_0^k, a_0^k)\}_{k=1}^K  = x_0\right \} 
    \le \frac{\delta}{|\cS||\cA|T^2}.
  \end{align} 

\paragraph{Proof of \eqref{eq:tmultimix}.} Notice that by the definition of $d_{\mathsf{TV}}$ and \eqref{eq:def_jointPt}, we have 
    \begin{align*}
d_{\mathsf{TV}} (P_{t}(\cdot| x_0),\, \mu ) \leq \sum_{k=1}^K d_{\mathsf{TV}} (P_{t}^k(\cdot| x_0^k),\,  \mu_{\mathsf{b}}^k ) 
  \end{align*}
for any $x_0 \in (\cS \times \cA)^K$. Hence, setting $t = \max_{k \in [K]}\tmix^k\left(\frac{\epsilon}{K}\right)$, we have
  $$\max_{x_0 \in (\cS \times \cA)^K} d_{\mathsf{TV}} (P_{t}(\cdot| x_0),\, \mu )  \leq \sum_{k=1}^K \frac{\epsilon}{ K} =   \epsilon ,$$
 which immediately implies
    \begin{align*} 
    \tmultimix(\epsilon)  & \leq   \max_k \tmix^k(\epsilon /K )     .
  \end{align*}
  The proof is complete by using the fact that $\tmix(\epsilon) \le 2 \tmix \log_2{\frac{2}{\epsilon}}  $ \citep{paulin2015concentration}, which leads to 
  $$\tmix \leq \max_{k \in [K]} \tmix^k \left (\frac{1}{4K} \right ) \le 4\log{8K} \max_{k \in [K]} \tmix^k.$$

\subsection{Proof of Lemma~\ref{lemma:asyncq-wsum-tighter}}
\label{proof:asyncq-wsum-tighter}
 
  First, \eqref{eq:asyncq-lambda-tighter} is derived as follows:
\begin{align}
  \lambda_{v_1,v_2}(s,a)
  = \frac{1}{K} \sum_{k=1}^K (1-\eta)^{N_{v_1,v_2}^{k}(s,a)} 
     \le \frac{1}{K} \sum_{k=1}^K \exp(-\eta N_{v_1, v_2}^{k}(s,a))
    & \le 1 - \frac{1}{2} \frac{1}{K} \sum_{k=1}^K    \eta N_{v_1, v_2}^{k}(s,a) \cr
    & \le \exp \left (- \frac{\eta}{2K} \sum_{k=1}^K    N_{v_1, v_2}^{k}(s,a) \right )  
\end{align}
using the fact that $1- x \le \exp(-x) \le 1 - \frac{x}{2}$ holds for any $0\le x <1$, and $ \eta N_{h \tau, (h+1)\tau}^{k'}(s,a) \le \eta \tau \le 1$.

  Next, we obtain \eqref{eq:asyncq-wsum-1-tighter} through the following derivation: 
  \begin{align}
    \sum_{k=1}^K \sum_{u \in \cU_{0, t}^k(s,a)} \omega_{u,t}^k (s,a)
    &= \sum_{k=1}^K \sum_{h=0}^{\nsyn(t)-1} \sum_{u \in \cU_{h\tau, (h+1)\tau}^k(s,a)} \omega_{u,t}^k (s,a) \cr
    &= \sum_{h=0}^{\nsyn(t)-1} \left ( \prod_{l = (h+1)}^{\nsyn(t)-1} \lambda_{l\tau, (l+1)\tau}(s,a) \right ) \sum_{k=1}^K \frac{1}{K}  \sum_{u \in \cU_{h\tau, (h+1)\tau}^k(s,a)}    \left ( \eta (1-\eta)^{N_{u+1,(h+1)\tau}^k(s,a)} \right )  \cr
    &\overset{\mathrm{(i)}}{=} \sum_{h=0}^{\nsyn(t)-1} \left ( \prod_{l = (h+1)}^{\nsyn(t)-1} \lambda_{l\tau, (l+1)\tau}(s,a) \right ) \sum_{k=1}^K \frac{1}{K}  (1-(1-\eta)^{N_{h\tau,(h+1)\tau}^k(s,a)})  \cr
    &\overset{\mathrm{(ii)}}{=}   \sum_{h=0}^{\nsyn(t)-1} \left ( \prod_{l = (h+1)}^{\nsyn(t)-1} \lambda_{l\tau, (l+1)\tau}(s,a) \right )  (1- \lambda_{h\tau, (h+1)\tau}(s,a)) \cr
&\overset{\mathrm{(iii)}}{=}  1 -\lambda_{0, \tau} \lambda_{\tau, 2\tau} \cdots \lambda_{ (\nsyn(t)-1) \tau, t} = 1-\omega_{0,t}(s,a),
  \end{align}
  where (i) follows from the geometric sum 
    \begin{align} \label{eq:asyncwq-wsum-1-1-tighter}
  \sum_{u \in \cU_{h\tau, (h+1)\tau}^k(s,a)} \eta (1-\eta)^{N_{u+1, (h+1)\tau}^k(s,a)} 
  &= \eta + \eta (1-\eta) + \cdots + \eta (1-\eta)^{N_{h\tau, (h+1)\tau}^k(s,a)-1} \nonumber \\
  &=    1- (1-\eta)^{N_{h\tau, (h+1)\tau}^k(s,a)}  ,
  \end{align}
  (ii) follows from the definition \eqref{eq:def_lambda_vvv-tighter}, and (iii) follows by cancellation.

  Similarly, \eqref{eq:asyncq-wsum-1-part-tighter} can be obtained with some algebraic calculations as follows:
  \begin{align}
    \sum_{k=1}^K \sum_{u \in \cU_{0, h'\tau}^k(s,a)} \omega_{u,t}^k (s,a)
    &= \sum_{k=1}^K \sum_{h=0}^{h'-1} \sum_{u \in \cU_{h\tau, (h+1)\tau}^k(s,a)} \omega_{u,t}^k (s,a) \cr
    &\overset{\mathrm{(i)}}{=} \sum_{h=0}^{h'-1} \left ( \prod_{l = (h+1)}^{\nsyn(t)-1} \lambda_{l\tau, (l+1)\tau}(s,a) \right )  (1- \lambda_{h\tau, (h+1)\tau}(s,a)) \nonumber \\
    &\overset{\mathrm{(ii)}}{\le}  \lambda_{h'\tau, (h'+1)\tau} \cdots \lambda_{ (\nsyn(t)-1) \tau, t} -\lambda_{0, \tau} \lambda_{\tau, 2\tau} \cdots \lambda_{ (\nsyn(t)-1) \tau, t} \nonumber \\
    &\leq  \lambda_{h'\tau, (h'+1)\tau} \cdots \lambda_{ (\nsyn(t)-1) \tau, t} \overset{\mathrm{(iii)}}{\le}  \prod_{h=h'}^{\nsyn(t)-1} \exp \left (- \frac{\eta}{2K} \sum_{k=1}^K    N_{h\tau, (h+1)\tau}^{k}(s,a) \right ) , \label{eq:proof_akiaki-tighter}
  \end{align}
  where (i) follows from similar derivations as above, (ii) follows by cancellation, and (iii) follows from \eqref{eq:asyncq-lambda-tighter}.

  Finally, \eqref{eq:asyncq-wsum-2-tighter} is derived as follows:
  \begin{align*}
    \sum_{k=1}^K \sum_{u \in \cU_{0, t}^k(s,a)} (\omega_{u,t}^k (s,a) )^2
    &= \sum_{k=1}^K \sum_{h=0}^{\nsyn(t)-1} \sum_{u \in \cU_{h\tau, (h+1)\tau}^k(s,a)} (\omega_{u,t}^k (s,a))^2 \cr
    &= \frac{1}{K}  \sum_{h=0}^{\nsyn(t)-1} \left ( \prod_{l = (h+1)}^{\nsyn(t)-1} \lambda_{l\tau, (l+1)\tau}(s,a) \right )^2 \sum_{k=1}^K \frac{1}{K}  \sum_{u \in \cU_{h\tau, (h+1)\tau}^k(s,a)}    \left ( \eta (1-\eta)^{N_{u+1,(h+1)\tau}^k(s,a)} \right )^2 \cr
    &\stackrel{\mathrm{(i)}}{\le} \frac{2\eta}{K} \sum_{h=0}^{\nsyn(t)-1} \left ( \prod_{l = (h+1)}^{\nsyn(t)-1} \lambda_{l\tau, (l+1)\tau}(s,a) \right ) \sum_{k=1}^K \frac{1}{K} \left(1- (1-\eta)^{(N_{h\tau, (h+1)\tau}^k(s,a))} \right)  \cr
    &= \frac{2\eta}{K} \sum_{h=0}^{\nsyn(t)-1} \left ( \prod_{l = (h+1)}^{\nsyn(t)-1} \lambda_{l\tau, (l+1)\tau}(s,a) \right ) \left(1- \lambda_{h\tau, (h+1)\tau}(s,a) \right) \cr
    &\stackrel{\mathrm{(ii)}}{\le} \frac{2\eta}{K}  ,
  \end{align*}
  where (i) holds since
  \begin{align} 
  \sum_{u \in \cU_{h\tau, (h+1)\tau}^k(s,a)}    \left ( \eta (1-\eta)^{N_{u+1,(h+1)\tau}^k(s,a)} \right )^2 
  &= \eta^2 + \eta^2 (1-\eta)^2 + \cdots + \eta (1-\eta)^{2(N_{u+1,(h+1)\tau}^k(s,a)-1)} \nonumber\\
  &\le \eta \left(1- (1-\eta)^{2 N_{u+1,(h+1)\tau}^k(s,a)} \right) \nonumber\\
  & \leq 2\eta \left(1- (1-\eta)^{  N_{u+1,(h+1)\tau}^k(s,a)} \right)
  \end{align}
and (ii) can be similarly derived to the proof of \eqref{eq:asyncq-wsum-1-part-tighter} (cf.~\eqref{eq:proof_akiaki-tighter}).

\subsection{Proof of Lemma~\ref{lemma:asyncq-error2-freedman-tighter}}
\label{proof:asyncq-error2-freedman-tighter}
Without loss of generality, we prove the claim for some fixed $1 \le t \le T$ and $(s,a) \in \cS \times \cA$. For notation simplicity, let
\begin{align}
    y_{u,t}^k(s,a)  = 
    \begin{cases}
         \omega_{u,t}^k(s,a) (P(s,a) -  P_{u+1}^k(s,a) ) V^k_{u} &\qquad \text{if}~(s_{u}^k, a_{u}^k) = (s,a) \\
         0 &\qquad \text{otherwise}
    \end{cases},
\end{align}
where 
\begin{align}
    \omega_{u,t}^k(s,a) = \frac{\eta}{K}   (1-\eta)^{N_{u+1,(\nsyn(u)+1)\tau}^k(s,a)} \prod_{h = \nsyn(u)+1}^{\nsyn(t)-1} \left(\frac{1}{K} \sum_{k'=1}^K (1-\eta)^{N_{h\tau, (h+1)\tau}^{k'}(s,a)}\right) ,
\end{align}
then $E_t^2(s,a)  = \gamma  \sum_{k=1}^K \sum_{u=0}^{t-1}    y_{u,t}^k(s,a)$. 
However, due to the dependency between $P_{u+1}^k(s,a)$ and $\omega_{u,t}^k(s,a)$ arising from the Markovian sampling, it is difficult to track the sum of $y : =\{ y_{u,t}^k(s,a)\}$ directly. To address this issue, we will first analyze the sum using a collection of approximate random variables $ \widehat{y} = \{ \widehat{y}_{u,t}^k(s,a) \} $ drawn from a carefully constructed set $\widehat{\cY}$, which is closely coupled with the target $\{y_{u,t}^k(s,a)\}_{0\le u <t}$, i.e.,
\begin{align} \label{eq:asyncq_proximity-tighter}
    D(y, \widehat{y}) &\defn  \left | \sum_{k=1}^K \sum_{u=0}^{t-1} \big(  y_{u,t}^k(s,a) - \widehat{y}_{u,t}^k(s,a) \big) \right |  
\end{align}
is sufficiently small. In addition, $ \widehat{y}$ shall exhibit some useful statistical independence and thus easier to control its sum; we shall control this over the entire set $\widehat{\cY}$. Finally, leveraging the proximity above, we can obtain the desired bound on $y$ via triangle inequality. We now provide details on executing this proof outline, where the crust is in designing the set $\widehat{\cY}$ with a controlled size. 

Before describing our construction, let's introduce the following useful event:
\begin{align}\label{eq:avgvisit_bound-tighter}
  \cB_M(s,a) &\defn  \bigcap_{u=0}^{  t-M\tau } \left \{{\frac{1}{4}} \muminavg(s,a) K  M\tau \le \sum_{k=1}^K N_{u,u+M\tau}^{k}(s,a) \le {2} \muminavg(s,a) K M\tau  \right \}, 
\end{align}
where $M= M(s,a):= \lfloor \frac{1}{8 \eta \muminavg(s,a) \tau} \rfloor$.
Note that $M\tau \ge \tau \ge \tth$ (see \eqref{eq:def_tth} for the definition of $\tth(s,a)$), and $1 \le 1/(16 \eta \muminavg(s,a)\tau ) \le  M(s,a) \le 1/(8 \eta \muminavg(s,a)\tau) $ if $\eta \tau \le 1/ 16$.
Then, $\cB_M(s,a)$ holds with probability at least $1 - \frac{\delta}{|\cS||\cA|T}$
according to Lemma~\ref{lemma:dist-asyncwq-multi-visit-mix}.
The rest of the proof shall be carried out under the event $ \cB_M(s,a)$.

\paragraph{Step 1: constructing $\widehat{\cY}$.}
To decouple dependency between $P_{u+1}^k(s,a)$ and $\omega_{u,t}^k(s,a)$, we will introduce approximates of $\omega_{u,t}^k(s,a)$ that only depend on history until $u$ by replacing a factor dependent on future with some constant. 
To gain insight, we first decompose $\omega_{u,t}^k(s,a)$ as follows:
\begin{align*} 
    \omega_{u,t}^k(s,a) 
    &=  \frac{\eta}{K} (1-\eta)^{-N_{\nsyn(u)\tau,u+1}^k(s,a)} 
    \frac{(1-\eta)^{N_{\nsyn(u)\tau,(\nsyn(u)+1)\tau}^k(s,a)}}{\sum_{k'=1}^K(1-\eta)^{N_{\nsyn(u)\tau,(\nsyn(u)+1)\tau}^{k'}(s,a)}}  
    \prod_{h=\nsyn(u)}^{\nsyn(t)-1} \left(\frac{1}{K} \sum_{k'=1}^K (1-\eta)^{N_{h\tau, (h+1)\tau}^{k'}(s,a)}\right) \cr 
    &=  \underbrace{\frac{\eta}{K} (1-\eta)^{-N_{\nsyn(u)\tau,u+1}^k(s,a)}  
    \prod_{h=\nsyn(u)}^{\nsyn(t)-1} \left(\frac{1}{K} \sum_{k'=1}^K (1-\eta)^{N_{h\tau, (h+1)\tau}^{k'}(s,a)}\right)}_{\defn \bar{\omega}_{u,t}^k(s,a)} \cr
    &\quad + \underbrace {\frac{\eta}{K} (1-\eta)^{-N_{\nsyn(u)\tau,u+1}^k(s,a)} 
    \left (\frac{(1-\eta)^{N_{\nsyn(u)\tau,(\nsyn(u)+1)\tau}^k(s,a)}}{\sum_{k'=1}^K(1-\eta)^{N_{\nsyn(u)\tau,(\nsyn(u)+1)\tau}^{k'}(s,a)}} - 1 \right )  
    \prod_{h=\nsyn(u)}^{\nsyn(t)-1} \left(\frac{1}{K} \sum_{k'=1}^K (1-\eta)^{N_{h\tau, (h+1)\tau}^{k'}(s,a)}\right) }_{\defn \chi_{u,t}^k(s,a)}.
\end{align*}
Considering that $\chi_{u,t}^k(s,a)$ can be made small enough, which will be shown in the following step, we analyze the dominant factor $\bar{\omega}_{u,t}^k(s,a)$ in detail as follows:
\begin{align} \label{eq:asyncq-omega-decomp-tighter}
    \bar{\omega}_{u,t}^k(s,a) 
    &=  \prod_{h=h_0(u,t)}^{\nsyn(u)-1} \left ( \left(\frac{1}{K} \sum_{k'=1}^K (1-\eta)^{N_{h\tau, (h+1)\tau}^{k'}(s,a)} \right) \left(\frac{1}{K} \sum_{k'=1}^K (1-\eta)^{N_{h\tau, (h+1)\tau}^{k'}(s,a)}\right )^{-1} \right ) \cr
    &\quad \times \frac{\eta}{K} (1-\eta)^{-N_{\nsyn(u)\tau,u+1}^k(s,a)}  
    \prod_{h=\nsyn(u)}^{\nsyn(t)-1} \left(\frac{1}{K} \sum_{k'=1}^K (1-\eta)^{N_{h\tau, (h+1)\tau}^{k'}(s,a)}\right)  \cr
    &=  \underbrace{ \frac{\eta}{K} (1-\eta)^{-N_{\nsyn(u)\tau,u+1}^k(s,a)} \prod_{h=h_0(u,t)}^{\nsyn(u)-1}   \left(\frac{1}{K} \sum_{k'=1}^K (1-\eta)^{N_{h\tau, (h+1)\tau}^{k'}(s,a)} \right)^{-1} }_{\text{dependent on history until } u} \cr
    &\quad \times \underbrace{\prod_{h=h_0(u,t)}^{\nsyn(t)-1} \left(\frac{1}{K} \sum_{k'=1}^K (1-\eta)^{N_{h\tau, (h+1)\tau}^{k'}(s,a)}\right) }_{\text{dependent on history and future until } t } \cr
    &=  \underbrace{ \frac{\eta}{K} (1-\eta)^{-N_{\nsyn(u)\tau,u+1}^k(s,a)} \prod_{h=h_0(u,t)}^{\nsyn(u)-1}   \left(\frac{1}{K} \sum_{k'=1}^K (1-\eta)^{N_{h\tau, (h+1)\tau}^{k'}(s,a)} \right)^{-1} }_{\defn x_{u}^k(s,a)} \cr
    &\quad \times \prod_{l=1}^{l(u,t)}  \underbrace{\prod_{h=\max\{0, \nsyn(t)- lM\}}^{\nsyn(t)- (l-1)M-1} \left(\frac{1}{K} \sum_{k'=1}^K (1-\eta)^{N_{h\tau, (h+1)\tau}^{k'}(s,a)}\right) }_{\defn z_{l}(s,a) } ,
\end{align}
where  we denote $h_0(u,t) = \max\{0, \nsyn(t)-l(u,t) M\}$, with $ l(u,t) \defn \lceil \frac{(t-u)}{M\tau} \rceil$.

Motivated by the above decomposition, we will construct $\widehat{\cY}$ by approximating the future-dependent parameter $z_{l}(s,a)$ for $1 \le l \le L$, where we define  
\begin{align} \label{def:approx-L}
L \defn \min \left\{ \left\lceil \frac{t}{M\tau} \right\rceil, \lceil 128 \log{(K/\eta)} \rceil\right\} .
\end{align}
We note that $L \le 128 \log{(TK)}$ for $\eta \ge 3/T$. 
Using the fact that $1- x \le \exp(-x) \le 1 - \frac{x}{2}$ holds for any $0\le x <1$, and $ \eta N_{h \tau, (h+1)\tau}^{k'}(s,a) \le \eta \tau \le \frac{1}{2}$, 
\begin{align} \label{eq:asyncq_expbound-tighter}
    \exp \left (- \frac{2 \eta }{K} \sum_{k'=1}^K  N_{h\tau, (h+1)\tau}^{k'}(s,a) \right) 
   &  \le 1 - \frac{\eta }{K} \sum_{k'=1}^K  N_{h\tau, (h+1)\tau}^{k'}(s,a)   \cr
    &\le \frac{1}{K} \sum_{k'=1}^K (1-\eta)^{N_{h\tau, (h+1)\tau}^{k'}(s,a)} \cr
    & \le \frac{1}{K} \sum_{k'=1}^K \exp(-\eta N_{h\tau, (h+1)\tau}^{k'}(s,a)) \cr
    & \le 1 - \frac{1}{2} \frac{1}{K} \sum_{k'=1}^K    \eta N_{h\tau, (h+1)\tau}^{k'}(s,a) \cr
    & \le \exp \left (- \frac{\eta}{2K} \sum_{k'=1}^K    N_{h\tau, (h+1)\tau}^{k'}(s,a) \right ) .  
\end{align}
Therefore, for $1\leq l<L$, under $\cB_{M}(s,a)$, the range of $z_{l}(s,a)$ is bounded as follows:
$$z_{l}(s,a) \in \left [\exp(-4\eta \muminavg(s,a) M \tau ), ~\exp(-\frac{1}{8}\eta \muminavg(s,a) M \tau ) \right ] .$$
Using this property, we construct a set of values that can cover possible realizations of $z_{l}(s,a)$ in a fine-grained manner as follows:
\begin{align}\label{eq:asyncq-zset-tighter}
    \cZ \defn \left \{ \exp\left(-\frac{1}{8}\eta \muminavg(s,a) M \tau - \frac{i \eta}{K} \right) ~~ \Big | i \in \mathbb{Z}: ~~0\le i   < 4 K \muminavg(s,a) M\tau \right \}.
\end{align}
Note that the distance of adjacent elements of $\cZ$  is bounded by $\eta / K e^{- 1/8\eta \muminavg(s,a) M \tau }$, and the size of the set is bounded by $4 K \muminavg(s,a) M\tau $. For $l=L$, because the number of iterations involved in $z_{L}(s,a)$ can be less than $M\tau$, it follows that $z_{L}(s,a) \in \left [\exp(-4\eta \muminavg(s,a) M \tau ), 1 \right ]$. Hence, we construct the set
\begin{align}\label{eq:asyncq-zset0-tighter}
    \cZ_0 \defn \left \{ \exp\left( - \frac{i \eta}{K} \right) ~~ \Big | i \in \mathbb{Z}:  ~~0\le i   <  {4} K \muminavg(s,a) M\tau \right \}.
\end{align}
In sum, we can always find $(\widehat{z}_1, \cdots, \widehat{z}_l, \cdots, \widehat{z}_L) \in \cZ^{L-1} \times \cZ_0$ where its entry-wise distance to $(z_{l}(s,a))_{l \in [L-1]}$ (resp. $z_L(s,a)$) is at most $\eta / K e^{- {1/8}\eta \muminavg(s,a) M \tau }$ (resp. $\eta/K$). 

Moreover, we approximate $x_{u}^k(s,a)$ by clipping it when the accumulated number of visits of all agents is not too large as follows:
\begin{align} \label{eq:asyncq-xclip-tighter}
    \widehat{x}_{u}^k(s,a)  = 
    \begin{cases}
         x_{u}^k(s,a) &\qquad \text{if}~ \sum_{k=1}^K N_{h_0(u,t) \tau, \nsyn(u)\tau}^{k}(s,a) \le  {2} K \muminavg(s,a) M\tau \\
         0 &\qquad \text{otherwise}
    \end{cases}.
\end{align}
Note that the clipping never occurs and $\widehat{x}_{u}^k(s,a) = x_{u}^k(s,a)$ for all $u$ as long as  {$\cB_{M}(s,a)$} holds. To provide useful properties of $\widehat{x}_{u}^k(s,a)$ that will be useful later, we record the following lemma whose proof is provided in Appendix~\ref{proof:asyncq-xsum-tighter}.
\begin{lemma} \label{lemma:asyncq-xsum-tighter} 
For any state-action pair $(s,a) \in \cS \times \cA$, consider any integers $1 \le t \le T$ and $1 \le l \le \lceil \frac{t}{M\tau} \rceil $, where  {$M=\lfloor \frac{1}{8 \eta \muminavg(s,a) \tau} \rfloor$}. Suppose that  {$4\eta \tau \le 1$}, then $\widehat{x}_{u}^k(s,a)$ defined in \eqref{eq:asyncq-xclip-tighter} satisfy
\begin{subequations}
 \begin{align}
    \label{eq:asyncq-xmax-tighter} \forall u \in [h_0, \nsyn(t)-(l-1)M) ~~:~~ \widehat{x}_{u}^k(s,a) &\le \frac{9\eta}{K} , \\
    \label{eq:asyncq-xsum-1-tighter}  \sum_{h=h_0}^{\nsyn(t)-(l-1)M-1} \sum_{u \in \cU_{h\tau, (h+1)\tau}^k(s,a)}  \sum_{k=1}^K \widehat{x}_{u}^k(s,a) &\le  {16}\eta \muminavg(s,a) M\tau, \\
    \label{eq:asyncq-xsum-2-tighter} \sum_{h=h_0}^{\nsyn(t)-(l-1)M-1} \sum_{u \in \cU_{h\tau, (h+1)\tau}^k(s,a)}  \sum_{k=1}^K (\widehat{x}_{u}^k(s,a))^2 & \le \frac{ {64}  \eta^2 \muminavg(s,a) M \tau }{K} ,
\end{align}
where $h_0 = \max\{0, \nsyn(t)-lM\}$.
\end{subequations} 
\end{lemma}

Finally, for each $ \bm{z} = (\widehat{z}_1, \cdots, \widehat{z}_L) \in \cZ^{L-1} \times \cZ_0$, setting 
\begin{align} \label{def:approx-omega}
\widehat{\omega}_{u,t}^k(s,a; \bm{z}) = \widehat{x}_{u}^k(s,a) \prod_{l=1}^{l(u,t)} \widehat{z}_l,
\end{align}
an approximate random sequence $\widehat{y}_{\bm{z}} = \{\widehat{y}_{u,t}^k(s,a; \bm{z})\}_{0\le u <t}$ can be constructed as follows:
\begin{align} \label{def:asyncq-widehaty-tighter}
    \widehat{y}_{u,t }^k(s,a ; \bm{z})  = 
    \begin{cases}
         \widehat{\omega}_{u,t}^k(s,a ; \bm{z}) (P(s,a) -  P_{u+1}^k(s,a) ) V^k_{u} &\qquad \text{if}~(s_{u}^k, a_{u}^k) = (s,a) ~\text{and}~ l(u,t) \le L \\
         0 &\qquad \text{otherwise}
    \end{cases}.
\end{align}
If $t> LM\tau$, for any $u < t-LM\tau$, i.e., $ l(u,t) > L$, we set $\widehat{y}_{u,t}^k(s,a;\bz) = 0$ since the magnitude of $\omega_{u,t}^k(s,a)$ becomes negligible when the time difference between $u$ and $t$ is large enough, and the fine-grained approximation using $\cZ$ is no longer needed, as shall be seen momentarily. 
Finally, denote a collection of the approximates induced by $\cZ^{L-1} \times \cZ_0$ as 
$$\widehat{\cY} = \{\widehat{y}_{ \bm{z}}: \quad \bm{z} \in \cZ^{L-1} \times \cZ_0 \}.$$

\paragraph{Step 2: bounding the approximation error $D(y, \widehat{y}_{\bm{z}} )$.}
We now show that under  {$\cB_{M}(s,a)$},  there exists $\widehat{y}_{ \bm{z}}: =\widehat{y}_{ \bm{z}(y)} \in \widehat{\cY}$ such that
\begin{equation}\label{eq:asyncq_peach-tighter}
D(y, \widehat{y}_{\bm{z}} ) < \frac{ {525}}{1-\gamma} \sqrt{ \frac{ \Csim \eta L }{K} \log{\frac{4|\cS||\cA|T^2}{\delta}}} 
\end{equation}
with at least probability $1-2\delta$.
To this end, we first decompose the approximation error as follows:
\begin{align*}
  &  \min_{\widehat{y}_{\bm{z}} \in \widehat{\cY}} D(y, \widehat{y}_{\bm{z}} ) \cr
   &  = \min_{\bm{z} \in \cZ^{L-1} \times \cZ_0} \left | \sum_{k=1}^K \sum_{u=0}^{t-1} \left( y_{u,t}^k(s,a) - \widehat{y}_{u,t}^k(s,a; \bm{z} ) \right) \right | \cr
   &\le \max_{\bm{z} \in \cZ^{L-1} \times \cZ_0} \left | \sum_{k=1}^K \sum_{u=0}^{t-LM\tau-1}  y_{u,t}^k(s,a) - \widehat{y}_{u,t}^k(s,a; \bm{z}) \right | 
   + \min_{\bm{z} \in \cZ^{L-1} \times \cZ_0} \left | \sum_{k=1}^K \sum_{u=t-LM\tau}^{t-1}  y_{u,t}^k(s,a) - \widehat{y}_{u,t}^k(s,a; \bm{z}) \right |  \cr
   &\le \underbrace{\max_{\bm{z} \in \cZ^{L-1} \times \cZ_0} \left | \sum_{k=1}^K \sum_{u=0}^{t-LM\tau-1}  y_{u,t}^k(s,a) - \widehat{y}_{u,t}^k(s,a; \bm{z}) \right | }_{ =: D_1} \cr
   &\quad + \underbrace{ \min_{\bm{z} \in \cZ^{L-1} \times \cZ_0} \left | \sum_{k=1}^K \sum_{u=t-LM\tau}^{t-1}  (\bar{\omega}_{u,t}^k(s,a) - \widehat{\omega}_{u,t}^k(s,a; \bm{z}) )  (P(s,a) -  P_{u+1}^k(s,a) ) V^k_{u} \right | }_{=: D_2} \cr
   &\quad + \underbrace{  \left | \sum_{k=1}^K \sum_{u=t-LM\tau}^{t-1}  \chi_{u,t}^k(s,a)  (P(s,a) -  P_{u+1}^k(s,a) ) V^k_{u} \right | }_{=: D_3} ,
\end{align*}
and will bound each term separately.
\begin{itemize}
    \item \textbf{Bounding $D_1$.}
    This term appears only when $t>LM\tau$. Since $\widehat{y}_{u,t}^k(s,a; \bz) =0$ for all $u < t-LM\tau$ regardless of $\bm{z}$ by construction, 
\begin{align*}
    \left | \sum_{k=1}^K \sum_{u=0}^{t-LM\tau-1}  y_{u,t}^k(s,a) - \widehat{y}_{u,t}^k(s,a; \bz) \right |
    &\le \sum_{k=1}^K \sum_{  u \in \cU_{0, t-LM\tau}^k(s,a)} \omega_{u,t}^k(s,a)  \| P(s,a) -  P_{u+1}^k(s,a) \|_{1} \| V^k_{u} \|_{\infty } \cr
    &\stackrel{\mathrm{(i)}}{\le} \frac{2}{1-\gamma} \sum_{k=1}^K \sum_{ u \in \cU_{0, t-LM\tau}^k(s,a)} \omega_{u,t}^k(s,a)   \cr
    &\le \frac{2}{1-\gamma} \prod_{h=\nsyn(t)-LM}^{\nsyn(t)-1} \left (\frac{1}{K} \sum_{k'=1}^K (1-\eta)^{N_{h\tau, (h+1)\tau}^{k'}(s,a)} \right) \cr
    &\stackrel{\mathrm{(ii)}}{\le} \frac{2}{1-\gamma} \exp\left ( - \frac{\eta}{2K} \sum_{k'=1}^K    N_{t-LM\tau, t}^{k'}(s,a) \right) \cr
    &\stackrel{\mathrm{(iii)}}{\le} \frac{2}{1-\gamma} \exp\left ( - {\frac{1}{8}}\eta  \muminavg(s,a) LM\tau \right)  \cr
    &\stackrel{\mathrm{(iv)}}{\le} \frac{2\eta}{(1-\gamma)K},
\end{align*}
where (i) holds since $\|P(s,a)\|_1,~\|P_{u}^k(s,a)\|_1 \le 1$ and $\|V_{u-1}^k\|_{\infty} \le \frac{1}{1-\gamma}$ (cf.~\eqref{eq:bounded_iterates}), (ii) follows from \eqref{eq:asyncq_expbound-tighter}, (iii) holds due to {$\cB_{M}(s,a)$}, and (iv) holds because {$L \ge  128 \log{\frac{K}{\eta}}  \ge \frac{8}{\eta \muminavg(s,a) M\tau} \log{\frac{K}{\eta}}$} given that {$\eta \muminavg(s,a) M\tau \ge 1/16$.}

    \item \textbf{Bounding $D_2$.}
    Since $\widehat{x}_{u}^k(s,a) = x_{u}^k(s,a)$  when  {$\cB_{M}(s,a)$} holds, in view of \eqref{def:asyncq-widehaty-tighter}, we have
    \begin{align*}
        &\min_{\bm{z} \in \cZ^{L-1} \times \cZ_0} \left | \sum_{k=1}^K \sum_{u=t-LM\tau}^{t-1}  (\bar{\omega}_{u,t}^k(s,a) - \widehat{\omega}_{u,t}^k(s,a; \bm{z}) )  (P(s,a) -  P_{u+1}^k(s,a) ) V^k_{u} \right | \cr
        & \le \min_{\bm{z} \in \cZ^{L-1} \times \cZ_0}  \sum_{k=1}^K \sum_{u \in \cU_{t-LM\tau, t}^k(s,a)} \big|\bar{\omega}_{u,t}^k(s,a) - \widehat{\omega}_{u,t}^k(s,a;\bz) \big|  \, \| P(s,a) -  P_{u+1}^k(s,a) \|_{1} \| V^k_{u} \|_{\infty } \cr
        &\le \frac{2}{1-\gamma} \min_{\bm{z} \in \cZ^{L-1} \times \cZ_0} \left( \sum_{l=1}^L \sum_{h=\nsyn(t)-lM}^{\nsyn(t)-(l-1)M-1} \sum_{u \in \cU_{h\tau, (h+1)\tau}^k(s,a)}  \sum_{k=1}^K \widehat{x}_{u}^k(s,a) \left |\prod_{l'=1}^{l} z_{l'}(s,a) - \prod_{l'=1}^{l} \widehat{z}_{l'} \right | 
    \right)  ,
    \end{align*}
where the last inequality holds since $\|P(s,a)\|_1,~\|P_{u}^k(s,a)\|_1 \le 1$ and $\|V_{u-1}^k\|_{\infty} \le \frac{1}{1-\gamma}$ (cf.~\eqref{eq:bounded_iterates}) and the definition of $\widehat{\omega}_{u,t}^k(s,a;\bz)$ defined in \eqref{def:approx-omega}.

Note that for any given $\{z_{l}(s,a)\}_{l \in [L]}$, under {$\cB_{M}(s,a)$}, there exists $\widehat{\bz}^{\star} = (\widehat{z}_1^{\star}, \ldots, \widehat{z}_l^{\star}, \ldots, \widehat{z}_L^{\star}) \in \cZ^{L-1} \times \cZ_0$ such that $|\widehat{z}_l^{\star} - z_{l}(s,a)| \le \frac{\eta}{K} \exp(-{1/8}\eta \muminavg(s,a) M \tau )$ for $l<L$ and $|\widehat{z}_L^{\star} - z_{L}(s,a)| \le \frac{\eta}{K} $. Also, recall that $z_l(s,a), ~ \widehat{z}_l^{\star} \le \exp(-{1/8}\eta \muminavg(s,a) M \tau ) $ for $l < L$ and $z_L(s,a), ~ \widehat{z}_L^{\star} \le 1$. Then, for any $l \le L$ it follows that:
\begin{align*}
     \left |\prod_{l'=1}^{l} z_{l'}(s,a) - \prod_{l'=1}^{l} \widehat{z}_{l'}^{\star} \right | 
    &\le \Big ( 
    \Big |\prod_{l'=1}^{l} z_{l'}(s,a) - \widehat{z}_{1}^{\star} \prod_{l'=2}^{l} z_{l'}(s,a) \Big | 
     + \cdots + \Big |  z_{l} \prod_{l'=1}^{l-1} \widehat{z}_{l'}^{\star} - \prod_{l'=1}^{l} \widehat{z}_{l'}^{\star} \Big |
     \Big ) \cr
     &\le \exp \Big (-{\frac{1}{8}}(l-1)\eta  \muminavg(s,a) M \tau \Big ) \sum_{l'=1}^l \frac{\eta}{K}   \cr
    &\le \exp \Big (-{\frac{1}{8}}(l-1) \eta  \muminavg(s,a) M \tau \Big ) \frac{L\eta}{K} .
\end{align*}
Then, applying the above bound and \eqref{eq:asyncq-xsum-1-tighter} in Lemma~\ref{lemma:asyncq-xsum-tighter},
\begin{align*}
    D_2
    &\le \frac{2}{1-\gamma}  \sum_{l=1}^L \sum_{h=\nsyn(t)-lM}^{\nsyn(t)-(l-1)M-1} \sum_{u \in \cU_{h\tau, (h+1)\tau}^k(s,a)}  \sum_{k=1}^K \widehat{x}_{u}^k(s,a) \left |\prod_{l'=1}^{l} z_{l'}(s,a) - \prod_{l'=1}^{l} \widehat{z}_{l'}^{\star} \right |  
     \cr
    &\le \frac{2}{1-\gamma}\frac{L\eta}{K}  \sum_{l=1}^L \exp \Big (- {\frac{1}{8}} (l-1) \eta  \muminavg(s,a) M \tau \Big )  \sum_{h=\nsyn(t)-lM}^{\nsyn(t)-(l-1)M-1} \sum_{u \in \cU_{h\tau, (h+1)\tau}^k(s,a)}  \sum_{k=1}^K \widehat{x}_{u}^k(s,a)
      \cr
    &\le \frac{2 }{1-\gamma}\frac{L\eta}{K}  \frac{1}{1-\exp(- {1/8}\eta  \muminavg(s,a) M \tau )}  ( {16}\eta \muminavg(s,a) M\tau) \cr
    &\stackrel{\mathrm{(i)}}{\le}  \frac{2 }{1-\gamma}\frac{L\eta}{K} \frac{ 16}{\eta  \muminavg(s,a) M \tau}  16\eta \muminavg(s,a) M\tau  
    \le \frac{ 512  \eta L}{(1-\gamma) K },
\end{align*}
where (i) holds since { $\eta  \muminavg(s,a) M \tau /8  \le 1 $ } and $e^{-x} \le 1-\frac{1}{2}x$ for any $0\le x \le 1$.

    \item \textbf{Bounding $D_3$.} 
    Applying Freedman's inequality, we can obtain the following bound, whose proof is provided in Appendix~\ref{proof:asyncq-error2-D3-tighter}.
    \begin{lemma}\label{lemma:asyncq-error2-D3-tighter}
        Consider any $\delta  \in (0,1)$ and $L$ defined in \eqref{def:approx-L}. For any $(s,a)  \in \cS \times \cA$ and $1 \le t \le T$, the following holds:
      \begin{align}
         D_3 
        \le \frac{ {9}}{1-\gamma} \sqrt{ \frac{ \Csim \eta L}{K} \log{\frac{4|\cS||\cA|T^2}{\delta}}}
      \end{align}
      with probability at least $1-2\delta$, as long as {$\tau \ge \tth$}, and {$\eta \le \min\{\frac{1}{4 \tau K} , \frac{1}{ K \Csim L \log{\frac{4|\cS||\cA|T^2}{\delta}}} \}$}.
    \end{lemma}

\end{itemize}

By combining the bounds obtained above,
\begin{align*}
    \min_{\widehat{y}_{\bm{z}} \in \widehat{\cY}} D(y, \widehat{y}_{\bm{z}} )
   &\le \frac{2\eta}{(1-\gamma)K}
   + \frac{{512} \eta L}{(1-\gamma)K} 
   + \frac{9}{1-\gamma} \sqrt{ \frac{ \Csim \eta L }{K}  \log{\frac{4|\cS||\cA|T^2}{\delta}}} \cr
   &\le \frac{ {525}}{1-\gamma} \sqrt{ \frac{ \Csim \eta L }{K} \log{\frac{4|\cS||\cA|T^2}{\delta}}} 
\end{align*}
since {$\eta \le \frac{K}{128 \log{(TK)}} \le K/L$} due to $L \le 128 \log{(TK)}$.

\paragraph{Step 3: concentration bound over $\cY$.} We now show that for all elements  in $\widehat{\cY} = \{\widehat{y}_{ \bm{z}}: \; \bm{z} \in \cZ^{L-1} \times \cZ_0 \}$ satisfy
 \begin{align} \label{eq:asyncq_banana_peel-tighter}
 \left |\sum_{k=1}^K \sum_{u=0}^{t-1}  \widehat{y}_{u,t}^k(s,a;\bz) \right |< \frac{115}{(1-\gamma)} \sqrt{\frac{\eta L}{K} \log{\frac{4|\cS||\cA|T^2 K}{\delta}}} 
\end{align}
with probability at least $1- \frac{\delta}{|\cS||\cA|T}$. It suffices to establish \eqref{eq:asyncq_banana_peel-tighter} for a fixed $\bm{z}   \in \cZ^{L-1} \times \cZ_0$ with probability at least $1- \frac{\delta}{|\cS||\cA|T|\cY|}$, where
\begin{align}\label{eq:asyncq_size_cY-tighter}
|\widehat{\cY}| = |\cZ^{L-1} \times \cZ_0 | \le {(4K \muminavg(s,a) M\tau )^{L} \le (K/\eta)^{L} \le (TK)^L}
\end{align}
because {$\eta \muminavg(s,a) M\tau \le 1/4$ and $\eta \ge 1/T$}.

For any fixed $\bm{z} = (\widehat{z}_1, \cdots, \widehat{z}_L) \in \cZ^{L-1} \times \cZ_0$, since $\widehat{\omega}_{u,t}^k(s,a; \bz) = \widehat{x}_{u}^k(s,a) \prod_{l=1}^{l(u,t)} \widehat{z}_l$ only depends on the events happened until $u$, which is independent to a transition at $u+1$.  Thus, we can apply Freedman's inequality to bound the sum of $\widehat{y}_{u,t}^k(s,a; \bz)$ since
\begin{align}
    \mexp[ \widehat{y}_{u,t}^k(s,a; \bz) | \mathcal{Y}_{u}] = 0,
\end{align}
where $\cY_u$ denotes the history of visited state-action pairs and updated values of all agents until $u$, i.e.,
$\cY_{u} = \{ (s_{v}^k, a_{v}^k), V_{v}^{k} \}_{k \in [K], v \le {u}}$. 
Before applying Freedman's inequality, we need to calculate the following quantities. First,
  \begin{align} 
      B_t(s,a)
      \defn \max_{k \in [K], 0 \le u < t} | \widehat{y}_{u,t}^k(s,a; \bz) | 
      \le \widehat{x}_{u}^k(s,a) \prod_{l=1}^{l(u,t)} \widehat{z}_l \| P(s,a) -  P_{u+1}^k(s,a) \|_{1} \| V^k_{u} \|_{\infty } 
      \le \frac{{18} \eta}{(1-\gamma)K} ,
    \end{align}
    where the last inequality follows from $\|P(s,a)\|_1,~\|P_{u}^k(s,a)\|_1 \le 1$,  $\|V_{u-1}^k\|_{\infty} \le \frac{1}{1-\gamma}$ (cf.~\eqref{eq:bounded_iterates}), $\hat{z}_l \le 1$, and \eqref{eq:asyncq-xmax-tighter} in Lemma~\ref{lemma:asyncq-xsum-tighter}. Next, we can bound the variance as
    \begin{align} 
      W_t(s,a)
      &\defn  \sum_{u=0}^{t} \sum_{k=1}^K \mexp [(\widehat{y}_{u,t}^k(s,a; \bz))^2 | \cY_{u} ] \cr
      & = \sum_{l=1}^L \sum_{h=\max\{0, \nsyn(t)-lM\}}^{\nsyn(t)-(l-1)M-1} \sum_{k=1}^K \sum_{u \in \cU_{h\tau, (h+1)\tau}^k(s,a)}  (\widehat{x}_{u}^k(s,a) \prod_{l'=1}^{l} \widehat{z}_{l'} )^2 \Var_{P(s,a)} (V_{u}^k) \cr
      & \stackrel{\mathrm{(i)}}{\le} \frac{2}{(1-\gamma)^2} \sum_{l=1}^L \left( \prod_{l'=1}^{l} \widehat{z}_{l'}^2 \right) \sum_{h=max\{0, \nsyn(t)-lM\}}^{\nsyn(t)-(l-1)M-1} \sum_{k=1}^K \sum_{u \in \cU_{h\tau, (h+1)\tau}^k(s,a)}  (\widehat{x}_{u}^k(s,a) )^2  \cr  
      & \stackrel{\mathrm{(ii)}}{\le} \frac{2}{(1-\gamma)^2} \sum_{l=1}^L \left( \prod_{l'=1}^{l} \widehat{z}_{l'}^2 \right) \frac{ {64}  \eta^2 \muminavg(s,a) M \tau }{K}   \cr   
      & \stackrel{\mathrm{(iii)}}{\le} \frac{{128} \eta^2 \muminavg(s,a) M \tau}{K(1-\gamma)^2} \sum_{l=1}^L \exp\left(- {1/4} (l-1) \eta  \muminavg(s,a) M \tau \right)   \cr   
      &\le \frac{{128} \eta^2 \muminavg(s,a) M \tau}{K(1-\gamma)^2} \frac{1}{1-\exp(- {1/4}\eta  \muminavg(s,a) M \tau )}  \cr
      &\stackrel{\mathrm{(iv)}}{\le} \frac{{128} \eta^2 \muminavg(s,a) M \tau}{K(1-\gamma)^2} \frac{ {8}}{\eta  \muminavg(s,a) M \tau} 
      = \frac{ {1024}\eta}{K(1-\gamma)^2} =:\sigma^2,
    \end{align}
    where (i) holds due to the fact that
    $\|\mathsf{Var}_{P}(V)\|_{\infty} \le \|P\|_1 (\|V\|_{\infty})^2 + (\|P\|_1 \|V\|_{\infty})^2 \le \frac{2}{(1-\gamma)^2}$
    because  $\|V\|_{\infty} \le \frac{1}{1-\gamma}$ (cf.~\eqref{eq:bounded_iterates}) and $\|P\|_1 \le 1$, 
    (ii) follows from \eqref{eq:asyncq-xsum-2-tighter} in Lemma~\ref{lemma:asyncq-xsum-tighter}, (iii) holds due to the range of $\cZ$ and $\cZ_0$ is bounded by $\exp(- {1/8}\eta  \muminavg(s,a) M \tau )$ and $1$, respectively, and (iv) holds since $e^{-x} \le 1-\frac{1}{2}x$ for any $0\le x \le 1$ and { $   \eta  \muminavg(s,a) M \tau /4  \le 1 $ }. 

Now, by substituting the above bounds of $W_t$ and $B_t$ into Freedman's inequality (see Theorem~\ref{thm:Freedman}) and setting $m=1$, it follows that for any $s\in \cS$, $a \in \cA$, $t\in [T]$ and $\widehat{y}_{\bm{z}} \in \widehat{\cY}$,  
\begin{align} \label{eq:asyncq-error2a-tighter}
  \left | \sum_{k=1}^K \sum_{u=0}^{t-1} \widehat{y}_{u,t}^k(s,a; \bz) \right |
  &\le \sqrt{8 \max{\{W_t(s,a), \frac{\sigma^2}{2^m} \}} \log{\frac{4m|\cS||\cA|T |\widehat{\cY}|}{\delta}}} + \frac{4}{3} B_t(s,a) \log{\frac{4m|\cS||\cA|T |\widehat{\cY}|}{\delta}} \cr
  &\le \sqrt{ {8192} \frac{\eta}{K(1-\gamma)^2} \log{\frac{4|\cS||\cA|T|\widehat{\cY}|}{\delta}}} + \frac{{24} \eta}{K(1-\gamma)} \log{\frac{4|\cS||\cA|T|\widehat{\cY}|}{\delta}} \cr
  &\stackrel{\mathrm{(i)}}{\le} \frac{ {115}}{(1-\gamma)} \sqrt{\frac{\eta  L}{K}  \log{\frac{4|\cS||\cA|T^2 K}{\delta}}} ,
\end{align}
with at least probability $1-\frac{\delta}{|\cS||\cA|T|\widehat{\cY}|}$, where (i) holds because $|\widehat{\cY}| \le (TK)^L$ (cf. \eqref{eq:asyncq_size_cY-tighter}), and $ {\frac{\eta L}{K} \log{\frac{4|\cS||\cA|T^2 K}{\delta}} \le 1}$ when  {$L \le 128 \log{(TK)}$ and $\eta \le \frac{K}{128 \log{(TK)} \log{\frac{4|\cS||\cA|T^2 K}{\delta}}}$}. Therefore, it follows that \eqref{eq:asyncq_banana_peel-tighter} holds. 

\paragraph{Step 4: putting things together.}
We now putting all the results obtained in the previous steps together to achieve the claimed bound.
Under {$\cB_{M}(s,a)$}, there exists $\widehat{y}_{ \bm{z}}: =\widehat{y}_{ \bm{z}(y)} \in \widehat{\cY}$ such that \eqref{eq:asyncq_peach-tighter} holds. Hence,
\begin{align} 
  \sum_{k=1}^K \sum_{u=0}^{t-1}    y_{u,t}^k(s,a) & \leq   \left | \sum_{k=1}^K \sum_{u=0}^{t-1} \widehat{y}_{u,t}^k(s,a; \bz) \right |  + D(y, \widehat{y}_{\bm{z}} )  \nonumber \cr
  & \leq \frac{115}{(1-\gamma)} \sqrt{\frac{\eta L }{K} \log{\frac{4|\cS||\cA|T^2 K}{\delta}}} +  \frac{525}{1-\gamma} \sqrt{\frac{ \Csim  \eta L }{K} \log{(TK)} \log{\frac{4|\cS||\cA|T^2}{\delta}}} \cr
  & \leq \frac{7241  }{(1-\gamma)} \sqrt{\frac{\Csim \eta }{K} \log{(TK)} \log{\frac{4|\cS||\cA|T^2 K}{\delta}}},
\end{align}
where the second line holds due to  \eqref{eq:asyncq_banana_peel-tighter} and \eqref{eq:asyncq_peach-tighter}, and the last line holds because {$L \le 128 \log{(TK)}$}. By taking a union bound over all $(s,a) \in \cS \times \cA$ and $t \in [T]$, we complete the proof.

\subsubsection{Proof of Lemma~\ref{lemma:asyncq-xsum-tighter}}
\label{proof:asyncq-xsum-tighter}

For notational simplicity, let $\overline{h}$ be the largest integer among $h \in (h_0, \nsyn(t)-(l-1)M)$ such that
\begin{align} \label{eq:asyncq-hclip-def-tighter}
     \sum_{k=1}^K N_{h_0 \tau, (h-1)\tau}^{k}(s,a) \le {2} K \muminavg(s,a) M\tau .
\end{align}
Then, the following holds:
\begin{align} \label{eq:asyncq-clipped-numvisit-tighter}
    \sum_{k=1}^K N_{h_0 \tau, \overline{h}\tau}^{k}(s,a) 
    &=  \sum_{k=1}^K N_{(\overline{h}-1)\tau, \overline{h}\tau}^{k}(s,a) + \sum_{k=1}^K N_{h_0 \tau, (\overline{h}-1)\tau}^{k}(s,a) \cr
    &\le K \tau + {2} K \muminavg(s,a) M\tau .
\end{align}
Also, for the following proofs, we provide an useful bound as follows:
\begin{align} \label{eq:asyncq_x_numerator_bound-tighter}
        \sum_{k'=1}^K \frac{(1-\eta)^{-N_{h \tau, (h+1)\tau}^{k'}(s,a)}}{K} 
        \le  \frac{ \sum_{k'=1}^K e^{\eta N_{h \tau, (h+1)\tau}^{k'}(s,a)} }{K}
        &\le 1+ 2 \eta \frac{\sum_{k'=1}^K  N_{h \tau, (h+1)\tau}^{k'}(s,a) }{K} \cr
        &\le \exp \left (2 \eta \frac{\sum_{k'=1}^K  N_{h \tau, (h+1)\tau}^{k'}(s,a) }{K} \right),
\end{align}
which holds since $1+x \le e^x \le 1+2x$ for any $x \in [0,1]$ and {$\eta N_{h \tau, (h+1)\tau}^{k'}(s,a) \le \eta \tau \le 1$}.

According to \eqref{eq:asyncq-xclip-tighter}, for any integer $u \in [\overline{h} \tau, t-(l-1)M\tau) $, $\widehat{x}_{u}^k(s,a)$ is clipped to zero. Now, we prove the bounds in Lemma~\ref{lemma:asyncq-xsum-tighter} respectively. 

\paragraph{Proof of \eqref{eq:asyncq-xmax-tighter}.} For $u \in [h_0\tau, \overline{h} \tau)$,
    \begin{align*} 
    \widehat{x}_{u}^k(s,a) 
    &= \frac{\eta}{K} (1-\eta)^{-N_{\nsyn(u)\tau,u+1}^k(s,a)} \prod_{h=h_0(u,t)}^{\nsyn(u)-1}   \left(\frac{1}{K} \sum_{k'=1}^K (1-\eta)^{N_{h\tau, (h+1)\tau}^{k'}(s,a)} \right)^{-1} \cr
    &\stackrel{\mathrm{(i)}}{\le} \frac{3\eta}{K} \prod_{h=h_0(u,t)}^{\nsyn(u)-1}   \left(\frac{1}{K} \sum_{k'=1}^K (1-\eta)^{N_{h\tau, (h+1)\tau}^{k'}(s,a)} \right)^{-1} \cr
    &\stackrel{\mathrm{(ii)}}{\le} \frac{3\eta}{K}  \exp \left(   \frac{2 \eta}{K} \sum_{k'=1}^K  N_{h_0 \tau, (\overline{h}-1)\tau}^{k'}(s,a) \right) \cr
    &\stackrel{\mathrm{(iii)}}{\le} \frac{3\eta}{K}  \exp \left(   4 \eta \muminavg(s,a) M \tau \right) 
     \stackrel{\mathrm{(iv)}}{\le}  \frac{9\eta}{K} ,
    \end{align*}
    where (i) holds since $(1+\eta)^x \le e^{\eta x}$ and {$\eta N_{\nsyn(u) \tau , u+1}^k(s,a) \le \eta \tau \le 1$}, (ii) holds due to \eqref{eq:asyncq_expbound-tighter} and the fact that $\nsyn(u) \le \overline{h}-1$, (iii) follows from the condition of $\overline{h}$ in \eqref{eq:asyncq-hclip-def-tighter}, and (iv) holds because {$4 \eta \muminavg(s,a) M\tau \le 1$}.
    
\paragraph{Proof of \eqref{eq:asyncq-xsum-1-tighter}.} By the definition of $\overline{h}$, it follows that 
    \begin{align*}
        \sum_{h=h_0}^{\nsyn(t)-(l-1)M-1} \sum_{u \in \cU_{h\tau, (h+1)\tau}^k(s,a)}  \sum_{k=1}^K \widehat{x}_{u}^k(s,a)
        = \sum_{h=h_0}^{\overline{h}-1} \sum_{u \in \cU_{h\tau, (h+1)\tau}^k(s,a)}  \sum_{k=1}^K x_{u}^k(s,a) .
    \end{align*}
    Using the following relation for each $h$:
    \begin{align*}
        & \sum_{u \in \cU_{h\tau, (h+1)\tau}^k(s,a)}  \sum_{k=1}^K x_{u}^k(s,a) \cr
        &=  \frac{1}{K} \left ( \sum_{k=1}^K  \sum_{u \in \cU_{h\tau, (h+1)\tau}^k(s,a)}  \eta (1-\eta)^{-N_{\nsyn(u)\tau,u+1}^k(s,a)} \right )
        \prod_{h'=h_0}^{h-1} \left(\frac{1}{K} \sum_{k'=1}^K (1-\eta)^{N_{h'\tau, (h'+1)\tau}^{k'}(s,a)} \right)^{-1}  \cr
        &=   \left ( \frac{1}{K} \sum_{k=1}^K  (1-\eta)^{-N_{h\tau, (h+1)\tau}^k(s,a) } -1  \right )
        \prod_{h'=h_0}^{h-1} \left(\frac{1}{K} \sum_{k'=1}^K (1-\eta)^{N_{h'\tau, (h'+1)\tau}^{k'}(s,a)} \right)^{-1}  \cr
        &\le   \left ( \frac{1}{K} \sum_{k=1}^K  (1-\eta)^{-N_{h\tau, (h+1)\tau}^k(s,a) } -1  \right )
        \prod_{h'=h_0}^{h-1} \left(\frac{1}{K} \sum_{k=1}^K  (1-\eta)^{-N_{h'\tau, (h'+1)\tau}^k(s,a) } \right)  ,
    \end{align*}
    where the last inequality follows from Jensen's inequality, and applying \eqref{eq:asyncq_x_numerator_bound-tighter}, we can complete the proof as follows:
    \begin{align*}
        \sum_{h=h_0}^{\overline{h}-1} \sum_{u \in \cU_{h\tau, (h+1)\tau}^k(s,a)}  \sum_{k=1}^K x_{u}^k(s,a)
        &\le \prod_{h'=h_0}^{\overline{h}-1} \left(\frac{1}{K} \sum_{k=1}^K  (1-\eta)^{-N_{h'\tau, (h'+1)\tau}^k(s,a) } \right) -1 \cr
        &\le \exp \left (\frac{2\eta \sum_{k'=1}^K  N_{h_0\tau, \overline{h}\tau}^{k'}(s,a)}{K} \right ) -1 \cr
        &\stackrel{\mathrm{(i)}}{\le} \exp \left ( {4}\eta \muminavg(s,a) M\tau + 2\eta \tau \right )   - 1  \cr
        &\stackrel{\mathrm{(ii)}}{\le} {16}\eta \muminavg(s,a) M\tau,        
    \end{align*}
    where (i) follows from \eqref{eq:asyncq-clipped-numvisit-tighter}, and (ii) holds because $e^x \le 1+2x$ for any $x \in [0,1]$, $2 \eta \tau \le 1/2$, and $4 \eta \muminavg(s,a) M\tau \le 1/2$.

\paragraph{Proof of \eqref{eq:asyncq-xsum-2-tighter}.}
Similarly,
    \begin{align*}
        \sum_{h=h_0}^{\nsyn(t)-(l-1)M-1} \sum_{u \in \cU_{h\tau, (h+1)\tau}^k(s,a)}  \sum_{k=1}^K (\widehat{x}_{u}^k(s,a))^2
        = \sum_{h=h_0}^{\overline{h}-1} \sum_{u \in \cU_{h\tau, (h+1)\tau}^k(s,a)}  \sum_{k=1}^K (x_{u}^k(s,a) )^2.
    \end{align*}
    Using the following relation for each $h$:
    \begin{align*}
        & \sum_{u \in \cU_{h\tau, (h+1)\tau}^k(s,a)}  \sum_{k=1}^K (x_{u}^k(s,a) )^2 \cr
        &=  \frac{1}{K^2} \left ( \sum_{k=1}^K  \sum_{u \in \cU_{h\tau, (h+1)\tau}^k(s,a)}  \eta^2 (1-\eta)^{-2N_{\nsyn(u)\tau,u+1}^k(s,a)} \right )
        \prod_{h'=h_0}^{h-1} \left(\frac{1}{K} \sum_{k'=1}^K (1-\eta)^{N_{h'\tau, (h'+1)\tau}^{k'}(s,a)} \right)^{-2}  \cr
        &\le  \frac{\eta }{K} \left ( \frac{1 }{K} \sum_{k=1}^K  (1-\eta)^{-2N_{h\tau, (h+1)\tau}^k(s,a) } -1  \right )
        \prod_{h'=h_0}^{h-1} \left(\frac{1}{K} \sum_{k'=1}^K (1-\eta)^{N_{h'\tau, (h'+1)\tau}^{k'}(s,a)} \right)^{-2}  \cr
        &\le   \frac{\eta }{K} \left ( \frac{1 }{K} \sum_{k=1}^K  (1-\eta)^{-2N_{h\tau, (h+1)\tau}^k(s,a) } -1  \right )
        \prod_{h'=h_0}^{h-1} \left(\frac{1 }{K} \sum_{k=1}^K  (1-\eta)^{-2N_{h'\tau, (h'+1)\tau}^k(s,a) } \right) ,
    \end{align*}
    where the last inequality follows from Jensen's inequality, and applying \eqref{eq:asyncq_x_numerator_bound-tighter} under the condition {$2\eta \tau \le 1$}, we can complete the proof as follows:
    \begin{align*} 
    \sum_{h=h_0}^{\overline{h}-1} \sum_{u \in \cU_{h\tau, (h+1)\tau}^k(s,a)}  \sum_{k=1}^K (x_{u}^k(s,a) )^2
    &\le \frac{\eta }{K} \prod_{h'=h_0}^{\overline{h}-1} \left(\frac{1}{K} \sum_{k=1}^K  (1-\eta)^{-2N_{h'\tau, (h'+1)\tau}^k(s,a) } \right) -1  \cr
    &\le \frac{\eta}{K}  \left ( \exp \left (4 \eta \frac{\sum_{k'=1}^K  N_{h_0 \tau, \overline{h}\tau}^{k'}(s,a) }{K} \right )   
       - 1 \right ) \cr
    &\stackrel{\mathrm{(i)}}{\le} \frac{\eta}{K}  \left ( \exp \left ({8} \eta \muminavg(s,a) M \tau + 4\eta \tau \right )   
       - 1 \right ) \cr   
    &\stackrel{\mathrm{(ii)}}{\le} \frac{ {64}  \eta^2 \muminavg(s,a) M \tau }{K}  ,
    \end{align*}
    where (i) follows from \eqref{eq:asyncq-clipped-numvisit-tighter}, and (ii) holds because $e^x \le 1+4x$ for any $x \in [0,2]$, $4\eta \tau\le 1$, and $8 \eta \muminavg(s,a) M\tau  \le 1$.

\subsubsection{Proof of Lemma~\ref{lemma:asyncq-error2-D3-tighter}}
\label{proof:asyncq-error2-D3-tighter}

Recall that
    \begin{align*}
        \chi_{u,t}^k(s,a) 
        &=  \frac{\eta}{K} (1-\eta)^{-N_{\nsyn(u)\tau,u+1}^k(s,a)} 
    \left (\frac{(1-\eta)^{N_{\nsyn(u)\tau,(\nsyn(u)+1)\tau}^k(s,a)}}{\sum_{k'=1}^K(1-\eta)^{N_{\nsyn(u)\tau,(\nsyn(u)+1)\tau}^{k'}(s,a)}} - 1 \right )  
    \prod_{h=\nsyn(u)}^{\nsyn(t)-1} \left(\frac{1}{K} \sum_{k'=1}^K (1-\eta)^{N_{h\tau, (h+1)\tau}^{k'}(s,a)}\right) \cr
    &= \left (\frac{(1-\eta)^{N_{\nsyn(u)\tau,(\nsyn(u)+1)\tau}^k(s,a)}}{\sum_{k'=1}^K(1-\eta)^{N_{\nsyn(u)\tau,(\nsyn(u)+1)\tau}^{k'}(s,a)}} - 1 \right )  \omega_{u,t}^k(s,a).
    \end{align*}
    We can observe that $\chi_{u,t}^k(s,a)$ and $\omega_{u,t}^k(s,a)$ are solely determined by the number of visits of agents during local steps, i.e., $(N_{h\tau,(h+1)\tau}^{k}(s,a))_{k \in [K], h \in [\nsyn(t)-LM,\nsyn(t)-1]} $. It thus suffice to consider $\{\chi_{u,t}^k(s,a; \bm{N})\}_{0\le u < t, k \in [K]}$ and $\{\omega_{u,t}^k(s,a; \bm{N})\}_{0\le u < t, k \in [K]}$ constructed with each of the possible combinations of number of visits for all $k \in [K]$ and $h \in [\nsyn(t)-LM,\nsyn(t)-1]$ , i.e.,
    $\bm{N} \in [0, \tau]^{KLM}$. 
    Then, by setting $X={ 9\sqrt{ \frac{ \Csim \eta L}{K(1-\gamma)^2} \log{\frac{4|\cS||\cA|T^2}{\delta}}} }$ and taking an union bound,
    \begin{align*}
        &\mprob \left[ \left | \sum_{k=1}^K \sum_{u=t-LM\tau}^{t-1}  \chi_{u,t}^k(s,a)  (P(s,a) -  P_{u+1}^k(s,a) ) V^k_{u} \right | \ge X \right]\cr
        &= \sum_{\bm{N} \in [0, \tau]^{KLM}} \mprob  \left[ \left | \sum_{k=1}^K \sum_{u=t-LM\tau}^{t-1}  \chi_{u,t}^k(s,a)  (P(s,a) -  P_{u+1}^k(s,a) ) V^k_{u} \right | \ge X , \chi_{u,t}^k(s,a) = \chi_{u,t}^k(s,a; \bm{N}) \right]\cr        
        &\le \sum_{\bm{N} \in [0, \tau]^{KLM}} \mprob  \left[ \left | \sum_{k=1}^K \sum_{u=t-LM\tau}^{t-1}  \chi_{u,t}^k(s,a; \bm{N})  (P(s,a) -  P_{u+1}^k(s,a) ) V^k_{u} \right | \ge X \right] ,
    \end{align*}
    and it suffices to show that 
    \begin{align*}
        \mprob  \left[ \left | \sum_{k=1}^K \sum_{u=t-LM\tau}^{t-1}  \chi_{u,t}^k(s,a; \bm{N})  (P(s,a) -  P_{u+1}^k(s,a) ) V^k_{u} \right | \ge  X  \right] \le \frac{\delta}{|\cS||\cA| T (1+\tau)^{KLM} }.
    \end{align*}

    Since $\chi_{u,t}^k(s,a; \bm{N})$ is a constant, which does not depend on $P_{u+1}^k(s,a)$, 
    \begin{align}
    \mexp[ \chi_{u,t}^k(s,a; \bm{N})  (P(s,a) -  P_{u+1}^k(s,a) ) V^k_{u} | \mathcal{Y}_{u}] = 0,
    \end{align}
    where $\cY_u$ denotes the history of visited state-action pairs and updated values of all agents until $u$, i.e.,
    $\cY_{u} = \{ (s_{v}^k, a_{v}^k), V_{v}^{k} \}_{k \in [K], v \le {u}}$, and thus, we can apply Freedman's inequality to bound the sum.

    Before applying Freedman's inequality, we need to calculate the following quantities. First,
    \begin{align*} 
      B_t(s,a)
      &\defn \max_{k \in [K], t-LM\tau \le u < t} |\chi_{u,t}^k(s,a; \bm{N})  (P(s,a) -  P_{u+1}^k(s,a) ) V^k_{u} | \cr
      &\le \max_{k \in [K], t-LM\tau \le u < t} \left | 1 -  \frac{\frac{1}{K} \sum_{k' =1}^K (1-\eta)^{N_{\nsyn(u)\tau,(\nsyn(u)+1)\tau}^{k'}(s,a)} }{(1-\eta)^{N_{\nsyn(u)\tau,(\nsyn(u)+1)\tau}^k(s,a)}} \right | \omega_{u,t}^k(s,a; \bm{N}) \| P(s,a) -  P_{u+1}^k(s,a) \|_{1} \| V^k_{u} \|_{\infty } \cr
      &\stackrel{\mathrm{(i)}}{\le} \frac{2}{1-\gamma}\max_{k \in [K], t-LM\tau \le u < t} \left | 1 -  \frac{\frac{1}{K} \sum_{k' =1}^K (1-\eta)^{N_{\nsyn(u)\tau,(\nsyn(u)+1)\tau}^{k'}(s,a)} }{(1-\eta)^{N_{\nsyn(u)\tau,(\nsyn(u)+1)\tau}^k(s,a)}} \right | \omega_{u,t}^k(s,a; \bm{N}) \cr
      &\stackrel{\mathrm{(ii)}}{\le} \frac{ 8 \eta  \mumax(s,a) \tau }{1-\gamma}\max_{k \in [K], t-LM\tau \le u < t}  \omega_{u,t}^k(s,a; \bm{N})  
      \stackrel{\mathrm{(iii)}}{\le} \frac{ 8 \eta^2  \mumax(s,a) \tau }{(1-\gamma)K},
    \end{align*}
    where (i) holds because $\|P(s,a)\|_1,~\|P_{u}^k(s,a)\|_1 \le 1$,  $\|V_{u-1}^k\|_{\infty} \le \frac{1}{1-\gamma}$ (cf.~\eqref{eq:bounded_iterates}), (ii) follows from the fact that (which will be shown at the end of the proof)
    \begin{align} \label{eq:asyncq_avgmaxgap-tighter}
        \left | 1 -  \frac{\frac{1}{K} \sum_{k' =1}^K (1-\eta)^{N_{\nsyn(u)\tau,(\nsyn(u)+1)\tau}^{k'}(s,a)} }{(1-\eta)^{N_{\nsyn(u)\tau,(\nsyn(u)+1)\tau}^k(s,a)}} \right |
        &\le 4 \eta  \mumax(s,a) \tau ,
    \end{align}
    with $\mumax(s,a) \defn \max_{k} \mu^k_{\mathsf{b}}(s,a)$,
    and (iii) holds due to the fact that $\omega_{u,t}^k(s,a; \bm{N}) \le \frac{\eta}{K}$.

    Next, we can bound the variance as
    \begin{align*} 
      W_t(s,a)
      &\defn  \sum_{u=\max\{0, t-LM\tau \}}^{t-1} \sum_{k=1}^K \mexp \Big[\Big(\chi_{u,t}^k(s,a; \bm{N})  (P(s,a) -  P_{u+1}^k(s,a) ) V^k_{u} \Big)^2 | \cY_{u} \Big] \cr
      & \stackrel{\mathrm{(i)}}{\le} ( {4} \eta  \mumax(s,a) \tau)^2  \sum_{h=\max\{0, \nsyn(t)-LM\}}^{\nsyn(t)-1} \sum_{u \in \cU_{h\tau, (h+1)\tau}^k(s,a)}  
        \sum_{k=1}^K  \Big (  \omega_{u,t}^k(s,a; \bm{N}) \Big )^2 \Var_{P(s,a)} (V_{u}^k) \cr
      & \stackrel{\mathrm{(ii)}}{\le} \frac{2 ( {4} \eta  \mumax(s,a) \tau)^2}{(1-\gamma)^2}  \sum_{h=\max\{0, \nsyn(t)-LM\}}^{\nsyn(t)-1}   \sum_{u \in \cU_{h\tau, (h+1)\tau}^k(s,a)} \sum_{k=1}^K \Big  (  \omega_{u,t}^k(s,a; \bm{N}) \Big )^2   \cr  
      & \stackrel{\mathrm{(iii)}}{\le} \frac{2 ( {4} \eta  \mumax(s,a) \tau)^2}{(1-\gamma)^2}  \frac{6\eta}{K} =:\sigma^2,
    \end{align*}
    where (i) follows from \eqref{eq:asyncq_avgmaxgap-tighter}, (ii) holds due to the fact that
    $\|\mathsf{Var}_{P}(V)\|_{\infty} \le \|P\|_1 (\|V\|_{\infty})^2 + (\|P\|_1 \|V\|_{\infty})^2 \le \frac{2}{(1-\gamma)^2}$
    because  $\|V\|_{\infty} \le \frac{1}{1-\gamma}$ (cf.~\eqref{eq:bounded_iterates}) and $\|P\|_1 \le 1$, 
    (iii) follows from \eqref{eq:asyncq-wsum-2-tighter} in Lemma~\ref{lemma:asyncq-wsum-tighter}.

    Now, by substituting the above bounds of $W_t$ and $B_t$ into Freedman's inequality (see Theorem~\ref{thm:Freedman}) and setting $m=1$, it follows that for any $s\in \cS$, $a \in \cA$, $t\in [T]$ and $\bm{N} = (N_{h\tau,(h+1)\tau}^{k}(s,a))_{k \in [K], h \in [\nsyn(t)-LM,\nsyn(t)-1]} \in [0, \tau]^{KLM}$,  
    \begin{align} 
      &\left | \sum_{k=1}^K \sum_{u=t-LM\tau}^{t-1}  \chi_{u,t}^k(s,a; \bm{N})  (P(s,a) -  P_{u+1}^k(s,a) ) V^k_{u} \right | \cr
      &\le \sqrt{8 \max{\{W_t(s,a), \frac{\sigma^2}{2^m} \}} \log{\frac{4m|\cS||\cA|T  (1+\tau)^{KLM}}{\delta}}} + \frac{4}{3} B_t(s,a) \log{\frac{4m|\cS||\cA|T  (1+\tau)^{KLM}}{\delta}} \cr
      &\le \sqrt{ 96 \frac{( {4} \eta  \mumax(s,a) \tau)^2 \eta  }{K(1-\gamma)^2} \log{\frac{4|\cS||\cA|T (1+\tau)^{KLM}}{\delta}}} + \frac{ 12 \eta^2  \mumax(s,a) \tau }{K(1-\gamma)} \log{\frac{4|\cS||\cA|T (1+\tau)^{KLM}}{\delta}} \cr
      &\le \sqrt{ 384 \frac{ (4 \eta \tau K) (  \mumax(s,a)^2 \eta  M \tau ) L \eta  }{K(1-\gamma)^2} \log{\frac{4|\cS||\cA|T (1+\tau)}{\delta}}} + \frac{12 L \eta  (\mumax(s,a)  \eta M \tau) }{(1-\gamma)} \log{\frac{4|\cS||\cA|T (1+\tau)}{\delta}} \cr
      &\stackrel{\mathrm{(i)}}{\le} \sqrt{ 48 \frac{ \Csim L \eta }{K(1-\gamma)^2} \log{\frac{4|\cS||\cA|T (1+\tau)}{\delta}}} + \frac{2 \Csim L \eta   }{(1-\gamma)} \log{\frac{4|\cS||\cA|T (1+\tau)}{\delta}} \cr
      & \stackrel{\mathrm{(ii)}}{\le} 9 \sqrt{ \frac{ \Csim \eta L}{K(1-\gamma)^2} \log{\frac{4|\cS||\cA|T^2}{\delta}}}
    \end{align}
    with at least probability $1-\frac{\delta}{|\cS||\cA|T (1+\tau)^{KLM}}$, where we invoke the definition of $\Csim$ (cf. \eqref{eq:def_csim}). Here, (i) holds because {$\eta \tau K \le 1/4$} and {$\mumax(s,a)  \eta M \tau \le \Csim \muminavg(s,a)  \eta M \tau \le \frac{\Csim}{8} $}, and (ii) follows from the fact that
    {$\eta \le \frac{1}{128  K \Csim \log{(TK)} \log{\frac{4|\cS||\cA|T^2}{\delta}}} \le \frac{1}{ K \Csim L \log{\frac{4|\cS||\cA|T^2}{\delta}}} $}.
    
    \paragraph{ Proof of \eqref{eq:asyncq_avgmaxgap-tighter}.}
    Using the fact that for $0 < \eta < 1$, 
    \begin{align*}
        (1-\eta)^{-n} \le e^{\eta n} \le 1+2 \eta n ~~\text{if}~~   n \ge 0 ~~\text{and}~~ \eta n \le 1,
        ~~\text{and}~~
        (1-\eta)^n  \ge 1-\eta n   ~~\text{if}~~   n \le 0 ~\text{or} ~ n \ge 1,
    \end{align*}
    we can obtain the bounds as follows:
    \begin{align*}
        1 - \frac{\eta}{K} \sum_{k' =1}^K  N_{\nsyn(u)\tau,(\nsyn(u)+1)\tau}^{k'}(s,a) 
        \le \frac{1}{K} \sum_{k' =1}^K (1-\eta)^{N_{\nsyn(u)\tau,(\nsyn(u)+1)\tau}^{k'}(s,a) }
        &\le \frac{\frac{1}{K} \sum_{k' =1}^K (1-\eta)^{N_{\nsyn(u)\tau,(\nsyn(u)+1)\tau}^{k'}(s,a)} }{(1-\eta)^{N_{\nsyn(u)\tau,(\nsyn(u)+1)\tau}^k(s,a)}} \cr
        &\le (1-\eta)^{ - N_{\nsyn(u)\tau,(\nsyn(u)+1)\tau}^k(s,a) }  \cr
        &\le 1 + 2 \eta N_{\nsyn(u)\tau,(\nsyn(u)+1)\tau}^k(s,a) .
    \end{align*}
    Thus, recalling $\mumax(s,a) \defn \max_{k} \mu^k_{\mathsf{b}}(s,a)$, and using the fact that for any $(s,a,k,u) \in \cS \times \cA \times [K] \times [T]$:
    \begin{align*}
      N_{\nsyn(u)\tau,(\nsyn(u)+1)\tau}^k(s,a) \le
      2 \mumax(s,a) \tau 
    \end{align*}
    at least with probability $1-\delta$, as long as {$ \tau \ge 443 \left( \frac{ \tmix^k }{\mumax (s,a)} \right) \log{\frac{4|\cS||\cA|T K}{\delta}}$}, which naturally holds if {$\tau \ge \tth $} (see \eqref{eq:def_tth} for the definition of $\tth$), according to Lemma~\ref{lemma:Bernstein-state-occupancy},
    \begin{align*}
        \left | 1 -  \frac{\frac{1}{K} \sum_{k' =1}^K (1-\eta)^{N_{\nsyn(u)\tau,(\nsyn(u)+1)\tau}^{k'}(s,a)} }{(1-\eta)^{N_{\nsyn(u)\tau,(\nsyn(u)+1)\tau}^k(s,a)}} \right |
        &\le 2 \eta \max\Big  \{ N_{\nsyn(u)\tau,(\nsyn(u)+1)\tau}^k(s,a), \frac{1}{K} \sum_{k' =1}^K  N_{\nsyn(u)\tau,(\nsyn(u)+1)\tau}^{k'}(s,a) \Big \} \cr
        &\le {4} \eta  \mumax(s,a) \tau.
    \end{align*}

\subsection{Proof of Lemma~\ref{lemma:asyncvq-e3}}
\label{proof:asyncvq-e3}
For any $t \ge \beta\tau$, the error term can be decomposed as follows:
\begin{align} \label{eq:asyncvq-e3-decomp}
  E_t^3(s,a)
  &= \gamma \sum_{k=1}^K \sum_{u \in \cU_{0, t}^k(s,a)} \omega_{u,t}^k(s,a)  P(s,a) (V^{\star} - V^k_{u}) \cr
  &= \underbrace{ \gamma \sum_{k=1}^K \sum_{u \in \cU_{0, (\nsyn(t)-\beta)\tau}^k(s,a)} \omega_{u,t}^k(s,a)  P(s,a) (V^{\star} - V^k_{u})  }_{=: E_t^{3a}(s,a)} \cr
  & \qquad\qquad + \underbrace{  \gamma \sum_{k=1}^K \sum_{u \in \cU_{(\nsyn(t)-\beta)\tau, t}^k(s,a)} \omega_{u,t}^k(s,a)  P(s,a) (V^{\star} - V^k_{u})  }_{=: E_t^{3b}(s,a)} .
\end{align}

We shall these two terms separately.
\begin{itemize}
    \item \textbf{Bounding $E_t^{3a}(s,a)$.} First, the bound on $E_t^{3a}(s,a)$ is derived as follows:  
\begin{align}
  |E_t^{3a}(s,a)|
  &\le \gamma \sum_{k=1}^K \sum_{u \in \cU_{0, (\nsyn(t)-\beta)\tau}^k(s,a)} \omega_{u,t}^k(s,a) \| P(s,a) \|_{1} \| (V^{\star} - V^k_{u}) \|_{\infty}\cr
  &\stackrel{\mathrm{(i)}}{\le} \frac{2\gamma}{1-\gamma} \sum_{k=1}^K \sum_{u \in \cU_{0, (\nsyn(t)-\beta)\tau}^k(s,a)}  \omega_{u,t}^k(s,a) \cr
  &\stackrel{\mathrm{(ii)}}{\le} \frac{2\gamma}{1-\gamma} \exp\left (-\frac{\eta}{2K} \sum_{k=1}^K N_{(\nsyn(t)-\beta)\tau,t}^k(s,a)\right) \cr
  &\stackrel{\mathrm{(iii)}}{\le} \frac{2\gamma}{1-\gamma} \exp\left (-\frac{\eta \muminavg \beta \tau}{8}   \right) ,
\end{align}
where (i) holds because $\|V^k_{u}\|_{\infty}, \|V^{\star}\|_{\infty} \le \frac{1}{1-\gamma}$ (cf. \eqref{eq:bounded_iterates}) and $\|P(s,a)\|_1 \le 1,$ (ii) holds due to \eqref{eq:asyncq-wsum-1-part-tighter} in Lemma~\ref{lemma:asyncq-wsum-tighter}, and (iii) follows from the fact that $ \sum_{k=1}^K N_{(\nsyn(t)-\beta)\tau,t}^k(s,a) \ge \frac{ K \muminavg \beta \tau}{ 4}$ according to Lemma~\ref{lemma:dist-asyncwq-multi-visit-mix} as long as {$\beta \tau \ge \tth$.}

\item \textbf{Bounding $E_t^{3b}(s,a)$.}  Next, we bound $E_t^{3b}(s,a)$ as follows:
\begin{align} \label{eq:asyncvq-e3b-decomp}
  |E_t^{3b}(s,a)|
  &\le \gamma \sum_{k=1}^K \sum_{u \in \cU_{(\nsyn(t)-\beta)\tau, t}^k(s,a)} \omega_{u,t}^k(s,a)  \left \|V^{\star} - V^k_{u} \right \|_{\infty} \cr
  &\stackrel{\mathrm{(i)}}{\le} \gamma \sum_{k=1}^K \sum_{h=\nsyn(t)-\beta}^{\nsyn(t) - 1} \sum_{u \in \cU_{h\tau, (h+1)\tau}^k(s,a)} \omega_{u,t}^k(s,a)  (\| \Delta_{h\tau} \|_{\infty} + \| Q^k_{u} - Q^k_{h\tau} \|_{\infty}) \cr
  &\stackrel{\mathrm{(ii)}}{\le} \gamma \sum_{k=1}^K \sum_{h=\nsyn(t)-\beta}^{\nsyn(t) - 1} \sum_{u \in \cU_{h\tau, (h+1)\tau}^k(s,a)} \omega_{u,t}^k(s,a)  ((1+2\eta \tau)\| \Delta_{h\tau} \|_{\infty} + \localvar) 
\end{align}
where (i) follows from the following bound, which will be shown in Appendix~\ref{proof:asyncq-vgap-indiv-bound}, 
\begin{align} \label{eq:asyncq-vgap-indiv-bound}
    \| V^{\star} - V^k_{u} \|_{\infty}
    &\le \| \Delta_{\syn(u)}^k \|_{\infty} + \| Q^k_{u} - Q^k_{\syn(u)} \|_{\infty},
\end{align}
and (ii) holds due to the following lemma.
\begin{lemma} \label{lemma:asyncq-tdgap-indiv} 
Assume {$ \eta \tau \le \frac{1}{2}$}. For any given $\delta \in (0,1)$,  the following holds for any $k \in [K]$ and $0 \le u < T$:
  \begin{align} \label{eq:asyncq-tdgap-indiv-bound}
    \|Q^k_{u} - Q^k_{\syn(u)} \|_{\infty} 
    \le 2\eta \tau \|\Delta_{\syn(u)}^k \|_{\infty} 
    +  \frac{8 \gamma \eta  \sqrt{\tau-1}}{1-\gamma}\sqrt{\log{\frac{2|\cS||\cA|T K }{\delta}}} 
  \end{align}
  with probability at least $1-\delta$.
\end{lemma}
Here, for notation simplicity, we denote $\localvar \defn \frac{8 \gamma \eta  \sqrt{\tau-1}}{1-\gamma}\sqrt{\log{\frac{2|\cS||\cA|T K }{\delta}}} $.  

Then, with some algebraic calculations, we can obtain the bound on $E_t^{3b}(s,a)$ as follows:
\begin{align}
    |E_t^{3b}(s,a)|
    &\stackrel{\mathrm{(i)}}{\le} \localvar + \gamma  \sum_{h=\nsyn(t)-\beta}^{\nsyn(t) - 1} (1 + 2 \eta \tau) \| \Delta_{h\tau} \|_{\infty}  \sum_{k=1}^K \sum_{u \in \cU_{h\tau, (h+1)\tau}^k(s,a)} \omega_{u,t}^k(s,a)  \cr
  &\stackrel{\mathrm{(ii)}}{\le} \localvar + \frac{1+\gamma}{2} \max_{\nsyn(t)-\beta \le h < \nsyn(t)}\| \Delta_{h\tau} \|_{\infty}  \sum_{k=1}^K \sum_{h=\nsyn(t)-\beta}^{\nsyn(t) - 1} \sum_{u \in \cU_{h\tau, (h+1)\tau}^k(s,a)} \omega_{u,t}^k(s,a)  \cr
  &\stackrel{\mathrm{(iii)}}{\le} \localvar + \frac{1+\gamma}{2} \max_{\nsyn(t)-\beta \le h < \nsyn(t)}\| \Delta_{h\tau} \|_{\infty}  ,
\end{align}
where (i) holds according to \eqref{eq:asyncq-wsum-1-tighter} of Lemma~\ref{lemma:asyncq-wsum-tighter}, 
(ii) holds when $\eta$ is small enough that {$\eta \le \frac{1-\gamma}{4\gamma \tau}$}, and (iii) follows from \eqref{eq:asyncq-wsum-1-tighter} of Lemma~\ref{lemma:asyncq-wsum-tighter}.

\end{itemize}

Now we have the bounds of $E_t^{3a}(s,a)$ and $E_t^{3b}(s,a)$ separately obtained above. By combining the bounds in \eqref{eq:asyncvq-e3-decomp}, we can claim the advertised bound, which completes the proof.

\subsubsection{Proof of \eqref{eq:asyncq-vgap-indiv-bound}}
\label{proof:asyncq-vgap-indiv-bound}
We prove the claim by showing
  \begin{align}
    \Delta_{\syn(u)}^k(s,a_{\syn(u)}^k(s)) - d_{\syn(u),u}^k(s,a^{\star}(s))
    \le V^{\star}(s) - V^k_{u}(s)
    \le \Delta_{\syn(u)}^k(s, a^{\star}(s))
    - d_{\syn(u),u}^k(s,a^{\star}(s))
  \end{align}
  for any $s \in \cS$.
  The upper bound is derived as follows:
  \begin{align}
    V^{\star}(s) - V_u^k(s)
    &= Q^{\star}(s, a^{\star}(s)) - Q^k_{u}(s, a_u^k(s)) \cr
    &\le Q^{\star}(s, a^{\star}(s)) - Q^k_{u}(s, a^{\star}(s)) \cr
    &= Q^{\star}(s, a^{\star}(s))
      - Q^k_{\syn(u)}(s, a^{\star}(s)) - \underbrace{ (Q^k_{u}(s, a^{\star}(s)) - Q^k_{\syn(u)}(s, a^{\star}(s))) }_{d_{\syn(u), u}^k(s,a^{\star}(s))}       
  \end{align}
  using the fact that
  $Q_u^k(s, a_u^k(s)) \ge Q_u^k(s, a^{\star}(s)).$
  Similarly, the lower bound is obtained as follows:
  \begin{align}
    V^{\star}(s) - V_u^k(s)
    &= Q^{\star}(s, a^{\star}(s)) - Q_u^k(s, a_u^k(s)) \cr
    &= Q^{\star}(s, a^{\star}(s)) - Q^k_{\syn(u)}(s, a_{\syn(u)}^k(s)) + Q^k_{\syn(u)}(s, a_{\syn(u)}^k(s)) - Q_u^k(s, a_u^k(s)) \cr
    &= Q^{\star}(s, a^{\star}(s)) - Q^k_{\syn(u)}(s, a_{\syn(u)}^k(s)) + Q^k_{\syn(u)}(s, a_{\syn(u)}^k(s)) - Q^k_{\syn(u)}(s, a_u^k(s)) -  d_{\syn(u), u}^k(s,a_u^k(s)) \cr
    &\ge Q^{\star}(s, a_{\syn(u)}^k(s)) - Q^k_{\syn(u)}(s, a_{\syn(u)}^k(s)) + Q^k_{\syn(u)}(s, a_{\syn(u)}^k(s)) - Q^k_{\syn(u)}(s, a_u^k(s)) - d_{\syn(u), u}^k(s,a_u^k(s))   \cr
    &\ge Q^{\star}(s, a_{\syn(u)}^k(s)) - Q^k_{\syn(u)}(s, a_{\syn(u)}^k(s)) - d_{\syn(u), u}^k(s,a_u^k(s))
  \end{align}
  using the fact that
  $Q^{\star}(s, a_{\syn(u)}^k(s)) \le Q^{\star}(s, a^{\star}(s))$
  and $Q^k_{\syn(u)}(s, a_{\syn(u)}^k(s)) \ge Q^k_{\syn(u)}(s, a_u^k(s)).$ 
 
\subsubsection{Proof of Lemma~\ref{lemma:asyncq-tdgap-indiv}}
 For any $0 \le u < T$, $k \in [K]$, and $(s,a) \in \cS \times \cA$, we can write the bound as
\begin{align} \label{eq:asyncq-td-decomp-2}
    |Q^k_{u} (s,a) - Q^k_{\syn(u)} (s,a) |
    &\le \underbrace{  2 \eta \sum_{v \in \cU_{\syn(u), u}^k(s,a)} \|\Delta_{v}^k\|_{\infty} }_{\defn B_1 } 
    + \underbrace{ \left |\gamma \eta \sum_{v \in \cU_{\syn(u), u}^k(s,a)} (P_{v+1}^k (s,a) - P(s,a)) V^{\star} \right|  }_{\defn B_2}.
\end{align}
The inequality holds by the local update rule:
  \begin{align}
    Q^k_{v+1}(s,a) - Q^k_{v}(s,a) 
    &= (1-\eta) Q^k_{v}(s,a) +  \eta (r(s,a) + \gamma P_{v+1}^k (s,a) V_{v}^k) - Q_{v}^k(s,a) \cr
    &= \eta (r(s,a) + \gamma P_{v+1}^k (s,a) V_{v}^k - Q_{v}^k(s,a)) \cr
    &= \eta (\gamma P_{v+1}^k (s,a) V_{v}^k  - \gamma P(s,a) V^{\star} + Q^{\star}(s,a) - Q_{v}^k(s,a)) \cr
    &= \gamma \eta P_{v+1}^k (s,a) (V_{v}^k - V^{\star}) + \gamma \eta (P_{v+1}^k (s,a) - P(s,a)) V^{\star} + \eta \Delta_{v}^k(s,a),
  \end{align}
  and  \begin{align} \label{eq:asyncq-td-decomp}
    | Q^k_{u} (s,a) - Q^k_{\syn(u)} (s,a) |
    &\le \sum_{v \in \cU_{\syn(u), u}^k(s,a)} |Q^k_{v+1}(s,a) - Q^k_{v}(s,a)| \cr
    &\le \sum_{v \in \cU_{\syn(u), u}^k(s,a)} \left (\eta |\Delta_{v}^k (s,a)| + \gamma \eta | P_{v+1}^k (s,a) (V_{v}^k - V^{\star})| \right)  \cr
    &\qquad + \left |\gamma \eta \sum_{v \in \cU_{\syn(u), u}^k(s,a)}  (P_{v+1}^k (s,a) - P(s,a)) V^{\star} \right | \cr
    &\le \sum_{v \in \cU_{\syn(u), u}^k(s,a)} 2 \eta \|\Delta_{v}^k\|_{\infty} 
     + \left |\gamma \eta \sum_{v \in \cU_{\syn(u), u}^k(s,a)}  (P_{v+1}^k (s,a) - P(s,a)) V^{\star} \right |    ,
  \end{align}
  where the last inequality holds since  $\|P_{v+1}^k (s,a)\|_{1} \le 1$ and $\|V_{v}^k - V^{\star} \|_{\infty} \le \|Q_{v}^k - Q^{\star} \|_{\infty}$ (cf. \eqref{eq:vgaptoqgap}).

Now, we shall bound each term separately.
\begin{itemize}
    \item \textbf{Bounding $B_1$.}
The local error $\|\Delta_{v}^k\|_{\infty}$ is bounded as follows.
\begin{lemma} \label{lemma:asyncq-error-indiv}
    Assume {$ \eta \tau \le \frac{1}{2}$}. For any given $\delta \in (0,1)$, the following holds for any $k \in [K]$ and $0 \le u < T$:
    \begin{align}
      \| \Delta_{u}^k \|_{\infty}
      \le \| \Delta_{\syn(u)}^k \|_{\infty} + \frac{2 \gamma}{1-\gamma}\sqrt{\eta \log{\frac{|\cS||\cA|T K }{\delta}}} 
    \end{align}
    with probability at least $1-\delta$.
\end{lemma}
Then, combining the fact that the number of local updates before the periodic averaging is at most $\tau-1$, we can conclude that
\begin{align}
    2\eta \sum_{v \in \cU_{\syn(u), u}^k(s,a)} \|\Delta_{v}^k \|_{\infty}
    &\le 2\eta |\cU_{\syn(u), u}^k(s,a)| \max_{v \in \cU_{\syn(u), u}^k(s,a)} \|\Delta_{v}^k \|_{\infty} \cr
    &\le 2\eta (\tau-1) \left (\|\Delta_{\syn(u)}^k \|_{\infty} +  \frac{2}{1-\gamma}\sqrt{\eta \log{\frac{|\cS||\cA|T K }{\delta}}} \right).
\end{align}

    \item \textbf{Bounding $B_2$.} Exploiting the independence of the transitions and applying the Hoeffding inequality and using the fact that $|\cU_{\syn(u), u}^k(s,a)|\le \tau-1$, $B_2$ is bounded as follows:
  \begin{align}
    B_2
   &\le \gamma \eta \sqrt{ \sum_{v \in \cU_{\syn(u), u}^k(s,a)} | (P_{v+1}^k (s,a) - P(s,a)) V^{\star} | \log{\frac{|\cS||\cA|T K }{\delta}} } \cr
    &\le \frac{2  \gamma \eta}{1-\gamma} \sqrt{ (\tau -1) \log{\frac{|\cS||\cA|TK}{\delta}}} 
  \end{align}
  for any $k \in [K]$, $(s,a) \in \cS \times \cA$, and $0 \le u < T$ with probability at least  $1-\delta$, where the last inequality follows from
  $\|V^{\star}\|_{\infty} \le \frac{1}{1-\gamma}$, $\| P_{v+1}^k (s,a) \|_{1}$, and $\|P(s,a) \|_{1} \le 1$. 
\end{itemize}

By substituting the bound on $B_1$ and $B_2$ into \eqref{eq:asyncq-td-decomp-2} and using the condition that $\eta \tau <1$, we can claim the stated bound holds and this completes the proof.

\subsubsection{Proof of Lemma~\ref{lemma:asyncq-error-indiv}}
 
  For each state-action $(s,a)\in \cS \times \cA$ and agent $k$, by invoking the recursive relation \eqref{eq:asynq-local-onestepdecomp-tighter} derived from the local Q-learning update in \eqref{eq:asyncq-local}, $\Delta_u^k$ is decomposed as follows:
  \begin{align}
    \Delta_u^k(s,a)
    &=  \underbrace{
      (1-\eta)^{N_{\syn(u), u}^k(s,a)} \Delta_{\syn(u)}^k(s,a)
      }_{=:D_1}  + \underbrace{
      \gamma \sum_{v \in \cU_{\syn(u), u}^k(s,a)} \eta (1-\eta)^{N_{v+1, u}^k(s,a)}  (P(s,a) -  P_{v+1}^k(s,a) ) V^{\star}
      }_{=:D_2}\cr
    &\quad + \underbrace{
      \gamma \sum_{v \in \cU_{\syn(u), u}^k(s,a)} \eta (1-\eta)^{N_{v+1, u}^k(s,a)}  P_{v+1}^k(s,a) (V^{\star} - V^k_{v})
      }_{=:D_3}.
  \end{align}

Now, we obtain the bound on the three decomposed terms separately. 
\begin{itemize}
\item \textbf{Bounding $D_1$.} The term $D_1$ can be bounded by
        \begin{align}
        |D_1| \le (1-\eta)^{N_{\syn(u),u}^k(s,a)} \| \Delta_{\syn(u)}^k\|_{\infty} .
        \end{align}
\item \textbf{Bounding $D_2$.}
    By applying the Hoeffding bound using the independence of transitions, the second term is bounded as follows:
  \begin{align}
    |D_2|
    &\le \gamma \sqrt{ \sum_{v \in \cU_{\syn(u), u}^k(s,a)}  (\eta (1-\eta)^{N_{v+1, u}^k(s,a)})^2 (\|V^{\star} \|_{\infty})^2 \log{\frac{|\cS||\cA|T K }{\delta}} } \cr
    &\le \frac{\gamma}{1-\gamma}\sqrt{\eta \log{\frac{|\cS||\cA|T K }{\delta}}} \defn \rho
  \end{align}
  with probability at least $1-\delta,$ where the last inequality holds due to the fact that $\|V^{\star}\|_{\infty} \le \frac{1}{1-\gamma}$ and $$\sum_{v \in \cU_{\syn(u), u}^k(s,a)}  (\eta (1-\eta)^{N_{v+1, u}^k(s,a)})^2 \le \eta^2 (1 + (1-\eta)^2 + (1-\eta)^4 + \cdots) \le \eta.$$ See \cite[Lemma~1]{li2021asyncq} for the detailed explanation of the bound.
\item \textbf{Bounding $D_3$.}
    Lastly, we bound the third term as follows:
  \begin{align}
    |D_3|
    &\le \gamma \sum_{v \in \cU_{\syn(u), u}^k(s,a)} \eta (1-\eta)^{N_{v+1, u}^k(s,a)}  \|P_{v+1}^k(s,a) \|_{1} \| V^{\star} - V^k_{v}
       \|_{\infty} \nonumber \\
    &\le \gamma \sum_{v \in \cU_{\syn(u), u}^k(s,a)} \eta (1-\eta)^{N_{v+1, u}^k(s,a)}   \| \Delta_{v}^k \|_{\infty} ,
  \end{align}
  where the last inequality follows from the fact that
  $\|P_{v+1}^k (s,a)\|_{1} = 1$ and
  $$Q_v^k(s,a^{\star}(s)) - Q^{\star}(s, a^{\star}(s)) \le V_{v}^k(s) - V^{\star}(s) \le Q_v^k(s,a_v^k(s)) - Q^{\star}(s,a_v^k(s))$$ for any $s \in \cS$, where we denote
  $a^{\star}(s) = \argmax_{a} Q^{\star}(s,a), ~ a_v^k(s) = \argmax_{a} Q_v^k(s,a)$.
  \end{itemize}
  
  By combining the bounds of the above three terms, we obtain the following recursive relation:
  \begin{align} \label{eq:asyncq-localerr-recursive}
    |\Delta_u^k(s,a) |
    &\le (1-\eta)^{N_{\syn(u),u}^k(s,a)} \| \Delta_{\syn(u)}^k\|_{\infty}
      + \rho
      + \gamma \sum_{v \in \cU_{\syn(u), u}^k(s,a)} \eta (1-\eta)^{N_{v+1, u}^k(s,a)}   \| \Delta_{v}^k \|_{\infty}.  
  \end{align}
  Using the recursive relation, we will prove that the following claim holds for any $0 \le m < \tau$ by induction: 
  \begin{align} \label{eq:asyncq-localerr-upbound}
    \| \Delta_{\syn(u) + m}^k \|_{\infty} \le \| \Delta_{\syn(u)}^k \|_{\infty} + 2 \rho,
  \end{align}
  which completes the proof of Lemma~\ref{lemma:asyncq-error-indiv}.
  First, if $m=0,$ the claim is obviously true. Suppose the claim holds for $\syn(u), \syn(u)+1, \cdots, \syn(u)+m-1.$ Then, for $u = \syn(u)+m$, by invoking the recursive relation \eqref{eq:asyncq-localerr-recursive}, we can show that the claim \eqref{eq:asyncq-localerr-upbound} holds for $m$ as follows:
  \begin{align}
    &| \Delta_{\syn(u)+m}^k(s,a) | \cr
    &\le (1-\eta)^{N_{\syn(u),u}^k(s,a)} \| \Delta_{\syn(u)}^k\|_{\infty}
      + \rho
      + \gamma \sum_{v \in \cU_{\syn(u), u}^k(s,a)} \eta (1-\eta)^{N_{v+1, u}^k(s,a)}   ( \| \Delta_{\syn(u)}^k \|_{\infty} + 2 \rho ) \cr
    &= ((1-\eta)^{N_{\syn(u),u}^k(s,a)} + \gamma \sum_{v \in \cU_{\syn(u), u}^k(s,a)} \eta (1-\eta)^{N_{v+1, u}^k(s,a)}) \| \Delta_{\syn(u)}^k \|_{\infty} + (1+2\gamma \sum_{v \in \cU_{\syn(u), u}^k(s,a)} \eta (1-\eta)^{N_{v+1, u}^k(s,a)}) \rho \cr
    &= ((1-\eta)^{N_{\syn(u),u}^k(s,a)} + \gamma (1-(1-\eta)^{N_{\syn(u),u}^k(s,a)}) \| \Delta_{\syn(u)}^k \|_{\infty} + (1+2\gamma (1-(1-\eta)^{N_{\syn(u),u}^k(s,a)})) \rho \cr
    &\le \| \Delta_{\syn(u)}^k \|_{\infty} + 2 \rho ,
  \end{align}
  where the last inequality holds since
  $$(1-\eta)^{N_{\syn(u),u}^k(s,a)} \ge (1-\eta)^{\tau} \ge (\frac{1}{4})^{\eta \tau} \ge \frac{1}{2}$$
  provided that {$ \eta \tau \le \frac{1}{2}$.}

%% file: fedasyncqweighted_proof.tex
\subsection{Proof of Lemma~\ref{lemma:asyncwq-wsum}}
\label{proof:asyncwq-wsum}

First, using the fact that 
  $$1 \le (1-\eta)^{-N_{t-\tau, t}^k(s,a)} \le e^{\eta \tau} \le 3$$
  given that $\eta\tau \le 1$, by the definition of $\alpha_t^k$ (cf.~\eqref{eq:asyncq-weight-skewed}),
  we derive \eqref{eq:asyncwq-alpha} as follows:
  \begin{align*} 
    \frac{1}{3K} \le \frac{ 1 }{K \max_{k'\in[K]} (1-\eta)^{-N_{t-\tau, t}^{k'}(s,a)} } \le \alpha_{t}^k(s,a)
    &= \frac{ (1-\eta)^{-N_{t-\tau, t}^k(s,a)} }{ \sum_{k'=1}^K (1-\eta)^{-N_{t-\tau, t}^{k'}(s,a)}} 
    \le \frac{ (1-\eta)^{-N_{t-\tau, t}^k(s,a)} }{K}
      \le \frac{3}{K} .
  \end{align*}
  Moving onto \eqref{eq:asyncwq-w0}, it follows that
  \begin{align*}  
    \widetilde{\omega}_{0,t} (s,a)
    &= \prod_{h=0}^{\nsyn(t)-1} \widetilde{\lambda}_{h\tau, (h+1)\tau} (s,a) \cr
    &= \prod_{h=0}^{\nsyn(t)-1} \sum_{k=1}^K \alpha_{(h+1)\tau}^k(s,a) (1-\eta)^{N_{h\tau, (h+1)\tau}^k (s,a)} \cr
    &\overset{\mathrm{(i)}}{=} \prod_{h=0}^{\nsyn(t)-1} \frac{K}{\sum_{k=1}^K (1-\eta)^{-N_{h\tau, (h+1)\tau}^k(s,a)}} \cr
    &\stackrel{\mathrm{(ii)}}{\le} \prod_{h=0}^{\nsyn(t)-1} \frac{1}{(1-\eta)^{-\frac{1}{K}\sum_{k=1}^K N_{h\tau, (h+1)\tau}^k(s,a)}} \cr
    &= (1-\eta)^{\sum_{h=0}^{\nsyn(t)-1} \frac{1}{K} \sum_{k=1}^K N_{h\tau, (h+1)\tau}^k(s,a)} = (1-\eta)^{\frac{1}{K} \sum_{k=1}^K N_{0,t}^k(s,a)} , 
  \end{align*}
  where (i) follows from the definition of $\alpha_t^k$ (cf.~\eqref{eq:asyncq-weight-skewed}), (ii) follows from Jensen's inequality.
  
  Next, we obtain \eqref{eq:asyncwq-wsum-1} through the following derivation:
  \begin{align} 
 &   \sum_{k=1}^K \sum_{u \in \cU_{0, t}^k(s,a)} \widetilde{\omega}_{u,t}^k(s,a)
    =\sum_{k=1}^K \sum_{h=0}^{\nsyn(t)-1} \sum_{u \in \cU_{h\tau, (h+1)\tau}^k(s,a)} \widetilde{\omega}_{u,t}^k(s,a) \nonumber \\
    &=\sum_{k=1}^K \sum_{h=0}^{\nsyn(t)-1} \alpha_{(h+1)\tau}^k (s,a) \sum_{u \in \cU_{h\tau, (h+1)\tau}^k(s,a)} \eta (1-\eta)^{N_{u+1, (h+1)\tau}^k(s,a)} \left  (\prod_{l=h+1}^{\nsyn(t)-1} \widetilde{\lambda}_{l\tau, (l+1)\tau}(s,a) \right)  \nonumber \\
    &=  \sum_{k=1}^K \sum_{h=0}^{\nsyn(t)-1} \alpha_{(h+1)\tau}^k (s,a) \left( 1- (1-\eta)^{N_{h\tau, (h+1)\tau}^k(s,a)} \right) \left  (\prod_{l=h+1}^{\nsyn(t)-1} \widetilde{\lambda}_{l\tau, (l+1)\tau}(s,a) \right)  \nonumber \\
    &\stackrel{\mathrm{(i)}}{=}   \sum_{h=0}^{\nsyn(t)-1} \left  (\prod_{l=h+1}^{\nsyn(t)-1} \widetilde{\lambda}_{l\tau, (l+1)\tau}(s,a) \right)  \sum_{k=1}^K \alpha_{(h+1)\tau}^k (s,a) \left(1- (1-\eta)^{N_{h\tau, (h+1)\tau}^k(s,a)} \right)  \nonumber \\
    &\stackrel{\mathrm{(ii)}}{=}    \sum_{h=0}^{\nsyn(t)-1} \left  (\prod_{l=h+1}^{\nsyn(t)-1} \widetilde{\lambda}_{l\tau, (l+1)\tau}(s,a) \right)  \left  ( 1 - \sum_{k=1}^K \alpha_{(h+1)\tau}^k (s,a) (1-\eta)^{N_{h\tau, (h+1)\tau}^k(s,a)} \right) \nonumber \\
    &=   \sum_{h=0}^{\nsyn(t)-1} \left  (\prod_{l=h+1}^{\nsyn(t)-1} \widetilde{\lambda}_{l\tau, (l+1)\tau}(s,a) \right)  \left  ( 1- \widetilde{\lambda}_{h \tau, (h+1)\tau}(s,a) \right) \nonumber \\
    &\stackrel{\mathrm{(iii)}}{=} 1 - \widetilde{\lambda}_{0, \tau}(s,a)  \widetilde{\lambda}_{\tau, 2\tau}(s,a) \cdots \widetilde{\lambda}_{ (\nsyn(t)-1) \tau, t}(s,a)  = 1 - \widetilde{\omega}_{0,t}(s,a), \label{eq:asyncwq-wsum-1-deriv}
  \end{align}
  where (i) follows by reordering the summation, (ii) follows by $\sum_{k=1}^K \alpha_t^k(s,a) =1$, and (iii) holds by cancellation.

  In a similar manner, \eqref{eq:asyncwq-wsum-1-part} is derived as follows:
  \begin{align*}
    \sum_{k=1}^K \sum_{u \in \cU_{0, h'\tau}^k(s,a)} \widetilde{\omega}_{u,t}^k (s,a)
    &=\sum_{k=1}^K \sum_{h=0}^{h'-1} \sum_{u \in \cU_{h\tau, (h+1)\tau}^k(s,a)} \widetilde{\omega}_{u,t}^k(s,a) \cr
    &=   \sum_{h=0}^{h'-1} \left  (\prod_{l=h+1}^{\nsyn(t)-1} \widetilde{\lambda}_{l\tau, (l+1)\tau}(s,a) \right)  \left  ( 1- \widetilde{\lambda}_{h \tau, (h+1)\tau}(s,a) \right) \cr
    &\le \prod_{l=h'}^{\nsyn(t)-1} \widetilde{\lambda}_{l\tau, (l+1)\tau}(s,a) \cr
    &\le (1-\eta)^{\frac{1}{K} \sum_{k=1}^k N_{h'\tau, t}^k(s,a)},
  \end{align*}
  where the last inequality follows from 
  \begin{align*}
    \prod_{l=h'}^{\nsyn(t)-1} \widetilde{\lambda}_{l\tau, (l+1)\tau}(s,a)
    = \prod_{h=h'}^{\nsyn(t)-1} \frac{K}{\sum_{k=1}^K (1-\eta)^{-N_{h\tau, (h+1)\tau}^k(s,a)}} 
    \le \prod_{h=h'}^{\nsyn(t)-1} \frac{1}{(1-\eta)^{-\frac{1}{K}\sum_{k=1}^K N_{h\tau, (h+1)\tau}^k(s,a)}} 
  \end{align*}
due to Jensen's inequality.
  
  Finally, with basic algebraic calculations, \eqref{eq:asyncwq-wsum-2} is derived as follows:
  \begin{align*}
&  \sum_{k=1}^K \sum_{u \in \cU_{0, t}^k(s,a)} (\widetilde{\omega}_{u,t}^k(s,a))^2 
    =\sum_{k=1}^K \sum_{h=0}^{\nsyn(t)-1} \sum_{u \in \cU_{h\tau, (h+1)\tau}^k(s,a)} (\widetilde{\omega}_{u,t}^k(s,a))^2 \cr
    &=\sum_{k=1}^K \sum_{h=0}^{\nsyn(t)-1} (\alpha_{(h+1)\tau}^k (s,a))^2 \left  (\prod_{l=h+1}^{\nsyn(t)-1} \widetilde{\lambda}_{l\tau, (l+1)\tau}(s,a) \right)^2  \sum_{u \in \cU_{h\tau, (h+1)\tau}^k(s,a)} \left(\eta (1-\eta)^{N_{u+1, (h+1)\tau}^k(s,a)} \right)^2  \cr
    &\stackrel{\mathrm{(i)}}{\le} 2 \sum_{k=1}^K \sum_{h=0}^{\nsyn(t)-1} (\alpha_{(h+1)\tau}^k (s,a))^2 \left  (\prod_{l=h+1}^{\nsyn(t)-1} \widetilde{\lambda}_{l\tau, (l+1)\tau}(s,a) \right)^2  \eta \left(1- (1-\eta)^{ N_{h\tau, (h+1)\tau}^k(s,a)} \right)  \cr 
    &\stackrel{\mathrm{(ii)}}{\le}  \frac{6\eta}{K} \sum_{h=0}^{\nsyn(t)-1}  \left  (\prod_{l=h+1}^{\nsyn(t)-1} \widetilde{\lambda}_{l\tau, (l+1)\tau}(s,a) \right)^2  \sum_{k=1}^K \alpha_{(h+1)\tau}^k (s,a)  \left(1- (1-\eta)^{ N_{h\tau, (h+1)\tau}^k(s,a)} \right)  \cr
    &\stackrel{\mathrm{(iii)}}{\le}  \frac{6\eta}{K},
  \end{align*}
  where (i) holds because
 \begin{align}
   \sum_{u \in \cU_{h\tau, (h+1)\tau}^k(s,a)} (\eta (1-\eta)^{N_{u+1, (h+1)\tau}^k(s,a)} )^2
   &= \eta^2 \frac{1- (1-\eta)^{2(N_{h\tau, (h+1)\tau}^k(s,a))}}{1-(1-\eta)^2} \cr
   &\le \eta (1- (1-\eta)^{2(N_{h\tau, (h+1)\tau}^k(s,a))}) \cr
   &\le 2\eta (1- (1-\eta)^{(N_{h\tau, (h+1)\tau}^k(s,a))}) 
 \end{align}
 given that $2x -x^2 \ge x$ for $x \le 1$ and $(1-x^2) \le 2(1-x)$,
  (ii) follows from \eqref{eq:asyncwq-alpha}, and (iii) follows from the same reasoning of \eqref{eq:asyncwq-wsum-1-deriv}.

\subsection{Proof of Lemma~\ref{lemma:asyncwq-error2-freedman}}
\label{proof:asyncwq-error2-freedman}

Without loss of generality, we prove the claim for some fixed $1 \le t \le T$ and $(s,a) \in \cS \times \cA$. For notation simplicity, let
\begin{align}
    \widetilde{y}_{u,t}^k(s,a)  = 
    \begin{cases}
         \widetilde{\omega}_{u,t}^k(s,a) (P(s,a) -  P_{u+1}^k(s,a) ) V^k_{u} &\qquad \text{if}~(s_{u}^k, a_{u}^k) = (s,a) \\
         0 &\qquad \text{otherwise}
    \end{cases},
\end{align}
where 
\begin{align}
    \widetilde{\omega}_{u,t}^k(s,a) 
    &= \frac{\eta(1-\eta)^{-N_{\nsyn(u) \tau , u+1}^k(s,a)}}{K} \prod_{h=\nsyn(u)}^{\nsyn(t)-1} \frac{K}{\sum_{k'=1}^K (1-\eta)^{-N_{h \tau, (h+1)\tau}^{k'}(s,a)}} ,
\end{align}
then $E_t^2(s,a)  = \gamma  \sum_{k=1}^K \sum_{u=0}^{t-1}    \widetilde{y}_{u,t}^k(s,a)$. 
However, due to the dependency between $P_{u+1}^k(s,a)$ and $\widetilde{\omega}_{u,t}^k(s,a)$ arising from the Markovian sampling, it is difficult to track the sum of $\widetilde{y}: =\{ \widetilde{y}_{u,t}^k(s,a)\}$ directly. To address this issue, we will first analyze the sum using a collection of approximate random variables $ \widehat{y} = \{ \widehat{y}_{u,t}^k(s,a) \} $ drawn from a carefully constructed set $\widehat{\cY}$, which is closely coupled with the target $\{\widetilde{y}_{u,t}^k(s,a)\}_{0\le u <t}$, i.e.,
\begin{align} \label{eq:proximity}
    D(\widetilde{y}, \widehat{y}) &\defn  \left | \sum_{k=1}^K \sum_{u=0}^{t-1} \big(  \widetilde{y}_{u,t}^k(s,a) - \widehat{y}_{u,t}^k(s,a) \big) \right |  
\end{align}
is sufficiently small. In addition, $ \widehat{y}$ shall exhibit some useful statistical independence and thus easier to control its sum; we shall control this over the entire set $\widehat{\cY}$. Finally, leveraging the proximity above, we can obtain the desired bound on $\widetilde{y}$ via triangle inequality. We now provide details on executing this proof outline, where the crust is in designing the set $\widehat{\cY}$ with a controlled size. 

Before describing our construction, let's introduce the following useful event:
\begin{align}\label{eq:avgvisit_bound}
 \cB_M &\defn  \bigcap_{u=0}^{  t-M\tau } \left \{{\frac{1}{4}} \muminavg(s,a) K  M\tau \le \sum_{k=1}^K N_{u,u+M\tau}^{k}(s,a) \le {2} \muminavg(s,a) K M\tau  \right \}, 
\end{align}
where $M= M(s,a):= \lfloor \frac{1}{8 \eta \muminavg(s,a) \tau} \rfloor$. Note that {$M \ge \frac{1}{16 \eta \muminavg(s,a) \tau}$} since $ \eta \tau \le 1/16$. Combining this with the assumption $\eta \le \frac{1}{16 \tth(s,a) \muminavg(s,a)}$ (see \eqref{eq:def_tth} for the definition of $\tth(s,a)$), it follows that $M\tau \ge \tth(s,a)$ always holds. Then, $ \cB_M $ holds with probability at least $1 - \frac{\delta}{|\cS||\cA|T}$
according to Lemma~\ref{lemma:dist-asyncwq-multi-visit-mix}. The rest of the proof shall be carried out under the event $ \cB_M $.

\paragraph{Step 1: constructing $\widehat{\cY}$.}
To decouple dependency between $P_{u+1}^k(s,a)$ and $\widetilde{\omega}_{u,t}^k(s,a)$, we will introduce approximates of $\widetilde{\omega}_{u,t}^k(s,a)$ that only depend on history until $u$ by replacing a factor dependent on future with some constant. To gain insight, we factorize $\widetilde{\omega}_{u,t}^k(s,a)$ into two components as follows:
\begin{align} \label{eq:asyncwq-omega-decomp}
    \widetilde{\omega}_{u,t}^k(s,a) 
    &= \prod_{h=h_0(u,t)}^{\nsyn(u)-1} \left (\frac{K}{\sum_{k'=1}^K (1-\eta)^{-N_{h \tau, (h+1)\tau}^{k'}(s,a)}} \frac{\sum_{k'=1}^K (1-\eta)^{-N_{h \tau, (h+1)\tau}^{k'}(s,a)}}{K} \right ) \cr
    &\qquad \times \frac{\eta(1-\eta)^{-N_{\nsyn(u) \tau , u+1}^k(s,a)}}{K} \prod_{h=\nsyn(u)}^{\nsyn(t)-1} \frac{K}{\sum_{k'=1}^K (1-\eta)^{-N_{h \tau, (h+1)\tau}^{k'}(s,a)}} \cr
    &= \underbrace{ \left (\prod_{h=h_0(u,t)}^{\nsyn(u)-1} \left (\frac{\sum_{k'=1}^K (1-\eta)^{-N_{h \tau, (h+1)\tau}^{k'}(s,a)}}{K} \right ) \frac{\eta(1-\eta)^{-N_{\nsyn(u) \tau , u+1}^k(s,a)}}{K} \right) }_{\text{dependent on history until } u}\cr
    &\qquad \times \underbrace{  \left (\prod_{h=h_0(u,t)}^{\nsyn(t)-1} \frac{K}{\sum_{k'=1}^K (1-\eta)^{-N_{h \tau, (h+1)\tau}^{k'}(s,a)}} \right ) }_{\text{dependent on history and future until } t} \cr
    &= \underbrace{ \left (\prod_{h=h_0(u,t)}^{\nsyn(u)-1} \left (\frac{\sum_{k'=1}^K (1-\eta)^{-N_{h \tau, (h+1)\tau}^{k'}(s,a)}}{K} \right ) \frac{\eta(1-\eta)^{-N_{\nsyn(u) \tau , u+1}^k(s,a)}}{K} \right) }_{\defn x_{u}^k(s,a)}\cr
    &\qquad \times    \prod_{l=1}^{l(u,t)} \underbrace{ \left ( \prod_{h=\max\{0, \nsyn(t)- lM\}}^{\nsyn(t)- (l-1)M-1} \frac{K}{\sum_{k'=1}^K (1-\eta)^{-N_{h \tau, (h+1)\tau}^{k'}(s,a)}} \right ) }_{\defn z_{l}(s,a) } . 
\end{align}
where we denote $ l(u,t) \defn \lceil \frac{(t-u)}{M\tau} \rceil$ and $h_0(u,t) = \max\{0, \nsyn(t)-l(u,t) M\}$.

Motivated by the above decomposition, we will construct $\widehat{\cY}$ by approximating the future-dependent parameter $z_{l}(s,a)$ for $1 \le l \le L$, where $L \defn \min \{ \lceil \frac{t}{M\tau} \rceil, \lceil 64 \log{(K/\eta)} \rceil\} $.
Using the fact that $1+x \le  \exp(x) \le 1+2x$ holds for any $0\le x <1$, and {$ \eta \frac{\sum_{k'=1}^K N_{h \tau, (h+1)\tau}^{k'}(s,a)}{K} \le \eta \tau \le 1$}, and applying Jensen's inequality, 
\begin{align*} 
 \exp \left (- \eta\frac{\sum_{k'=1}^K N_{h \tau, (h+1)\tau}^{k'}(s,a)}{K} \right ) \geq    \frac{K}{\sum_{k'=1}^K (1-\eta)^{-N_{h \tau, (h+1)\tau}^{k'}(s,a)}} 
    & \ge \frac{K}{ \sum_{k'=1}^K e^{ \eta  N_{h \tau, (h+1)\tau}^{k'}(s,a)}}  \cr
     &\ge \frac{1}{ 1 + 2  \eta \sum_{k'=1}^K  \frac{  \sum_{k'=1}^K N_{h \tau, (h+1)\tau}^{k'}(s,a)}{K}}  \cr
     &\ge \exp \left (- 2 \eta \frac{  \sum_{k'=1}^K N_{h \tau, (h+1)\tau}^{k'}(s,a)}{K} \right )  .
\end{align*}
Therefore, for $1\leq l<L$, under $\cB_M$, the range of $z_{l}(s,a)$ is bounded as follows:
$$z_{l}(s,a) \in \left [\exp(-{4}\eta \muminavg(s,a) M \tau ), ~\exp(-{\frac{1}{4}}\eta \muminavg(s,a) M \tau ) \right ] .$$
Using this property, we construct a set of values that can cover possible realizations of $z_{l}(s,a)$ in a fine-grained manner as follows:
\begin{align}\label{eq:asyncwq-zset}
    \cZ \defn \left \{ \exp\left(-{\frac{1}{4}}\eta \muminavg(s,a) M \tau - \frac{i \eta}{K} \right) ~~ \Big | i \in \mathbb{Z}: ~~0\le i   < {4} K \muminavg(s,a) M\tau \right \}.
\end{align}
Note that the distance of adjacent elements of $\cZ$  is bounded by $\eta / K e^{-{1/4}\eta \muminavg(s,a) M \tau }$, and the size of the set is bounded by ${4} K \muminavg(s,a) M\tau $. For $l=L$, because the number of iterations involved in $z_{L}(s,a)$ can be less than $M\tau$, it follows that $z_{L}(s,a) \in \left [\exp(-{4}\eta \muminavg(s,a) M \tau ), 1 \right ]$. Hence, we construct the set
\begin{align}\label{eq:asyncwq-zset0}
    \cZ_0 \defn \left \{ \exp\left( - \frac{i \eta}{K} \right) ~~ \Big | i \in \mathbb{Z}:  ~~0\le i   < {4} K \muminavg(s,a) M\tau \right \}.
\end{align}
In sum, we can always find $(\widehat{z}_1, \cdots, \widehat{z}_l, \cdots, \widehat{z}_L) \in \cZ^{L-1} \times \cZ_0$ where its entry-wise distance to $(z_{l}(s,a))_{l \in [L-1]}$ (resp. $z_L(s,a)$) is at most $\eta / K e^{-{1/4}\eta \muminavg(s,a) M \tau }$ (resp. $\eta/K$).

Moreover, we approximate $x_{u}^k(s,a)$ by clipping it when the accumulated number of visits of all agents is not too large as follows:
\begin{align} \label{eq:asyncwq-xclip}
    \widehat{x}_{u}^k(s,a)  = 
    \begin{cases}
         x_{u}^k(s,a) &\qquad \text{if}~ \sum_{k=1}^K N_{h_0(u,t) \tau, \nsyn(u)\tau}^{k}(s,a) \le {2} K \muminavg(s,a) M\tau \\
         0 &\qquad \text{otherwise}
    \end{cases}.
\end{align}
Note that the clipping never occurs and $\widehat{x}_{u}^k(s,a) = x_{u}^k(s,a)$ for all $u$ as long as $\cB_M$ holds. To provide useful properties of $\widehat{x}_{u}^k(s,a)$ that will be useful later, we record the following lemma whose proof is provided in Appendix~\ref{proof:asyncwq-xsum}.
\begin{lemma} \label{lemma:asyncwq-xsum} 
For any state-action pair $(s,a) \in \cS \times \cA$, consider any integers $1 \le t \le T$ and $1 \le l \le \lceil \frac{t}{M\tau} \rceil $, where {$M=\lfloor \frac{1}{8 \eta \muminavg(s,a) \tau} \rfloor$}. Suppose that {$4\eta \tau \le 1$}, then $\widehat{x}_{u}^k(s,a)$ defined in \eqref{eq:asyncwq-xclip} satisfy
\begin{subequations}
 \begin{align}
    \label{eq:asyncwq-xmax} \forall u \in [h_0, \nsyn(t)-(l-1)M) ~~:~~ \widehat{x}_{u}^k(s,a) &\le \frac{9\eta}{K} , \\
    \label{eq:asyncwq-xsum-1}  \sum_{h=h_0}^{\nsyn(t)-(l-1)M-1} \sum_{u \in \cU_{h\tau, (h+1)\tau}^k(s,a)}  \sum_{k=1}^K \widehat{x}_{u}^k(s,a) &\le  {16}\eta \muminavg(s,a) M\tau, \\
    \label{eq:asyncwq-xsum-2} \sum_{h=h_0}^{\nsyn(t)-(l-1)M-1} \sum_{u \in \cU_{h\tau, (h+1)\tau}^k(s,a)}  \sum_{k=1}^K (\widehat{x}_{u}^k(s,a))^2 & \le \frac{ {64}  \eta^2 \muminavg(s,a) M \tau }{K} ,
\end{align}
where $h_0 = \max\{0, \nsyn(t)-lM\}$.
\end{subequations} 
\end{lemma}

Finally, for each $ \bm{z} = (\widehat{z}_1, \cdots, \widehat{z}_L) \in \cZ^{L-1} \times \cZ_0$, setting $\widehat{\omega}_{u,t}^k(s,a; \bm{z}) = \widehat{x}_{u}^k(s,a) \prod_{l=1}^{l(u,t)} \widehat{z}_l$, an approximate random sequence $\widehat{y}_{\bm{z}} = \{\widehat{y}_{u,t}^k(s,a; \bm{z})\}_{0\le u <t}$ can be constructed as follows:
\begin{align} \label{def:widehaty}
    \widehat{y}_{u,t }^k(s,a ; \bm{z})  = 
    \begin{cases}
         \widehat{\omega}_{u,t}^k(s,a ; \bm{z}) (P(s,a) -  P_{u+1}^k(s,a) ) V^k_{u} &\qquad \text{if}~(s_{u}^k, a_{u}^k) = (s,a) ~\text{and}~ l(u,t) \le L \\
         0 &\qquad \text{otherwise}
    \end{cases}.
\end{align}
If $t> LM\tau$, for any $u < t-LM\tau$, i.e., $ l(u,t) > L$, we set $\widehat{y}_{u,t}^k(s,a;\bz) = 0$ since the magnitude of $\widetilde{\omega}_{u,t}^k(s,a)$ becomes negligible when the time difference between $u$ and $t$ is large enough, and the fine-grained approximation using $\cZ$ is no longer needed, as shall be seen momentarily. 
Finally, denote a collection of the approximates induced by $\cZ^{L-1} \times \cZ_0$ as 
$$\widehat{\cY} = \{\widehat{y}_{ \bm{z}}: \quad \bm{z} \in \cZ^{L-1} \times \cZ_0 \}.$$

\paragraph{Step 2: bounding the approximation error $D(\widetilde{y}, \widehat{y}_{\bm{z}} )$.}
We now show that under $\cB_M$,  there always exists $\widehat{y}_{ \bm{z}}: =\widehat{y}_{ \bm{z}(\widetilde{y})} \in \widehat{\cY}$ such that 
\begin{equation}\label{eq:peach}
D(\widetilde{y}, \widehat{y}_{\bm{z}} ) <   \frac{{129}}{1-\gamma} \sqrt{\frac{ L\eta}{K}}.
\end{equation}
To this end, we first decompose the approximation error as follows:
\begin{align*}
  &  \min_{\widehat{y}_{\bm{z}} \in \widehat{\cY}} D(\widetilde{y}, \widehat{y}_{\bm{z}} ) \cr
   &  = \min_{\bm{z} \in \cZ^{L-1} \times \cZ_0} \left | \sum_{k=1}^K \sum_{u=0}^{t-1} \left( \widetilde{y}_{u,t}^k(s,a) - \widehat{y}_{u,t}^k(s,a; \bm{z} ) \right) \right | \cr
   &\le \underbrace{\max_{\bm{z} \in \cZ^{L-1} \times \cZ_0} \left | \sum_{k=1}^K \sum_{u=0}^{t-LM\tau-1}  \widetilde{y}_{u,t}^k(s,a) - \widehat{y}_{u,t}^k(s,a; \bm{z}) \right | }_{ =: D_1}
   + \underbrace{ \min_{\bm{z} \in \cZ^{L-1} \times \cZ_0} \left | \sum_{k=1}^K \sum_{u=t-LM\tau}^{t-1}  \widetilde{y}_{u,t}^k(s,a) - \widehat{y}_{u,t}^k(s,a; \bm{z}) \right | }_{=: D_2} 
\end{align*}

\begin{itemize}
    \item \textbf{Bounding $D_1$.}
    This term appears only when $t>LM\tau$. Since $\widehat{y}_{u,t}^k(s,a; \bz) =0$ for all $u < t-LM\tau$ regardless of $\bm{z}$ by construction, 
\begin{align*}
    \left | \sum_{k=1}^K \sum_{u=0}^{t-LM\tau-1}  \widetilde{y}_{u,t}^k(s,a) - \widehat{y}_{u,t}^k(s,a; \bz) \right |
    &\le \sum_{k=1}^K \sum_{  u \in \cU_{0, t-LM\tau}^k(s,a)} \widetilde{\omega}_{u,t}^k(s,a)  \| P(s,a) -  P_{u+1}^k(s,a) \|_{1} \| V^k_{u} \|_{\infty } \cr
    &\stackrel{\mathrm{(i)}}{\le} \frac{2}{1-\gamma} \sum_{k=1}^K \sum_{ u \in \cU_{0, t-LM\tau}^k(s,a)} \widetilde{\omega}_{u,t}^k(s,a)   \cr
    &\stackrel{\mathrm{(ii)}}{\le} \frac{2}{1-\gamma} (1-\eta)^{\frac{1}{K}\sum_{k=1}^K N_{t-LM\tau, t}^k(s,a)} \cr
    &\stackrel{\mathrm{(iii)}}{\le} \frac{2}{1-\gamma} e^{-\eta {\frac{1}{4}}\muminavg(s,a) LM\tau} \cr
    &\stackrel{\mathrm{(iv)}}{\le} \frac{2\eta}{(1-\gamma)K},
\end{align*}
where (i) holds since $\|P(s,a)\|_1,~\|P_{u}^k(s,a)\|_1 \le 1$ and $\|V_{u-1}^k\|_{\infty} \le \frac{1}{1-\gamma}$ (cf.~\eqref{eq:bounded_iterates}), (ii) follows from \eqref{eq:asyncwq-wsum-1-part} in Lemma~\ref{lemma:asyncwq-wsum}, (iii) holds due to $\cB_M$, and (iv) holds because {$L \ge  64 \log{\frac{K}{\eta}}  \ge \frac{4}{\eta \muminavg(s,a) M\tau} \log{\frac{K}{\eta}}$} given that {$\eta \muminavg(s,a) M\tau \ge 1/16$}.
    \item \textbf{Bounding $D_2$.}
    Since $\widehat{x}_{u}^k(s,a) = x_{u}^k(s,a)$  when $\cB_M$ holds, in view of \eqref{def:widehaty}, we have
\begin{align*}
    & \min_{\bm{z} \in \cZ^{L-1} \times \cZ_0} \left | \sum_{k=1}^K \sum_{u=t-LM\tau}^{t-1}  \widetilde{y}_{u,t}^k(s,a) - \widehat{y}_{u,t}^k(s,a; \bz) \right | \cr
    &\le \min_{\bm{z} \in \cZ^{L-1} \times \cZ_0}  \sum_{k=1}^K \sum_{u \in \cU_{t-LM\tau, t}^k(s,a)} \big|\widetilde{\omega}_{u,t}^k(s,a) - \widehat{\omega}_{u,t}^k(s,a;\bz) \big|  \, \| P(s,a) -  P_{u+1}^k(s,a) \|_{1} \| V^k_{u} \|_{\infty } \cr
    &\le \frac{2}{1-\gamma} \min_{\bm{z} \in \cZ^{L-1} \times \cZ_0} \left( \sum_{l=1}^L \sum_{h=\nsyn(t)-lM}^{\nsyn(t)-(l-1)M-1} \sum_{u \in \cU_{h\tau, (h+1)\tau}^k(s,a)}  \sum_{k=1}^K \widehat{x}_{u}^k(s,a) \left |\prod_{l'=1}^{l} z_{l'}(s,a) - \prod_{l'=1}^{l} \widehat{z}_{l'} \right | 
    \right)  ,
\end{align*}
where the last inequality holds since $\|P(s,a)\|_1,~\|P_{u}^k(s,a)\|_1 \le 1$ and $\|V_{u-1}^k\|_{\infty} \le \frac{1}{1-\gamma}$ (cf.~\eqref{eq:bounded_iterates}).

Note that for any given $\{z_{l}(s,a)\}_{l \in [L]}$, under $\cB_M$, there exists $\widehat{\bz}^{\star} = (\widehat{z}_1^{\star}, \ldots, \widehat{z}_l^{\star}, \ldots, \widehat{z}_L^{\star}) \in \cZ^{L-1} \times \cZ_0$ such that $|\widehat{z}_l^{\star} - z_{l}(s,a)| \le \frac{\eta}{K} \exp(-{1/4}\eta \muminavg(s,a) M \tau )$ for $l<L$ and $|\widehat{z}_L^{\star} - z_{L}(s,a)| \le \frac{\eta}{K} $. Also, recall that $z_l(s,a), ~ \widehat{z}_l^{\star} \le \exp(-{1/4}\eta \muminavg(s,a) M \tau ) $ for $l < L$ and $z_L(s,a), ~ \widehat{z}_L^{\star} \le 1$. Then, for any $l \le L$ it follows that:
\begin{align*}
     \left |\prod_{l'=1}^{l} z_{l'}(s,a) - \prod_{l'=1}^{l} \widehat{z}_{l'}^{\star} \right | 
    &\le \Big ( 
    \Big |\prod_{l'=1}^{l} z_{l'}(s,a) - \widehat{z}_{1}^{\star} \prod_{l'=2}^{l} z_{l'}(s,a) \Big | 
     + \cdots + \Big |  z_{l} \prod_{l'=1}^{l-1} \widehat{z}_{l'}^{\star} - \prod_{l'=1}^{l} \widehat{z}_{l'}^{\star} \Big |
     \Big ) \cr
     &\le \exp \Big (-{\frac{1}{4}}(l-1)\eta  \muminavg(s,a) M \tau \Big ) \sum_{l'=1}^l \frac{\eta}{K}   \cr
    &\le \exp \Big (-{\frac{1}{4}}(l-1) \eta  \muminavg(s,a) M \tau \Big ) \frac{L\eta}{K} .
\end{align*}
Then, applying the above bound and \eqref{eq:asyncwq-xsum-1} in Lemma~\ref{lemma:asyncwq-xsum},
\begin{align*}
    & \min_{\bm{z} \in \cZ^{L-1} \times \cZ_0} \left | \sum_{k=1}^K \sum_{u=t-LM\tau}^{t-1}  \widetilde{y}_{u,t}^k(s,a) - \widehat{y}_{u,t}^k(s,a; \bz) \right | \cr
    &\le \frac{2}{1-\gamma}  \sum_{l=1}^L \sum_{h=\nsyn(t)-lM}^{\nsyn(t)-(l-1)M-1} \sum_{u \in \cU_{h\tau, (h+1)\tau}^k(s,a)}  \sum_{k=1}^K \widehat{x}_{u}^k(s,a) \left |\prod_{l'=1}^{l} z_{l'}(s,a) - \prod_{l'=1}^{l} \widehat{z}_{l'}^{\star} \right |  
     \cr
    &\le \frac{2}{1-\gamma}\frac{L\eta}{K}  \sum_{l=1}^L \exp \Big (-{\frac{1}{4}} (l-1) \eta  \muminavg(s,a) M \tau \Big )  \sum_{h=\nsyn(t)-lM}^{\nsyn(t)-(l-1)M-1} \sum_{u \in \cU_{h\tau, (h+1)\tau}^k(s,a)}  \sum_{k=1}^K \widehat{x}_{u}^k(s,a)
      \cr
    &\le \frac{2 }{1-\gamma}\frac{L\eta}{K}  \frac{1}{1-\exp(-{1/4}\eta  \muminavg(s,a) M \tau )}  ({16}\eta \muminavg(s,a) M\tau) \cr
    &\stackrel{\mathrm{(i)}}{\le}  \frac{2 }{1-\gamma}\frac{L\eta}{K} \frac{{8}}{\eta  \muminavg(s,a) M \tau} {16}\eta \muminavg(s,a) M\tau  
    \le \frac{{256} L \eta}{(1-\gamma) K },
\end{align*}
where (i) holds since { ${1/4}\eta  \muminavg(s,a) M \tau   \le 1 $ } and $e^{-x} \le 1-\frac{1}{2}x$ for any $0\le x \le 1$.

\end{itemize}

By combining the bounds obtained above and using the fact that ${\frac{4\eta L}{K}  \le 1}$ and {$L \le 64 \log{(TK)}$}, we can conclude that
\begin{align*}
    \min_{\widehat{y}_{\bm{z}} \in \widehat{\cY}} D(\widetilde{y}, \widehat{y}_{\bm{z}} )
   &\le \frac{2\eta}{(1-\gamma)K}
   + \frac{{256} L\eta}{(1-\gamma)K} 
   \le 
   \frac{{129}}{1-\gamma} \sqrt{\frac{ L\eta}{K}}.
\end{align*}

\paragraph{Step 3: concentration bound over $\cY$.} We now show that for all elements  in $\widehat{\cY} = \{\widehat{y}_{ \bm{z}}: \; \bm{z} \in \cZ^{L-1} \times \cZ_0 \}$ satisfy
 \begin{align} \label{eq:banana_peel}
 \left |\sum_{k=1}^K \sum_{u=0}^{t-1}  \widehat{y}_{u,t}^k(s,a;\bz) \right |< \frac{{624}}{(1-\gamma)} \sqrt{\frac{\eta }{K} \log{(TK)}\log{\frac{4|\cS||\cA|T^2 K}{\delta}}} 
\end{align}
with probability at least $1- \frac{\delta}{|\cS||\cA|T}$. It suffices to establish \eqref{eq:banana_peel} for a fixed $\bm{z}   \in \cZ^{L-1} \times \cZ_0$ with probability at least $1- \frac{\delta}{|\cS||\cA|T|\cY|}$, where
\begin{align}\label{eq:size_cY}
|\widehat{\cY}| = |\cZ^{L-1} \times \cZ_0 | \le {(4K \muminavg(s,a) M\tau )^{L} \le (K/\eta)^{L} \le (TK)^L}.
\end{align}


For any fixed $\bm{z} = (\widehat{z}_1, \cdots, \widehat{z}_L) \in \cZ^{L-1} \times \cZ_0$, since $\widehat{\omega}_{u,t}^k(s,a; \bz) = \widehat{x}_{u}^k(s,a) \prod_{l=1}^{l(u,t)} \widehat{z}_l$ only depends on the events happened until $u$, which is independent to a transition at $u+1$.  Thus, we can apply Freedman's inequality to bound the sum of $\widehat{y}_{u,t}^k(s,a; \bz)$ since
\begin{align}
    \mexp[ \widehat{y}_{u,t}^k(s,a; \bz) | \mathcal{Y}_{u}] = 0,
\end{align}
where $\cY_u$ denotes the history of visited state-action pairs and updated values of all agents until $u$, i.e.,
$\cY_{u} = \{ (s_{v}^k, a_{v}^k), V_{v}^{k} \}_{k \in [K], v \le {u}}$. 
Before applying Freedman's inequality, we need to calculate the following quantities. First,
  \begin{align} 
      B_t(s,a)
      \defn \max_{k \in [K], 0 \le u < t} | \widehat{y}_{u,t}^k(s,a; \bz) | 
      \le \widehat{x}_{u}^k(s,a) \prod_{l=1}^{l(u,t)} \widehat{z}_l \| P(s,a) -  P_{u+1}^k(s,a) \|_{1} \| V^k_{u} \|_{\infty } 
      \le \frac{{18} \eta}{(1-\gamma)K} ,
    \end{align}
    where the last inequality follows from $\|P(s,a)\|_1,~\|P_{u}^k(s,a)\|_1 \le 1$,  $\|V_{u-1}^k\|_{\infty} \le \frac{1}{1-\gamma}$ (cf.~\eqref{eq:bounded_iterates}), $\hat{z}_l \le 1$, and \eqref{eq:asyncwq-xmax} in Lemma~\ref{lemma:asyncwq-xsum}. Next, we can bound the variance as
    \begin{align} 
      W_t(s,a)
      &\defn  \sum_{u=t-LM\tau}^{t-1} \sum_{k=1}^K \mexp [(\widehat{y}_{u,t}^k(s,a; \bz))^2 | \cY_{u} ] \cr
      & = \sum_{l=1}^L \sum_{h=\max\{0, \nsyn(t)-lM\}}^{\nsyn(t)-(l-1)M-1} \sum_{k=1}^K \sum_{u \in \cU_{h\tau, (h+1)\tau}^k(s,a)}  (\widehat{x}_{u}^k(s,a) \prod_{l'=1}^{l} \widehat{z}_{l'} )^2 \Var_{P(s,a)} (V_{u}^k) \cr
      & \stackrel{\mathrm{(i)}}{\le} \frac{2}{(1-\gamma)^2} \sum_{l=1}^L \left( \prod_{l'=1}^{l} \widehat{z}_{l'}^2 \right) \sum_{h=max\{0, \nsyn(t)-lM\}}^{\nsyn(t)-(l-1)M-1} \sum_{k=1}^K \sum_{u \in \cU_{h\tau, (h+1)\tau}^k(s,a)}  (\widehat{x}_{u}^k(s,a) )^2  \cr  
      & \stackrel{\mathrm{(ii)}}{\le} \frac{2}{(1-\gamma)^2} \sum_{l=1}^L \left( \prod_{l'=1}^{l} \widehat{z}_{l'}^2 \right) \frac{ {64}  \eta^2 \muminavg(s,a) M \tau }{K}   \cr   
      & \stackrel{\mathrm{(iii)}}{\le} \frac{{128} \eta^2 \muminavg(s,a) M \tau}{K(1-\gamma)^2} \sum_{l=1}^L \exp\left(-{1/2} (l-1) \eta  \muminavg(s,a) M \tau \right)   \cr   
      &\le \frac{{128} \eta^2 \muminavg(s,a) M \tau}{K(1-\gamma)^2} \frac{1}{1-\exp(-{1/2}\eta  \muminavg(s,a) M \tau )}  \cr
      &\stackrel{\mathrm{(iv)}}{\le} \frac{{128} \eta^2 \muminavg(s,a) M \tau}{K(1-\gamma)^2} \frac{{4}}{\eta  \muminavg(s,a) M \tau} 
      = \frac{{512}\eta}{K(1-\gamma)^2} \defn \sigma^2,
    \end{align}
    where (i) holds due to the fact that
    $\|\mathsf{Var}_{P}(V)\|_{\infty} \le \|P\|_1 (\|V\|_{\infty})^2 + (\|P\|_1 \|V\|_{\infty})^2 \le \frac{2}{(1-\gamma)^2}$
    because  $\|V\|_{\infty} \le \frac{1}{1-\gamma}$ (cf.~\eqref{eq:bounded_iterates}) and $\|P\|_1 \le 1$, 
    (ii) follows from \eqref{eq:asyncwq-xsum-2} in Lemma~\ref{lemma:asyncwq-xsum}, (iii) holds due to the range of $\cZ$ and $\cZ_0$ is bounded by $\exp(-{1/4}\eta  \muminavg(s,a) M \tau )$ and $1$, respectively, and (iv) holds since $e^{-x} \le 1-\frac{1}{2}x$ for any $0\le x \le 1$ and { $1/2 \eta  \muminavg(s,a) M \tau  \le 1 $ }.

Now, by substituting the above bounds of $W_t$ and $B_t$ into Freedman's inequality (see Theorem~\ref{thm:Freedman}) and setting $m=1$, it follows that for any $s\in \cS$, $a \in \cA$, $t\in [T]$ and $\widehat{y}_{\bm{z}} \in \widehat{\cY}$,  
\begin{align} \label{eq:asyncwq-error2a}
  \left | \sum_{k=1}^K \sum_{u=0}^{t-1} \widehat{y}_{u,t}^k(s,a; \bz) \right |
  &\le \sqrt{8 \max{\{W_t(s,a), \frac{\sigma^2}{2^m} \}} \log{\frac{4m|\cS||\cA|T |\widehat{\cY}|}{\delta}}} + \frac{4}{3} B_t(s,a) \log{\frac{4m|\cS||\cA|T |\widehat{\cY}|}{\delta}} \cr
  &\le \sqrt{{4096} \frac{\eta}{K(1-\gamma)^2} \log{\frac{4|\cS||\cA|T|\widehat{\cY}|}{\delta}}} + \frac{{24} \eta}{K(1-\gamma)} \log{\frac{4|\cS||\cA|T|\widehat{\cY}|}{\delta}} \cr
  &\stackrel{\mathrm{(i)}}{\le} \frac{{78}}{(1-\gamma)} \sqrt{\frac{\eta L }{K} \log{\frac{4|\cS||\cA|T^2 K}{\delta}}} ,
\end{align}
with at least probability $1-\frac{\delta}{|\cS||\cA|T|\widehat{\cY}|}$, where (i) holds because $|\widehat{\cY}| \le (TK)^L $ given that {$\eta \muminavg(s,a) M\tau \le 1/4$}, and ${\frac{4\eta L}{K} \log{\frac{4|\cS||\cA|T^2 K}{\delta}} \le 1}$. Therefore, it follows that \eqref{eq:banana_peel} holds.

\paragraph{Step 4: putting things together.}
We now putting all the results obtained in the previous steps together to achieve the claimed bound.
Under $\cB_M$, there always exists $\widehat{y}_{ \bm{z}}: =\widehat{y}_{ \bm{z}(\widetilde{y})} \in \widehat{\cY}$ such that \eqref{eq:peach} holds. Hence, setting {$q=\frac{2064}{(1-\gamma)} \sqrt{\frac{\eta }{K} \log{(TK)} \log{\frac{4|\cS||\cA|T^2 K}{\delta}}}$},
\begin{align} 
  \sum_{k=1}^K \sum_{u=0}^{t-1}    \widetilde{y}_{u,t}^k(s,a) & \leq   \left | \sum_{k=1}^K \sum_{u=0}^{t-1} \widehat{y}_{u,t}^k(s,a; \bz) \right |  + D(\widetilde{y}, \widehat{y}_{\bm{z}} )  \nonumber \\
  & \leq \frac{{78}}{(1-\gamma)} \sqrt{\frac{\eta L }{K} \log{\frac{4|\cS||\cA|T^2 K}{\delta}}} +  \frac{{129}}{1-\gamma} \sqrt{\frac{ L\eta}{K}}\nonumber \\
  & \leq \frac{2064}{(1-\gamma)} \sqrt{\frac{\eta }{K} \log{(TK)} \log{\frac{4|\cS||\cA|T^2 K}{\delta}}},
\end{align}
where the second line holds due to  \eqref{eq:banana_peel} and \eqref{eq:peach}, and the last line holds due to {$L \le 64 \log{(TK)}$}. By taking a union bound over all $(s,a) \in \cS \times \cA$ and $t \in [T]$, we complete the proof.

\subsubsection{Proof of Lemma~\ref{lemma:asyncwq-xsum}}
\label{proof:asyncwq-xsum}

For notational simplicity, let $\overline{h}$ be the largest integer among $h \in (h_0, \nsyn(t)-(l-1)M)$ such that
\begin{align} \label{eq:asyncwq-hclip-def}
     \sum_{k=1}^K N_{h_0 \tau, (h-1)\tau}^{k}(s,a) \le {2} K \muminavg(s,a) M\tau .
\end{align}
Then, the following holds:
\begin{align} \label{eq:asyncwq-clipped-numvisit}
    \sum_{k=1}^K N_{h_0 \tau, \overline{h}\tau}^{k}(s,a) 
    &=  \sum_{k=1}^K N_{(\overline{h}-1)\tau, \overline{h}\tau}^{k}(s,a) + \sum_{k=1}^K N_{h_0 \tau, (\overline{h}-1)\tau}^{k}(s,a) \cr
    &\le K \tau + {2} K \muminavg(s,a) M\tau .
\end{align}
Also, for the following proofs, we provide an useful bound as follows:
\begin{align} \label{eq:x_numerator_bound}
        \sum_{k'=1}^K \frac{(1-\eta)^{-N_{h \tau, (h+1)\tau}^{k'}(s,a)}}{K} 
        \le  \frac{ \sum_{k'=1}^K e^{\eta N_{h \tau, (h+1)\tau}^{k'}(s,a)} }{K}
        &\le 1+ 2 \eta \frac{\sum_{k'=1}^K  N_{h \tau, (h+1)\tau}^{k'}(s,a) }{K} \cr
        &\le \exp \left (2 \eta \frac{\sum_{k'=1}^K  N_{h \tau, (h+1)\tau}^{k'}(s,a) }{K} \right),
\end{align}
which holds since $1+x \le e^x \le 1+2x$ for any $x \in [0,1]$ and {$\eta N_{h \tau, (h+1)\tau}^{k'}(s,a) \le \eta \tau \le 1$}.

According to \eqref{eq:asyncwq-xclip}, for any integer $u \in [\overline{h} \tau, t-(l-1)M\tau) $, $\widehat{x}_{u}^k(s,a)$ is clipped to zero. Now, we prove the bounds in Lemma~\ref{lemma:asyncwq-xsum} respectively. 

\paragraph{Proof of \eqref{eq:asyncwq-xmax}.} For $u \in [h_0\tau, \overline{h} \tau)$,
    \begin{align} \label{eq:x_max}
    \widehat{x}_{u}^k(s,a) 
    &= \prod_{h=h_0}^{\nsyn(u)-1} \left (\frac{\sum_{k'=1}^K (1-\eta)^{-N_{h \tau, (h+1)\tau}^{k'}(s,a)}}{K} \right ) \frac{\eta(1-\eta)^{-N_{\nsyn(u) \tau , u+1}^k(s,a)}}{K} \cr
    &\stackrel{\mathrm{(i)}}{\le} \prod_{h=h_0}^{\nsyn(u)-1} \left (\frac{\sum_{k'=1}^K (1-\eta)^{-N_{h \tau, (h+1)\tau}^{k'}(s,a)}}{K} \right ) \frac{3\eta}{K} \cr
    &\stackrel{\mathrm{(ii)}}{\le}  \exp \left (\frac{ 2\eta}{K} \sum_{k'=1}^K   N_{h_0 \tau, (\overline{h}-1)\tau}^{k'}(s,a)  \right )  \frac{3\eta}{K}  \cr
    &\stackrel{\mathrm{(iii)}}{\le}  \exp( {4} \eta \muminavg(s,a) M \tau)    \frac{3\eta}{K}   
     \stackrel{\mathrm{(iv)}}{\le}  \frac{9\eta}{K} ,
    \end{align}
    where (i) holds since $(1+\eta)^x \le e^{\eta x}$ and {$\eta N_{\nsyn(u) \tau , u+1}^k(s,a) \le \eta \tau \le 1$}, (ii) holds due to \eqref{eq:x_numerator_bound} and the fact that $\nsyn(u) \le \overline{h}-1$, (iii) follows from the definition of $\overline{h}$ in \eqref{eq:asyncwq-hclip-def}, and (iv) holds because {$4 \eta \muminavg(s,a) M\tau \le 1$}.
    
\paragraph{Proof of \eqref{eq:asyncwq-xsum-1}.} By the definition of $\overline{h}$, it follows that 
    \begin{align*}
        \sum_{h=h_0}^{\nsyn(t)-(l-1)M-1} \sum_{u \in \cU_{h\tau, (h+1)\tau}^k(s,a)}  \sum_{k=1}^K \widehat{x}_{u}^k(s,a)
        = \sum_{h=h_0}^{\overline{h}-1} \sum_{u \in \cU_{h\tau, (h+1)\tau}^k(s,a)}  \sum_{k=1}^K x_{u}^k(s,a) .
    \end{align*}
    Using the following relation for each $h$:
    \begin{align*}
        &\sum_{u \in \cU_{h\tau, (h+1)\tau}^k(s,a)}  \sum_{k=1}^K x_{u}^k(s,a) \cr
        &=  \left( \prod_{h'=h_0}^{h-1} \frac{\sum_{k'=1}^K (1-\eta)^{-N_{h' \tau, (h'+1)\tau}^{k'}(s,a)}}{K} \right )  \sum_{k=1}^K \frac{ \sum_{u \in \cU_{h\tau, (h+1)\tau}^k(s,a)} \eta(1-\eta)^{-N_{h \tau , u+1}^k(s,a)}}{K} \cr
        &= \left( \prod_{h'=h_0}^{h-1} \frac{\sum_{k'=1}^K (1-\eta)^{-N_{h' \tau, (h'+1)\tau}^{k'}(s,a)}}{K} \right )  \sum_{k=1}^K \frac{ (1-\eta)^{-N_{h \tau , (h+1) \tau}^k(s,a)} -1 }{K} \cr
        &= \left( \prod_{h'=h_0}^{h} \frac{\sum_{k'=1}^K (1-\eta)^{-N_{h' \tau, (h'+1)\tau}^{k'}(s,a)}}{K} \right )  - \left( \prod_{h'=h_0}^{h-1} \frac{\sum_{k'=1}^K (1-\eta)^{-N_{h' \tau, (h'+1)\tau}^{k'}(s,a)}}{K} \right ),
    \end{align*}
    and applying \eqref{eq:x_numerator_bound}, we can complete the proof as follows:
    \begin{align*}
        \sum_{h=h_0}^{\overline{h}-1} \sum_{u \in \cU_{h\tau, (h+1)\tau}^k(s,a)}  \sum_{k=1}^K x_{u}^k(s,a)
        &\le \prod_{h'=h_0}^{\overline{h}-1} \exp \left (\frac{2\eta \sum_{k'=1}^K  N_{h' \tau, (h'+1)\tau}^{k'}(s,a)}{K} \right )   - 1 \cr
        &\le \exp \left (\frac{2\eta \sum_{k'=1}^K  N_{h_0\tau, \overline{h}\tau}^{k'}(s,a)}{K} \right )   - 1 \cr
        &\stackrel{\mathrm{(i)}}{\le} \exp \left ( {4}\eta \muminavg(s,a) M\tau + 2\eta \tau \right )   - 1  \cr
        &\stackrel{\mathrm{(ii)}}{\le} {16}\eta \muminavg(s,a) M\tau,
    \end{align*}
    where (i) follows from \eqref{eq:asyncwq-clipped-numvisit}, and (ii) holds because $e^x \le 1+2x$ for any $x \in [0,1]$ and {$2 \eta \tau \le 4 \eta \muminavg(s,a) M\tau \le 1/2$}.

\paragraph{Proof of \eqref{eq:asyncwq-xsum-2}.}
Similarly,
    \begin{align*}
        \sum_{h=h_0}^{\nsyn(t)-(l-1)M-1} \sum_{u \in \cU_{h\tau, (h+1)\tau}^k(s,a)}  \sum_{k=1}^K (\widehat{x}_{u}^k(s,a))^2
        = \sum_{h=h_0}^{\overline{h}-1} \sum_{u \in \cU_{h\tau, (h+1)\tau}^k(s,a)}  \sum_{k=1}^K (x_{u}^k(s,a) )^2.
    \end{align*}
    Using the following relation for each $h$:
    \begin{align} \label{eq:x_sqsum_tau}
    &\sum_{u \in \cU_{h\tau, (h+1)\tau}^k(s,a)}  \sum_{k=1}^K (x_{u}^k(s,a) )^2 \cr
    &= \left( \prod_{h'=h_0}^{h-1} \frac{\sum_{k'=1}^K (1-\eta)^{-N_{h' \tau, (h'+1)\tau}^{k'}(s,a)}}{K} \right )^2  \sum_{k=1}^K \frac{ \sum_{u \in \cU_{h\tau, (h+1)\tau}^k(s,a)} \eta^2 (1-\eta)^{-2N_{h \tau , u+1}^k(s,a)}}{K^2} \cr
    &\le \left( \prod_{h'=h_0}^{h-1} \frac{\sum_{k'=1}^K (1-\eta)^{-N_{h' \tau, (h'+1)\tau}^{k'}(s,a)}}{K} \right )^2  \sum_{k=1}^K \frac{  \eta ((1-\eta)^{-2N_{h \tau , (h+1)\tau }^k(s,a)} -1 )}{K^2} \cr
    &\le \frac{\eta}{K} \left( \prod_{h'=h_0}^{h-1} \exp \left (2 \eta \frac{\sum_{k'=1}^K  N_{h' \tau, (h'+1)\tau}^{k'}(s,a) }{K} \right ) \right )^2  
     \left (  \exp \left (4 \eta \frac{\sum_{k'=1}^K  N_{h \tau, (h+1)\tau}^{k'}(s,a) }{K} \right )   -1 \right )\cr
    &= \frac{\eta}{K}  \exp \left (4 \eta \frac{\sum_{k'=1}^K  N_{h_0 \tau, h\tau}^{k'}(s,a) }{K} \right )   
     \left (  \exp \left (4 \eta \frac{\sum_{k'=1}^K  N_{h \tau, (h+1)\tau}^{k'}(s,a) }{K} \right )   -1 \right )\cr 
    &= \frac{\eta}{K}  \left ( \exp \left (4 \eta \frac{\sum_{k'=1}^K  N_{h_0 \tau, (h+1)\tau}^{k'}(s,a) }{K} \right )   
       - \exp \left (4 \eta \frac{\sum_{k'=1}^K  N_{h_0 \tau, h\tau}^{k'}(s,a) }{K} \right ) \right ),
    \end{align}
    where the inequality is derived similarly to \eqref{eq:x_numerator_bound} under the condition {$2\eta \tau \le 1$}, 
    we can complete the proof as follows:
    \begin{align} \label{eq:x_sqsum}
    \sum_{h=h_0}^{\overline{h}-1} \sum_{u \in \cU_{h\tau, (h+1)\tau}^k(s,a)}  \sum_{k=1}^K (x_{u}^k(s,a) )^2
    &\le \frac{\eta}{K}  \left ( \exp \left (4 \eta \frac{\sum_{k'=1}^K  N_{h_0 \tau, \overline{h}\tau}^{k'}(s,a) }{K} \right )   
       - 1 \right ) \cr
    &\stackrel{\mathrm{(i)}}{\le} \frac{\eta}{K}  \left ( \exp \left ({8} \eta \muminavg(s,a) M \tau + 4\eta \tau \right )   
       - 1 \right ) \cr   
    &\stackrel{\mathrm{(ii)}}{\le} \frac{ {64}  \eta^2 \muminavg(s,a) M \tau }{K}  ,
    \end{align}
    where (i) follows from \eqref{eq:asyncwq-clipped-numvisit}, and (ii) holds because $e^x \le 1+4x$ for any $x \in [0,2]$ and {$4\eta \tau \le 8 \eta \muminavg(s,a) M\tau  \le 1$}.

\subsection{Proof of Lemma~\ref{lemma:asyncwq-e3}}
\label{proof:asyncwq-e3}

The proof follows a similar structure to that of Lemma~\ref{lemma:asyncvq-e3}. We omit common parts of the proofs and refer to Appendix~\ref{proof:asyncvq-e3} to check the detailed derivations. 
First, we decompose the error term as follows:
\begin{align} \label{eq:asyncwq-e3-decomp}
  E_t^3(s,a)
  &= \underbrace{ \gamma \sum_{k=1}^K \sum_{u \in \cU_{0, (\nsyn(t)-\beta)\tau}^k(s,a)} \widetilde{\omega}_{u,t}^k(s,a)  P(s,a) (V^{\star} - V^k_{u})  }_{=: E_t^{3a}(s,a)} \cr
  &\qquad\qquad  + \underbrace{  \gamma \sum_{k=1}^K \sum_{u \in \cU_{(\nsyn(t)-\beta)\tau, t}^k(s,a)} \widetilde{\omega}_{u,t}^k(s,a)  P(s,a) (V^{\star} - V^k_{u})  . }_{=: E_t^{3b}(s,a)}
\end{align}

We shall bound these two terms separately.

\begin{itemize}
    \item \textbf{Bounding $E_t^{3a}(s,a)$.} First, the bound of $E_t^{3a}(s,a)$ is derived as follows:  
\begin{align}
  |E_t^{3a}(s,a)|
  &\le \gamma \sum_{k=1}^K \sum_{u \in \cU_{0, (\nsyn(t)-\beta)\tau}^k(s,a)} \widetilde{\omega}_{u,t}^k(s,a) \| P(s,a) \|_{1} \| V^{\star} - V^k_{u} \|_{\infty}\cr
  &\stackrel{\mathrm{(i)}}{\le} \frac{2}{1-\gamma} (1-\eta)^{ \frac{1}{K}\sum_{k=1}^K N_{(\nsyn(t)-\beta)\tau, t}^{k}(s,a)} \cr
  &\stackrel{\mathrm{(ii)}}{\le} \frac{2}{1-\gamma} (1-\eta)^{ \frac{\muminavg \beta \tau }{4} },
\end{align}
where 
(i) holds due to  Lemma~\ref{lemma:asyncwq-wsum} (cf. \eqref{eq:asyncwq-wsum-1-part}), and (ii) follows fromapplying Lemma~\ref{lemma:dist-asyncwq-multi-visit-mix} that with probability at least $1-\delta$,
$$\sum_{k=1}^K N_{(\nsyn(t)-\beta)\tau, t}^{k}(s,a) \ge \frac{K \beta \tau \muminavg}{4}$$ holds for all $(s,a) \in \cS \times \cA$ and $0 \le u < v \le T$ as long as {$\beta \tau \ge \tth$}.

    \item \textbf{Bounding $E_t^{3b}(s,a)$.} Combining \eqref{eq:asyncq-vgap-indiv-bound} and Lemma~\ref{lemma:asyncq-tdgap-indiv} to bound $\|V^{\star} - V^k_{u} \|_{\infty}$, we bound $E_t^{3b}(s,a)$ as follows:
\begin{align} \label{eq:asyncwq-e3b-decomp}
  |E_t^{3b}(s,a)|
  &\le \gamma \sum_{k=1}^K \sum_{u \in \cU_{(\nsyn(t)-\beta)\tau, t}^k(s,a)} \widetilde{\omega}_{u,t}^k(s,a)  \left \|V^{\star} - V^k_{u} \right \|_{\infty} \cr
  &\le \gamma \sum_{k=1}^K \sum_{h=\nsyn(t)-\beta}^{\nsyn(t) - 1} \sum_{u \in \cU_{h\tau, (h+1)\tau}^k(s,a)} \widetilde{\omega}_{u,t}^k(s,a)  ((1+2\eta \tau)\| \Delta_{h\tau} \|_{\infty} + \localvar) \cr
  &\le \localvar + \frac{1+\gamma}{2} \max_{\nsyn(t)-\beta \le h < \nsyn(t)}\| \Delta_{h\tau} \|_{\infty} 
\end{align}
where we denote $\localvar \defn \frac{8 \gamma \eta  \sqrt{\tau-1}}{1-\gamma}\sqrt{\log{\frac{2|\cS||\cA|T K }{\delta}}} $ for notational simplicity, and the last inequality follows from Lemma~\ref{lemma:asyncwq-wsum} (cf. \eqref{eq:asyncwq-wsum-1}) and the assumption that {$\eta \le \frac{1-\gamma}{4\gamma \tau}$}.

\end{itemize}

Now we have the bounds of $E_t^{3a}(s,a)$ and $E_t^{3b}(s,a)$ separately obtained above. By combining the bounds in \eqref{eq:asyncwq-e3-decomp}, we can claim the advertised bound, which completes the proof.